\documentclass[lettersize,journal]{IEEEtran}
\usepackage{amssymb,amsmath,amsthm,amsfonts}
\usepackage{algorithmic}
\usepackage{array}
\usepackage[caption=false,font=normalsize,labelfont=sf,textfont=sf]{subfig}
\usepackage{textcomp}
\usepackage{stfloats}
\usepackage{url}
\usepackage{verbatim}
\usepackage{graphicx}
\usepackage{cite}
\usepackage{lipsum}
\usepackage{booktabs}
\usepackage{multicol}
\usepackage{multirow}
\usepackage{hyperref}
\usepackage{float}
\usepackage{threeparttable}

\usepackage{changes}
\usepackage{morefloats}

\newtheorem{assumption}{Assumption}
\newtheorem{definition}{Definition}
\newtheorem{theorem}{Theorem}
\newtheorem{lemma}{Lemma}

\newtheorem{example}{Example}
\newtheorem{remark}{Remark}
\newtheorem{corollary}{Corollary}

\def\BibTeX{{\rm B\kern-.05em{\sc i\kern-.025em b}\kern-.08em
    T\kern-.1667em\lower.7ex\hbox{E}\kern-.125emX}}
\usepackage{balance}

\begin{document}
\title{Byzantine-Robust Distributed Online Learning: Taming Adversarial
	Participants in \\ An Adversarial Environment}
\author{Xingrong Dong, Zhaoxian Wu, Qing Ling, and Zhi Tian
\thanks{Xingrong Dong, Zhaoxian Wu and Qing Ling are with the School of Computer Science and Engineering and the Guangdong Provincial Key Laboratory of Computational Science, Sun Yat-Sen University, Guangzhou, Guangdong 510006, China. Zhi Tian is with the Department of Electrical and Computer Engineering, George Mason University, Fairfax, VA 22030, USA. Corresponding author: Qing Ling (lingqing556@mail.sysu.edu.cn).}
\thanks{Qing Ling is supported by National Natural Science Foundation of China grants 61973324 and 12126610, Guangdong Basic and Applied Basic Research Foundation grant 2021B1515020094, and Guangdong Provincial Key Laboratory of Computational Science grant 2020B1212060032. A short preliminary version of this paper has appeared in ICASSP 2023 \cite{Dong2023icassp}.}}


\maketitle

\begin{abstract}
This paper studies distributed online learning under Byzantine attacks. The performance of an online learning algorithm is often characterized by (adversarial) regret, which
evaluates the quality of one-step-ahead decision-making when an environment incurs adversarial losses, and a sublinear regret bound is preferred. But we prove that, even with a class of state-of-the-art robust aggregation rules, in an adversarial environment and in the presence of Byzantine participants, distributed online gradient descent can only achieve a linear adversarial regret bound, which is tight.
This is the inevitable consequence of Byzantine attacks, even though we can control the constant of the linear adversarial regret to a reasonable level.
Interestingly, when the environment is not fully adversarial so that the losses of the honest participants are i.i.d. (independent and identically distributed), we show that sublinear stochastic regret, in contrast to the aforementioned adversarial regret, is possible.
We develop Byzantine-robust distributed online momentum algorithms to attain such sublinear stochastic regret bounds for a class of robust aggregation rules. Numerical experiments corroborate our theoretical analysis.
\end{abstract}


\vspace{-0.1em}

\section{Introduction}
\IEEEPARstart{O}{nline} learning is a powerful tool to process streaming data in a timely manner \cite{zinkevich2003online,hazan2007logarithmic,hazan2016introduction}.
In response to an environment that provides (adversarial) losses sequentially, an online learning algorithm makes one-step-ahead decisions. Its performance is characterized by (adversarial) regret, which measures the accumulative difference between the losses of the online decisions and those of the overall best solution in hindsight.
It is preferred that the adversarial regret increases sublinearly in time, which would lead to asymptotically vanishing performance degradation.
When the streaming data are separately collected by multiple participants and data privacy is a concern, distributed online learning becomes a natural choice \cite{fosson2020centralized,li2019online,paternain2020constrained,yi2020distributed}.
Each participant makes a local decision, and a server aggregates all the local decisions to a global one \cite{tsianos2012distributed,hosseini2016online}.
Exemplary applications include online web ranking and online advertisement recommendation, etc \cite{shalev2012online,dekel2006online,jin2015collaborating,chen2020asynchronous}.

In addition to the sequential losses caused by the adversarial environment, distributed online learning faces a new challenge in terms of robustness, because not all the participants are guaranteed to be trustworthy. Some participants may intentionally or unintentionally send wrong messages, instead of true local decisions, to the server. These adversarial participants are termed as Byzantine participants following the notion in distributed systems to describe the worst-case attacks \cite{lamport1982byzantine}.
Therefore, an interesting question arises: \textit{Is it possible to develop a Byzantine-robust distributed online learning algorithm with provable sublinear adversarial regret, in an adversarial environment and in the presence of adversarial participants?}

In this paper, we provide a rather negative answer to this question. We show that even equipped with a class of state-of-the-art robust aggregation rules, distributed online gradient descent algorithms can only achieve linear adversarial regret bounds, which are tight. This rather negative result highlights the difficulty of Byzantine-robust distributed online learning. The joint impact from the adversarial environment and the adversarial participants leads the online decisions to deviate from the overall best solution in hindsight, no matter how long the learning time is. Nevertheless, we stress that it is the necessary price for handling arbitrarily malicious Byzantine attacks from the adversarial participants, and with the help of the state-of-the-art robust aggregation rules, we can control the constant of linear adversarial regret to a reasonable value.

On the other hand, we further show that if the environment is not fully adversarial so that the losses of the honest participants are i.i.d. (independent and identically distributed), then sublinear stochastic regret \cite{hazan2011beyond}, in contrast to the aforementioned adversarial regret, is possible. Accordingly, we develop a family of Byzantine-robust distributed online gradient descent algorithms enhanced with momentum to attain such sublinear stochastic regret bounds.

The rest of this paper is organized as follows. We briefly survey the related works in Section \ref{Related works}, and give the problem statement in Section \ref{Problem Statement}. The linear adversarial regret bounds of Byzantine-robust distributed online gradient descent are established in Section \ref{section-linear}, and the sublinear stochastic regret bounds of Byzantine-robust distributed online momentum are shown in Section \ref{section-sublinear}. We conduct numerical experiments in Section \ref{Numerical Experiments}, followed by conclusions in Section \ref{Conclusions}.

\section{Related works}
\label{Related works}

Online learning aims at sequentially making one-step-ahead decisions in an environment that provides (adversarial) losses. Classical online learning algorithms include but are not limited to online gradient descent \cite{mokhtari2016online}, online conditional gradient \cite{garber2016linearly}, online mirror descent \cite{hall2015online}, adaptive gradient \cite{ding2015adaptive}. We focus on online gradient descent and its variants in this paper. Their performance is often characterized by (adversarial) regret, which measures the accumulative difference between the losses of the online decisions and those of the overall best solution in hindsight. These algorithms have provable adversarial regret bounds of $\mathcal{O}(\sqrt{T})$ and $\mathcal{O}(\log T)$ for convex and strongly convex losses, respectively, where $T$ is the time horizon. When $T$ goes to infinity, such sublinear adversarial regret bounds imply asymptotically vanishing performance degradation in the long run.


When the streaming data are separately collected by multiple participants, data privacy becomes a big concern. Therefore, distributed online learning, which avoids transmitting raw data from the participants to the server, has attracted extensive research attention \cite{fosson2020centralized,li2019online}. Similar to their centralized counterparts, the distributed online gradient descent algorithms have provable adversarial regret bounds of $\mathcal{O}(\sqrt{T})$ and $\mathcal{O}(\log T)$ for convex and strongly convex losses, respectively \cite{wan2020projection,yan2012distributed}.

However, in a distributed online learning system, some of the participants can be adversarial. They do not follow the prescribed algorithmic protocol but send arbitrarily malicious messages to the server. We characterize these adversarial participants with the classical Byzantine attacks model \cite{lamport1982byzantine}. Interestingly, Byzantine-robust distributed online learning, which investigates reliable decision-making in an adversarial environment and in the presence of adversarial participants, is rarely studied. The work of \cite{ganesh2023online} focuses on the case that the environment provides linear losses, which is different to ours. The proposed asynchronous distributed online learning algorithm in \cite{ganesh2023online} also lacks regret bound analysis. The work of \cite{yao2022robust} considers online mean estimation over a decentralized network without a server. There is only one malicious participant, which has a limited budget to attack and only pollutes a faction of its messages to be transmitted. The performance metric is the Euclidean distance between the true mean and the estimate. In contrast, our work considers a general distributed online learning problem, the Byzantine participants have unlimited budgets to attack and can pollute all of their messages to be transmitted, and the performance metrics are adversarial and stochastic regrets.  
The work of \cite{sahoo2022distributed} considers decentralized online learning, but relaxes the problem to minimizing a convex combination of the losses. Accordingly, an $\mathcal{O}(\log^2 T)$ relaxed adversarial regret bound is established. In our work, we do not introduce any relaxation and the $\mathcal{O}(\log^2 T)$ relaxed adversarial regret bound in \cite{sahoo2022distributed} is not comparable to ours. The work of \cite{odeyomi2023byzantine} also considers decentralized online learning, but confines the number of Byzantine participants to be small so as to establish a dynamic regret bound. In contrast, our analysis on the static regret bounds allows nearly up to half of the participants to be Byzantine. Byzantine-robust decentralized meta learning is investigated in \cite{odeyomi2023online}, and a stochastic regret bound is established.


Several recent works investigate distributed bandits under Byzantine attacks. Different from online learning, participants receive values of losses, instead of gradients or functions, from an environment. It has been shown in \cite{kapoor2019corruption} that the proposed Byzantine-robust algorithms have linear adversarial regret bounds for multi-armed and linear-contextual problems. This is consistent with our result. Some works make the i.i.d. assumption \cite{jadbabaie2022byzantine,mitra2022collaborative,zhu2023byzantine}. The work of \cite{jadbabaie2022byzantine} proves $\mathcal{O}(T^{3/4})$ regret for linear bandits with high probability. The work of \cite{mitra2022collaborative} reaches $\mathcal{O}(\sqrt{T})$ regret but requires the action set to be finite. Our proposed algorithm, with the aid of momentum, attains the $\mathcal{O}(\sqrt{T})$ stochastic regret bound. The work of \cite{zhu2023byzantine} considers multi-armed bandits, and uses historic information to reach $\mathcal{O}(\log T)$ regret, which is consistent with our stochastic regret bound established for Byzantine-robust distributed online momentum.
The work of \cite{demirel2022federated} is free of the i.i.d. assumption, but the regret for multi-armed bandits is defined with respect to a suboptimal solution other than the optimal one. Therefore, the derived $\mathcal{O}(\log T)$ sublinear regret bound is not comparable to others.

Another tightly related area is Byzantine-robust distributed stochastic optimization \cite{pillutla2022robust,yu2023secure,jian2022filter}.
Therein, the basic idea is to replace the vulnerable mean aggregation rule in distributed stochastic gradient descent with robust aggregation rules, including coordinate-wise median \cite{yin2018byzantine}, trimmed mean \cite{yin2018byzantine,su2020byzantine}, geometric median \cite{chen2017distributed}, Krum \cite{blanchard2017machine}, centered clipping \cite{karimireddy2021learning}, Phocas \cite{xie2018phocas}, FABA \cite{xia2019faba}, etc. Most of them belong to the category of robust bounded aggregation rules (see Definition \ref{def:robust-bounded}). We will incorporate these robust bounded aggregation rules with distributed online gradient descent and momentum to enable Byzantine-robustness.

In Table \ref{table:results}, we compare the adversarial regret bounds of distributed online gradient descent with the mean aggregation rule and without Byzantine attacks, the derived adversarial regret bounds of Byzantine-robust distributed online gradient descent with robust bounded aggregation rules, as well as the derived stochastic regret bounds of Byzantine-robust distributed online momentum with robust bounded aggregation rules.

\begin{table}
		\caption{Regret bounds of different algorithms}
	\label{table:results}
	\vspace{-1em}
	\begin{center}
	\begin{threeparttable}
	\begin{tabular}{@{}ccc@{}}
		\toprule
		           &  constant step size& diminishing step size\\ \midrule
		Byzantine-free mean\tnote{1} & $\mathcal{O}(\sqrt{T})$ & $\mathcal{O}(\log T)$ \\
		Byzantine-robust\tnote{2} & $\mathcal{O}(T)$        & $\mathcal{O}(T)$      \\
		Byzantine-robust momentum\tnote{3} &  $\mathcal{O}(\sqrt{T})$ & $\mathcal{O}(\log T)$ \\ \bottomrule
	\end{tabular}
\begin{tablenotes}
	\footnotesize
	\item[1] Adversarial regret bounds of distributed online gradient descent with the mean aggregation rule and without Byzantine attacks.
	\item[2] Adversarial regret bounds of Byzantine-robust distributed online gradient descent with robust bounded aggregation rules.
	\item[3] Stochastic regret bounds of Byzantine-robust distributed online momentum with robust bounded aggregation rules.
\end{tablenotes}
\end{threeparttable}
\end{center}
\end{table}

\section{Problem Statement}
\label{Problem Statement}

Consider $n$ participants in a set $\mathcal{N}$, among which $h$ are honest and in subset $\mathcal{H}$, while $b$ are Byzantine and in subset $\mathcal{B}$. We know $n=h+b$, but the identities of Byzantine participants are unknown and we can only roughly estimate an upper bound of $b$. At step $t$, each honest participant $j$ makes its local decision of the model parameters $w_t^j \in \mathbb{R}^d$ and sends it to the server, while each Byzantine participant $j$ sends an arbitrarily malicious message to the server. For notational convenience, denote $z_t^j \in \mathbb{R}^d$ as the message sent by participant $j$ to the server at step $t$, no matter whether it is from an honest or Byzantine participant. Upon receiving all $z_t^j$, the server aggregates them to yield a global decision of the model parameters $w_t \in \mathbb{R}^d$. The quality of the sequential decisions over $T$ steps is often evaluated by adversarial regret with respect to the overall best solution in hindsight, given by
\begin{align}
	R_T := \sum_{t=1}^{T}f_t(w_t) - \min_{w\in \mathbb{R}^d}\sum_{t=1}^{T}f_t(w),
	\label{doco}
\end{align}
where
\begin{align}
	f_t(w):= \frac{1}{h}\sum_{j\in \mathcal{H}}f_t^j(w),
\end{align}
and $f_t^j$ is the loss revealed to $j \in \mathcal{H}$ at the end of step $t$.

For distributed online gradient descent, each honest participant $j \in \mathcal{H}$ makes its local decision following
\begin{align}
	\label{eq:descent}
	w_{t+1}^j = w_t - \eta_t \nabla f_t^j(w_t),
\end{align}
where $\eta_t > 0$ is the step size, and sends $z_{t+1}^j = w_{t+1}^j$ to the server. The server aggregates the messages $z_{t+1}^j$ to yield the mean value
\begin{align}
	w_{t+1} = \frac{1}{n}\sum_{j=1}^{n} z_{t+1}^j.
	\label{eq:mean}
\end{align}
However, messages $z_{t+1}^j$ from the Byzantine participants $j \in \mathcal{B}$ are arbitrarily malicious, such that $w_{t+1}$ can be manipulated to reach arbitrarily large adversarial regret.

Motivated by the recent advances of Byzantine-robust distributed stochastic optimization, one may think of using robust aggregation rules to replace the vulnerable mean aggregation rule in \eqref{eq:mean}. Denote $AGG$ as a proper robust aggregation rule. Now, the server makes the decision as
\begin{align}
	\label{eq:robust}
	w_{t+1} = AGG(z_{t+1}^1,z_{t+1}^2,\cdots, z_{t+1}^n).
\end{align}

Below we introduce two exemplary robust aggregation rules. More examples can be found in Appendix \ref{C-alpha-proof}. For notational convenience, we denote $\mathcal{Z}_{t+1}:=\{z_{t+1}^1, z_{t+1}^2, \cdots, z_{t+1}^n\}$ as the set of the $n$ received messages and $\mathcal{Z}_{t+1}[k]:=\{z_{t+1}^1[k], z_{t+1}^2[k], \cdots, z_{t+1}^n[k]\}$ as the set of their $k$-th elements, where $k\in[d]$.

\noindent \textbf{Coordinate-wise median}. It yields the median for each dimension, given by
\begin{align}
   & \operatorname{coomed}({\mathcal{Z}_{t+1}}) \\
:= & [\operatorname{med}(\mathcal{Z}_{t+1}[1]); \operatorname{med}(\mathcal{Z}_{t+1}[2]); \cdots; \operatorname{med}(\mathcal{Z}_{t+1}[d])], \notag
\end{align}
where $\operatorname{med}(\cdot)$ calculates the median of the input scalars.

\noindent \textbf{Trimmed mean.} It is also coordinate-wise. Let $q < \frac{n}{2}$ be the estimated number of Byzantine participants. Given $\mathcal{Z}_{t+1}[k]$, trimmed mean removes the largest $q$ inputs and the smallest $q$ inputs, and then averages the rest to yield $\operatorname{trimean}(\mathcal{Z}_{t+1}[k])$. The results of the $d$ dimensions are stacked to yield
\begin{align}
\hspace{-1em}   & \operatorname{trimean}({\mathcal{Z}_{t+1}}) \\
\hspace{-1em}:= & [\operatorname{trimean}(\mathcal{Z}_{t+1}[1]); \operatorname{trimean}(\mathcal{Z}_{t+1}[2]); \cdots; \operatorname{trimean}(\mathcal{Z}_{t+1}[d])]. \notag
\end{align}

\section{Linear Adversarial Regret Bounds of Byzantine-Robust Distributed Online Gradient Descent}
\label{section-linear}

Robust aggregation rules have been proven effective in distributed stochastic optimization, given that the fraction of Byzantine participants $\alpha = \frac{b}{n}$ is less than $\frac{1}{2}$ \cite{yin2018byzantine,chen2017distributed,su2020byzantine,blanchard2017machine,karimireddy2021learning,xie2018phocas,xia2019faba}. Thus, one may wonder whether the Byzantine-robust distributed online gradient descent updates \eqref{eq:descent} and \eqref{eq:robust} can achieve sublinear adversarial regret.

Our answer is negative. Even with a wide class of \textit{robust bounded aggregation} rules, the tight adversarial regret bounds are linear.
\begin{definition}{(Robust bounded aggregation rule)} Consider $n$ messages $z_t^1,z_t^2,\cdots, z_t^n \in \mathbb{R}^d$ from $h$ honest participants in $\mathcal{H}$ and $b$ Byzantine participants in $\mathcal{B}$. The fraction of Byzantine participants $\alpha = \frac{b}{n} < \frac{1}{2}$.	An aggregation rule $AGG$ is a robust bounded aggregation rule, if the difference between its output and the mean of the honest messages is bounded by
	\begin{align}
		\nonumber \|w_t-\bar{z}_t\|^2 =& \|AGG(z_t^1,z_t^2,\cdots, z_t^n)-\bar{z}_t\|^2
		\leq C_{\alpha}^2\zeta^2,
	\end{align}
	where $\bar{z}_t := \frac{1}{h}\sum_{j\in\mathcal{H}}z_t^j$ is the mean of the honest messages, $\zeta^2$ is the largest deviation of the honest messages such that $\|\bar{z}_t - z_t^j \|^2 \leq \zeta^2$ for all $j \in \mathcal{H}$, and $C_{\alpha}$ is an aggregation-specific constant determined by $\alpha$.
	\label{def:robust-bounded}
\end{definition}

\begin{table}
	\caption{Constants $C_{\alpha}$ of robust bounded aggregation rules, with $\alpha$ being the fraction of Byzantine participants}
	\label{C-alpha}
	\vspace{-1em}
	\begin{center}
		\begin{small}
			\begin{tabular}{@{}cc@{}}
				\toprule
				                           & $C_{\alpha}$                                   \\ \midrule
				coordinate-wise median     & $\mathcal{O}(\frac{1}{1-\alpha})$                                  \\
				trimmed mean               & $\mathcal{O}(\frac{\sqrt{\alpha(1-\alpha)}}{(1-2\alpha)})$    \\
				geometric median           & $\mathcal{O}(\frac{1-\alpha}{1-2\alpha})$               \\
				Krum                       & $\mathcal{O}(1+\sqrt{\frac{1-\alpha}{1-2\alpha}})$                                       \\	
				centered clipping          & $\mathcal{O}(\sqrt{\alpha})$                                    \\
				Phocas                     & $\mathcal{O}(\sqrt{1+\frac{\alpha(1-\alpha)}{(1-2\alpha)^2}})$ \\	
				FABA                       & $\mathcal{O}(\frac{\alpha}{1-3\alpha})$                  \\
				\bottomrule
			\end{tabular}
		\end{small}
	\end{center}
	\vskip -0.1in
\end{table}

In Definition \ref{def:robust-bounded}, $\zeta^2$ characterizes the heterogeneity of the messages to be aggregated. For a robust bounded aggregation rule, the difference between its output and the mean of the honest messages is bounded by $C_{\alpha}^2 \zeta^2$. Therefore, a robust bounded aggregation rule with a smaller $C_{\alpha}^2$ can better handle the heterogeneity of the messages to be aggregated.

We show that a number of state-of-the-art robust aggregation rules, including coordinate-wise median \cite{yin2018byzantine}, trimmed mean \cite{yin2018byzantine,su2020byzantine}, geometric median \cite{chen2017distributed}, Krum \cite{blanchard2017machine}, centered clipping\footnote{Centered clipping requires $\alpha \leq 0.1$.} \cite{karimireddy2021learning}, Phocas \cite{xie2018phocas}, and FABA\footnote{FABA requires $\alpha < \frac{1}{3}$.} \cite{xia2019faba}, all belong to robust bounded aggregation rules. Their corresponding constants $C_\alpha$ are listed in Table \ref{C-alpha} and the derivations of these constants are in Appendix \ref{C-alpha-proof}. 

Note that $\bar{z}_t$ is only used for the purpose of theoretical analysis. The server does not need to calculate the value of $\bar{z}_t$ during implementing a robust bounded aggregation rule.

To analyze the adversarial regret bounds, we make the following standard assumptions on the losses of any honest participant $j \in \mathcal{H}$.

\begin{assumption}[$L$-smoothness]
	$f_t^j$ is differentiable and has Lipschitz continuous gradients. For any $x, y \in \mathbb{R}^d$, there exists a constant $L > 0$ such that	
	\begin{equation}
		\vert| \nabla f_t^j(x) - \nabla f_t^j(y) \vert| \leq L \vert| x- y\vert| .
		\label{ass-1-equa}
	\end{equation}
	\label{ass-1}
\end{assumption}

\vspace{-2em}

\begin{assumption}[$\mu$-strong convexity]
	$f_t^j$ is strongly convex. For any $x, y \in \mathbb{R}^d$, there exists a constant $\mu > 0$ such that
	\begin{align}
		\langle \nabla{f_t^j(x)},x-y\rangle \geq f_t^j(x) - f_t^j(y) + \frac{\mu}{2}\|x-y\|^2.
		\label{ass-2-equa}
	\end{align}
	\label{ass-2}
\end{assumption}

\vspace{-2em}


\begin{assumption}[Bounded deviation]
	Define $\nabla f_t(w_t):= \frac{1}{h}\sum_{j\in \mathcal{H}}$ $\nabla{f_t^j(w_t)}$. The deviation between each honest gradient $\nabla{f_t^j(w_t)}$ and the mean of the honest gradients is bounded by
	\begin{align}
		\vert| \nabla{f_t^j(w_t)}- \nabla f_t(w_t) \vert|^2\leq \sigma^2.
		\label{ass-3-equa}
	\end{align}
	\label{ass-3}
\end{assumption}

\vspace{-2em}

\begin{assumption}[Bounded gradient at the overall best solution]
	Define $w^* = {\arg\min}_{w\in \mathbb{R}^d}\sum_{t=1}^T f_t(w)$ as the overall best solution. The mean of the honest gradients at this point is upper bounded by
	\begin{align}
		\|\frac{1}{h}\sum_{j\in \mathcal{H}}\nabla{f_t^j(w^*)}\|^2 \leq  \xi^2.
		\label{ass-4-equa}
	\end{align}
	\label{ass-4}
\end{assumption}

\vspace{-1em}

These assumptions are common in the analysis of online learning algorithms. Some works make stronger assumptions \cite{zinkevich2003online,hazan2007logarithmic,hazan2016introduction}, for example, bounded variable or bounded gradient that yields Assumptions \ref{ass-3} and \ref{ass-4}.

Next, we show that the Byzantine-robust distributed online gradient descent algorithm with a robust bounded aggregation rule can only reach a linear adversarial regret bound under Byzantine attacks. The proof is in Appendix \ref{theorem-T}. In contrast, the distributed online gradient descent algorithm with the mean aggregation rule can achieve a sublinear adversarial regret bound without Byzantine attacks, as shown in Appendix \ref{app:corollary-DOGD}. To distinguish the adversarial regret bounds with different step sizes, we denote $R_{T:\eta}, R_{T:\frac{1}{\sqrt{T}}}, R_{T:\frac{1}{t}}$ as the adversarial regrets with a constant step size $\eta$, a special constant step size $ \mathcal{O}(\frac{1}{\sqrt{T}})$, and a diminishing step size $\mathcal{O}(\frac{1}{t})$, respectively.

\begin{theorem} 	
	\label{theorem:bound-T}
	Suppose that the fraction of Byzantine participants $\alpha = \frac{b}{n} < \frac{1}{2}$. Under Assumptions \ref{ass-1}, \ref{ass-2}, \ref{ass-3}, and \ref{ass-4}, the Byzantine-robust distributed online gradient descent updates \eqref{eq:descent} and \eqref{eq:robust} with a robust bounded aggregation and a constant step size $\eta_t = \eta \in (0,\frac{1}{4L}]$ have an adversarial regret bound
	\begin{align}
		\hspace{-1.5em} R_{T:\eta} \leq & \frac{1}{\eta} \|w_1 - w^*\|^2 + \left(2\eta+\frac{8L^2\eta^2}{\mu}\right) \xi^2 T +\frac{2}{\mu}C_{\alpha}^2\sigma^2 T.
	\end{align}
	In particular, if $\eta_t = \eta = \frac{c}{\sqrt{T}}$ where $c$ is a sufficiently small positive constant, then the adversarial regret bound becomes
	\begin{align}
		 R_{T:\frac{1}{\sqrt{T}}} \leq & \frac{8L^2c^2}{\mu}\xi^2 + \left( \frac{\|w_1 - w^*\|^2}{c}+2c\xi^2 \right) \sqrt{T} \\
		\nonumber& + \frac{2}{\mu}C_{\alpha}^2\sigma^2 T.
	\end{align}
	If we use a diminishing step size $\eta_t = \min\{\frac{1}{4L},\frac{8}{\mu t}\}$, then the adversarial regret bound is
	\begin{align}
		 R_{T:\frac{1}{t}} \leq & 4L\|w_1 - w^*\|^2 + \frac{48L}{\mu^2}\xi^2\log T \\
        \nonumber& + \frac{2}{\mu}C_{\alpha}^2\sigma^2T.
	\end{align}
\end{theorem}

We construct the following counter-example to show that the derived $\mathcal{O}(\sigma^2 T)$ linear adversarial regret bound is tight.

\begin{example}
	\label{ex-sigma}
	Consider a distributed online learning system with $3$ participants, among which participant 3 is Byzantine. Thus, $\mathcal{N}= \{1,2,3\}$,  $\mathcal{H}= \{1,2\}$ and $\mathcal{B}= \{3\}$. Suppose that at any step $t$, the losses of participants 1 and 2 are respectively given by
	\begin{align}
		&f_t^1(w) = \frac{1}{2}(w-\sigma)^2, \quad
		&f_t^2(w) = \frac{1}{2}(w+\sigma)^2. \nonumber
	\end{align}
	It is easy to check that these losses satisfy Assumptions \ref{ass-1}, \ref{ass-2}, \ref{ass-3}, and \ref{ass-4}. To be specific, the overall best solution $w^*=0$, $L=1$, $\mu=1$, and $\xi^2=0$.
	
	Take geometric median as an exemplary aggregation rule. Suppose that the algorithm is initialized by $w_1 = \sigma$. At step $2$, participant $1$ sends $z_2^1 = w_2^1 = w_1 - \eta_1(w_1-\sigma) = \sigma$, while participant $2$ sends $z_2^2 = w_2^2 = w_1 - \eta_1(w_1+\sigma) = \sigma - 2 \eta_1 \sigma$. In this circumstance, participant $3$, who is Byzantine, can send $z_2^3 = \sigma$ so that the aggregation result is $w_2 = \sigma$. As such, for any step $t$, $w_t = \sigma$, $f_t(w_t) = \sigma^2$, $f_t(w^*) = \frac{1}{2}\sigma^2$, and the adversarial regret is $\frac{1}{2}\sigma^2 T$.
	
	For other robust bounded aggregation rules, we can observe that the mean of the honest messages is $\bar{z}_{t+1} = (1-\eta_t)\sigma$ and the largest deviation is $\zeta^2 = \eta_t^2\sigma^2$. According to Definition \ref{def:robust-bounded}, participant 3 can always manipulate its message so that the aggregation result is in the order of $\sigma$, which eventually yields linear adversarial regret. If the aggregation rule is majority-voting-based, such as coordinate-wise median and trimmed mean, sending $
	z_{t+1}^3 = \sigma$ is effective. For centered clipping, participant 3 can send $z_{t+1}^3 = \sigma + 2 \eta_t \sigma$ instead.
\end{example}

Note that Example \ref{ex-sigma} holds for both constant and diminishing step sizes. Meanwhile, Example \ref{ex-sigma} can be extended to a larger number of participants.


The linear adversarial regret bound seems frustrating, but is the necessary price for handling arbitrarily malicious Byzantine attacks from the adversarial participants. With the help of robust bounded aggregation rules, we are able to control the constant of linear adversarial regret to a reasonable value $\frac{2}{\mu}C_{\alpha}^2\sigma^2$, which is determined by the property of losses, the robust bounded aggregation rule, the fraction of Byzantine participants, and the gradient deviation among honest participants.

\section{Sublinear Stochastic Regret Bounds of Byzantine-Robust Distributed Online Momentum}
\label{section-sublinear}

According to Theorem \ref{theorem:bound-T}, the established linear adversarial regret bounds are proportional to $\sigma^2$, the maximum between the honest gradients and the mean of the honest gradients. This makes sense as the disagreement among the honest participants is critical, especially in an adversarial environment. This observation motivates us to investigate whether it is possible to attain sublinear regret bounds when the disagreement among the honest participants is well-controlled.

To this end, suppose that the environment provides all the honest participants with independent losses from the same distribution $\mathcal{D}$ at all steps. Define the expected loss $F(w):=\mathbb{E}_{\mathcal{D}}f_t^j(w)$ for all $j \in \mathcal{H}$ and all $t$. Then, stochastic regret is defined as
\begin{align}
	S_T:=\mathbb{E} \sum_{t=1}^T F(w_t)-T \cdot \min_{w\in\mathbb{R}^d}F(w),
\end{align}
where the expectation is taken over the stochastic process \cite{hazan2011beyond}. In such an i.i.d. setting, the notion of stochastic regret is natural and has been widely adopted \cite{hazan2011beyond, chen2023optimistic,li2023survey}.
Note that the works of \cite{jadbabaie2022byzantine} and \cite{mitra2022collaborative}, which investigate the problem of Byzantine-robust distributed bandits, also make a similar i.i.d. assumption.

However, naively applying robust bounded aggregation rules to \eqref{eq:descent} and \eqref{eq:robust} cannot guarantee sublinear stochastic regret, since the random perturbations of the honest losses still accumulate over time and the disagreement among the honest participants does not diminish.
Motivated by the successful applications of variance reduction techniques in Byzantine-robust distributed stochastic optimization \cite{wu2020federated,khanduri2019byzantine,el2020distributed,karimireddy2021learning,gorbunov2022variance}, we let each honest participant perform momentum steps, instead of gradient descent steps, to gradually eliminate the disagreement during the learning process.

In Byzantine-robust distributed online gradient descent with momentum, each honest participant $j$ maintains a momentum vector
\begin{align}
    \label{eq:momentum-def}
	m_t^j=\nu_t \nabla f_t^j(w_t)+(1-\nu_t) m_{t-1}^j,
\end{align}
where $\nu_t \in (0, 1)$ is the momentum parameter. Then, it makes the local decision following
\begin{align}
	\label{eq:momentum-descent}
	w_{t+1}^j = w_t - \eta_t m_t^j,
\end{align}
instead of \eqref{eq:descent} and sends to the server. The server still aggregates the messages and makes the decision as \eqref{eq:robust}.

The ensuing analysis needs the following assumptions on the expected loss, in lieu of Assumptions \ref{ass-1}, \ref{ass-2} and \ref{ass-3} on the individual losses.
\begin{assumption}[$L$-smoothness]
	$F$ is differentiable and has Lipschitz continuous gradients. For any $x, y \in \mathbb{R}^d$, there exists a constant $L > 0$ such that
	\begin{align}
		\vert| \nabla F(x) - \nabla F(y) \vert| \leq L \vert| x- y\vert|.
		\label{ass-1-new-equa}
	\end{align}
	\label{ass-1-new}
\end{assumption}

\vspace{-2em}

\begin{assumption}[$\mu$-strong convexity]
	$F$ is strongly convex. For any $x, y \in \mathbb{R}^d$, there exists a constant $\mu > 0$ such that
	\begin{align}
		\langle \nabla{F(x)},x-y\rangle \geq F(x) - F(y) + \frac{\mu}{2}\|x-y\|^2.
		\label{ass-2-new-equa}
	\end{align}
	\label{ass-2-new}
\end{assumption}

\vspace{-2em}

\begin{assumption}[Bounded variance]
	The variance of each honest gradient $\nabla{f_t^j(w_t)}$ is bounded by
	\begin{align}
		 \mathbb{E}\vert| \nabla{f_t^j(w_t)}- \nabla{F(w_t)} \vert|^2
		\leq \sigma^2.
		\label{ass-3-new-equa}
	\end{align}
	\label{ass-3-new}
\end{assumption}

\vspace{-1em}

In the investigated i.i.d. setting, the overall best solution $w^* = {\arg\min}_{w\in \mathbb{R}^d}$ $F(w)$ makes $\nabla F(w^*) = 0$, such that we no longer need to bound the gradient at the overall best solution as in Assumption \ref{ass-4}.

%
%

\begin{theorem}
	\label{theorem-DROGDM}
	Suppose that the fraction of Byzantine participants $\alpha = \frac{b}{n} < \frac{1}{2}$ and that each honest participant $j$ draws its loss $f_t^j$ at step $t$ from distribution $\mathcal{D}$ with expectation $F:=\mathbb{E}_{\mathcal{D}}f_t^j$. Under Assumptions \ref{ass-1-new}, \ref{ass-2-new} and \ref{ass-3-new}, the Byzantine-robust distributed online momentum updates \eqref{eq:momentum-def}, \eqref{eq:momentum-descent} and \eqref{eq:robust} with a robust bounded aggregation rule, a constant step size $\eta_t = \eta \in (0, \frac{\mu}{16 L^2})$ and a constant momentum parameter $\nu_{t} = \nu = \frac{8\sqrt{3}L^2}{\mu}\eta$ have a stochastic regret bound
	\begin{align}
		S_{T:\eta} \leq \mathcal{O}\left(\frac{1}{\eta}+\frac{\sigma^2}{h} \Big( 1+ h^2C_{\alpha}^2 \Big)\frac{L^4}{\mu^4}\eta T\right).
	\end{align}
In particular, if $\eta_t = \eta = \mathcal{O}(\frac{1}{\sqrt{T}})$ and $\nu_t = \nu = \mathcal{O}(\frac{1}{\sqrt{T}})$ are properly chosen, then the stochastic regret bound becomes
	\begin{align}
		\label{eq:sublinearbound}
		S_{T:\frac{1}{\sqrt{T}}}=\mathcal{O}\left(\sqrt{T}+\frac{\sigma^2}{h} \Big( 1+ h^2C_{\alpha}^2 \Big)\frac{L^4}{\mu^4}\sqrt{T}\right).
	\end{align}
If we use a proper diminishing step size $\eta_t=\mathcal{O}(\frac{1}{t})$ and a proper momentum parameter $\nu_t=\mathcal{O}(\frac{1}{t})$, then the stochastic regret bound is
	\begin{align}
		\label{eq:sublinearlogbound}		S_{T:\frac{1}{t}}=\mathcal{O}\left(\frac{\sigma^2}{h} \Big( 1+ h^2 C_{\alpha}^2 \Big)\frac{L^4}{\mu^4}\log T\right).
	\end{align}
\end{theorem}

The proof is left to Appendix \ref{stochasic regret analysis}.
With proper constant and diminishing step sizes, Theorem \ref{theorem-DROGDM} establishes the $\mathcal{O}(\sqrt{T})$ and $\mathcal{O}(\log{T})$ stochastic regret bounds of Byzantine-robust distributed online momentum in the i.i.d. setting. In the sublinear stochastic regret bounds \eqref{eq:sublinearbound} and \eqref{eq:sublinearlogbound}, the coefficient $\frac{\sigma^2}{h}$ is inversely proportional to $h$, the number of honest participants, which highlights the benefit of collaboration. The constant is also determined by $C_\alpha$ that characterizes the defense ability of the robust bounded aggregation rule. Smaller $C_\alpha$ yields smaller stochastic regret.
Besides, some robust bounded aggregation rules, including trimmed mean, centered clipping and FABA, have $C_\alpha=0$ when $\alpha=0$, namely, no Byzantine participants are present. In this case, the derived stochastic regret bounds respectively degenerate to $\mathcal{O} ((\sigma^2 / h)\sqrt T )$ and $\mathcal{O} ((\sigma^2 / h)\log T )$.

The i.i.d. assumption is essential to the sublinear stochastic regret bound. Without the i.i.d. assumption, we show that Byzantine-robust distributed online momentum has a tight linear stochastic regret bound in Example \ref{ex-noniid}, similar to the construction in Example \ref{ex-sigma}.

\begin{example}
	\label{ex-noniid}
	Consider a distributed online learning system with $3$ participants, among which participant 3 is Byzantine. Thus, $\mathcal{N}= \{1,2,3\}$,  $\mathcal{H}= \{1,2\}$ and $\mathcal{B}= \{3\}$. Suppose that at any step $t$, the losses of participants 1 and 2 are respectively given by
	\begin{align}
		&f_t^1(w) = \frac{1}{2}(w-\sigma)^2, \quad
		&f_t^2(w) = \frac{1}{2}(w+\sigma)^2. \nonumber
	\end{align}
	The losses of participants 1 and 2 are non-i.i.d. and the expected loss of the honest participants is
	\begin{align*}
		F_t(w) = \frac{1}{2}(\frac{1}{2}(w-\sigma)^2+\frac{1}{2}(w+\sigma)^2)=\frac{1}{2} (w^2 +\sigma^2).
	\end{align*}
	It is easy to check that these losses satisfy Assumptions \ref{ass-1-new}, \ref{ass-2-new} and \ref{ass-3-new}. To be specific, the overall best solution $w^*=0$, $L=1$, and $\mu=1$.
	
	Take geometric median as an exemplary aggregation rule. Suppose that the algorithm is initialized by $w_1 = \sigma$, $m_0^1 = 0$ and $m_0^2 = 2\sigma$. Thus,
$m_1^1 = \nu_1(w_1-\sigma)+(1-\nu_1)m_0^1=0$ and $m_1^2 = \nu_1(w_1+\sigma)+(1-\nu_1)m_0^2 = 2\sigma$. At step $2$, participant $1$ sends $z_2^1 = w_2^1 = w_1 - \eta_1 m_1^1 = \sigma$, while participant $2$ sends $z_2^2 = w_2^2 = w_1 - \eta_1 m_1^2  = \sigma - 2 \eta_1 \sigma$. In this circumstance, participant $3$, who is Byzantine, can send $z_2^3 = \sigma$ so that the aggregation result is $w_2 = \sigma$. As such, for any step $t$, $w_t = \sigma$, $F_t(w_t) = \sigma^2$, $F_t(w^*) = \frac{1}{2}\sigma^2$, and the stochastic regret is $\mathbb{E}\sum_{t=1}^{T}(F_t(w_t)-F_t(w^*))=\frac{1}{2}\sigma^2 T$.
	
	For other robust bounded aggregation rules, we can observe that the mean of the honest messages is $\bar{z}_{t+1}=(1-\eta_t)\sigma$ and the largest deviation is $\zeta^2 = \eta_t^2\sigma^2$. According to Definition \ref{def:robust-bounded}, participant 3 can always manipulate its message so that the aggregation result is in the order of $\sigma$, which eventually yields linear stochastic regret. If the aggregation rule is majority-voting-based, such as coordinate-wise median and trimmed mean, sending $
	z_{t+1}^3 = \sigma$ is effective. For centered clipping, participant 3 can send $z_{t+1}^3 = \sigma + 2 \eta_t \sigma$ instead.
\end{example}

But on the other hand, the momentum technique is critical to the sublinear stochastic regret bound. In contrast, we can show that Byzantine-robust distributed online gradient descent without momentum has an undesired tight linear stochastic regret bound even with the i.i.d. assumption; see Example \ref{ex-iid-without-momentum}.

\begin{example}
	\label{ex-iid-without-momentum}
	Consider a distributed online learning system with $3$ participants, among which participant 3 is Byzantine. Thus, $\mathcal{N}= \{1,2,3\}$,  $\mathcal{H}= \{1,2\}$ and $\mathcal{B}= \{3\}$. Suppose that at any step $t$, the losses of participants 1 and 2 are independently sampled from the following two functions with the same probability:
	\begin{align}
		&f_1(w) = \frac{1}{2}(w-\sigma)^2, \quad
		&f_2(w) = \frac{1}{2}(w+\sigma)^2. \nonumber
	\end{align}
	The losses of participants 1 and 2 are i.i.d. and the expected loss of honest participants is
	\begin{align*}
		F_t(w) = \frac{1}{2}(\frac{1}{2}(w-\sigma)^2+\frac{1}{2}(w+\sigma)^2)=\frac{1}{2} (w^2 +\sigma^2).
	\end{align*}
	It is easy to check that these losses satisfy Assumptions \ref{ass-1-new}, \ref{ass-2-new} and \ref{ass-3-new}. To be specific, the overall best solution $w^*=0$, $L=1$, and $\mu=1$.
	
	Take geometric median as an exemplary aggregation rule. Suppose that the algorithm is initialized by $w_1 = \frac{\sigma}{2}$. At step $2$, participant $1$ sends $z_2^1 = w_2^1 = w_1 - \eta_1(w_1-\sigma) = (1-\eta_1)w_1 + \eta_1\sigma$ or $z_2^1 = w_2^1 = w_1 - \eta_1(w_1+\sigma) = (1-\eta_1)w_1 - \eta_1\sigma$, each with a $50\%$ probability. Participant $2$ sends $z_2^2$ whose distribution is the same as that of $z_2^1$. In this circumstance, participant $3$, who is Byzantine, can send $z_2^3 = (1-\eta_1)w_1 + \eta_1\sigma$ so that the aggregation result is $w_2 = (1-\eta_1)w_1 + \eta_1\sigma$ with a $75\%$ probability or $w_2 = (1-\eta_1)w_1 - \eta_1\sigma$ with a $25\%$ probability. Thus, the expected aggregation result at step $2$ is $\mathbb{E}w_2 = (1-\eta_1)w_1 + \eta_1\frac{\sigma}{2}$.
	
	At step $3$, participant $1$ sends $z_3^1 = w_3^1 = w_2 - \eta_2(w_2-\sigma) = (1-\eta_2)w_2 + \eta_2\sigma$ or $z_3^1 = w_3^1 = w_2 - \eta_2(w_2+\sigma) = (1-\eta_2)w_2 - \eta_2\sigma$, each with a $50\%$ probability. Participant $2$ sends $z_3^2$ whose distribution is the same as that of $z_3^1$. In this circumstance, participant $3$, who is Byzantine, can send $z_3^3 = (1-\eta_2)w_2 + \eta_2\sigma$ so that the aggregation result is $w_3 = (1-\eta_2)w_2 + \eta_2\sigma$ with a $75\%$ probability or $w_3 = (1-\eta_2)w_2 - \eta_2\sigma$ with a $25\%$ probability. Thus, the expected aggregation result at step $3$ is $\mathbb{E}w_3 = (1-\eta_2)\mathbb{E}w_2 + \eta_2\frac{\sigma}{2}$.
	
	 As such, for any step $t+1$, we have $\mathbb{E}w_{t+1}=(1-\eta_t)\mathbb{E}w_t + \eta_t\frac{\sigma}{2}$. With the initialization $w_1 = \frac{\sigma}{2}$, for any step $t$, we get $\mathbb{E}w_t = \frac{\sigma}{2}$, $F_t(w_t) = \frac{1}{2}(w_t^2+\sigma^2)$, $F_t(w^*) = \frac{1}{2}\sigma^2$, and the stochastic regret is at least
	 \begin{align*}
	 \mathbb{E}\sum_{t=1}^{T}(F_t(w_t)-F_t(w^*)) &= \frac{1}{2}\mathbb{E}\sum_{t=1}^{T}w_t^2\\
	 &\geq \frac{1}{2}\sum_{t=1}^{T}(\mathbb{E}w_t)^2
	 =\frac{1}{8}\sigma^2 T.
	 \end{align*}
	 	
	For other robust bounded aggregation rules, we can observe that the expected mean of the honest messages is $\mathbb{E} \bar{z}_{t+1} = (1-\eta_t) \frac{\sigma}{2}$ and
the largest deviation is $\zeta^2 = \eta_t^2\sigma^2$. According to Definition \ref{def:robust-bounded}, participant 3 can always manipulate its message so that the expected aggregation result is in the order of $\sigma$, which eventually yields linear stochastic regret. If the aggregation rule is majority-voting-based, such as coordinate-wise median and trimmed mean, sending $z_{t+1}^3 = (1-\eta_t)w_t + \eta_t\sigma$ is effective. For centered clipping, participant 3 can send $z_{t+1}^3 = (1-\eta_t)w_t + 3\eta_t\sigma$ if $z_{t+1}^1 \neq z_{t+1}^2$ or $z_{t+1}^3=z_{t+1}^1$ if $z_{t+1}^1 = z_{t+1}^2$ instead.
\end{example}

\begin{remark}
When the environment provides all the honest participants with independent losses from the same distribution $\mathcal{D}$, the online learning and stochastic optimization formulations share similarities. However, they are from different perspectives, one is sequentially making decisions against a possibly adversarial environment with the objective of minimizing the regret, while another is actively sampling losses to approach the minimizer of the expected loss. In addition, from the online learning perspective, we can adopt different performance metrics, such as dynamic regret when the underlying distribution is time-varying \cite{zinkevich2003online}. Our results can also be extended to new online learning algorithms, such as online conditional gradient \cite{garber2016linearly}, online mirror descent \cite{hall2015online}, adaptive gradient \cite{ding2015adaptive}, etc.
\end{remark}

\section{Numerical Experiments}
\label{Numerical Experiments}

In this section, we show the performance of the Byzantine-robust distributed online gradient descent and momentum algorithms through numerical experiments, including least-squares regression on synthetic datasets, softmax regression on the MNIST dataset and Resnet18 training on the CIFAR10 dataset. Due to the page limit, we left Resnet18 training on the CIFAR10 dataset to Appendix \ref{experimentCifar10}. The source code is available online\footnote{\url{https://github.com/wanger521/OGD}}.
	
In addition to the non-robust mean aggregation rule, we test seven robust bounded aggregation rules, including coordinate-wise median, trimmed mean, geometric median, Krum, centered clipping, Phocas, and FABA. We consider the following three commonly-used Byzantine attacks.

\noindent \textbf{Sign-flipping attack.} Each Byzantine participant sends a negative multiple of its true message, and the coefficient is set as $-3$, $-1$ and $-1$ for the three numerical experiments, respectively.

\noindent \textbf{Gaussian attack.} Each Byzantine participant sends a random message, where each element follows the Gaussian distribution $\mathcal{N}(0,500)$, $\mathcal{N}(0,200)$ and $\mathcal{N}(0,200)$ for the three numerical experiments, respectively.

\noindent \textbf{Sample-duplicating attack.} The Byzantine participants jointly choose one honest participant, and duplicate its message to send. This amounts to that the Byzantine participants duplicate the samples of the chosen honest participant.

\subsection{Least-squares regression on synthetic datasets}
We start with least-squares regression on synthetic datasets, each of which contains 60,000 training samples. The dimensionality of decision variable is $d=10$. During training, the batch size is 1. We launch one server and 30 participants. Under Byzantine attacks, $5$ randomly chosen participants are adversarial.

We take into account two data distributions. In the i.i.d. setting, each element of the regressors is drawn from the Gaussian distribution $\mathcal{N}(0,1)$. We also randomly generate each dimension of the ground-truth solution from the Gaussian distribution $\mathcal{N}(0,1)$. Then, the labels are obtained via multiplying the regressors by the ground-truth solution, followed by adding Gaussian noise $\mathcal{N}(0,0.1)$. These training samples are evenly distributed to all participants. In the non-i.i.d. setting, each element of the regressors and the ground-truth solution is evenly drawn from three pairs of Gaussian distributions:
$(\mathcal{N}(0,1), g+\mathcal{N}(0,0.5))$, $(\mathcal{N}(1,1),g+ \mathcal{N}(0.2,0.5))$, and $(\mathcal{N}(2,1), g+\mathcal{N}(0.4,0.5))$, where $g \sim \mathcal{N}(0,1)$.
The added Gaussian noise is still from $\mathcal{N}(0,0.1)$. For each of the three classes, the training samples are evenly distributed to 10 participants.

The performance metrics are adversarial regret and stochastic regret for the i.i.d. setting, and adversarial regret for the non-i.i.d. setting. We repeat generating the datasets and conducing the experiments for 10 times to calculate the regrets. This way, taking the average approximates the stochastic regret bound, while choosing the worst approximates the adversarial regret bound.

When the step size $\eta$ and the momentum parameter $\nu$ are constant, they are set to 0.01 for the i.i.d. setting and 0.005 for the non-i.i.d. setting. For the diminishing step size $\eta_t$ and momentum parameter $\nu_t$, they are set to 0.008 in the first 500 iterations, and $\frac{4}{t}$ afterwards.

\begin{figure*}
 	\begin{center}
 		\centerline{\includegraphics[width=0.8\textwidth]{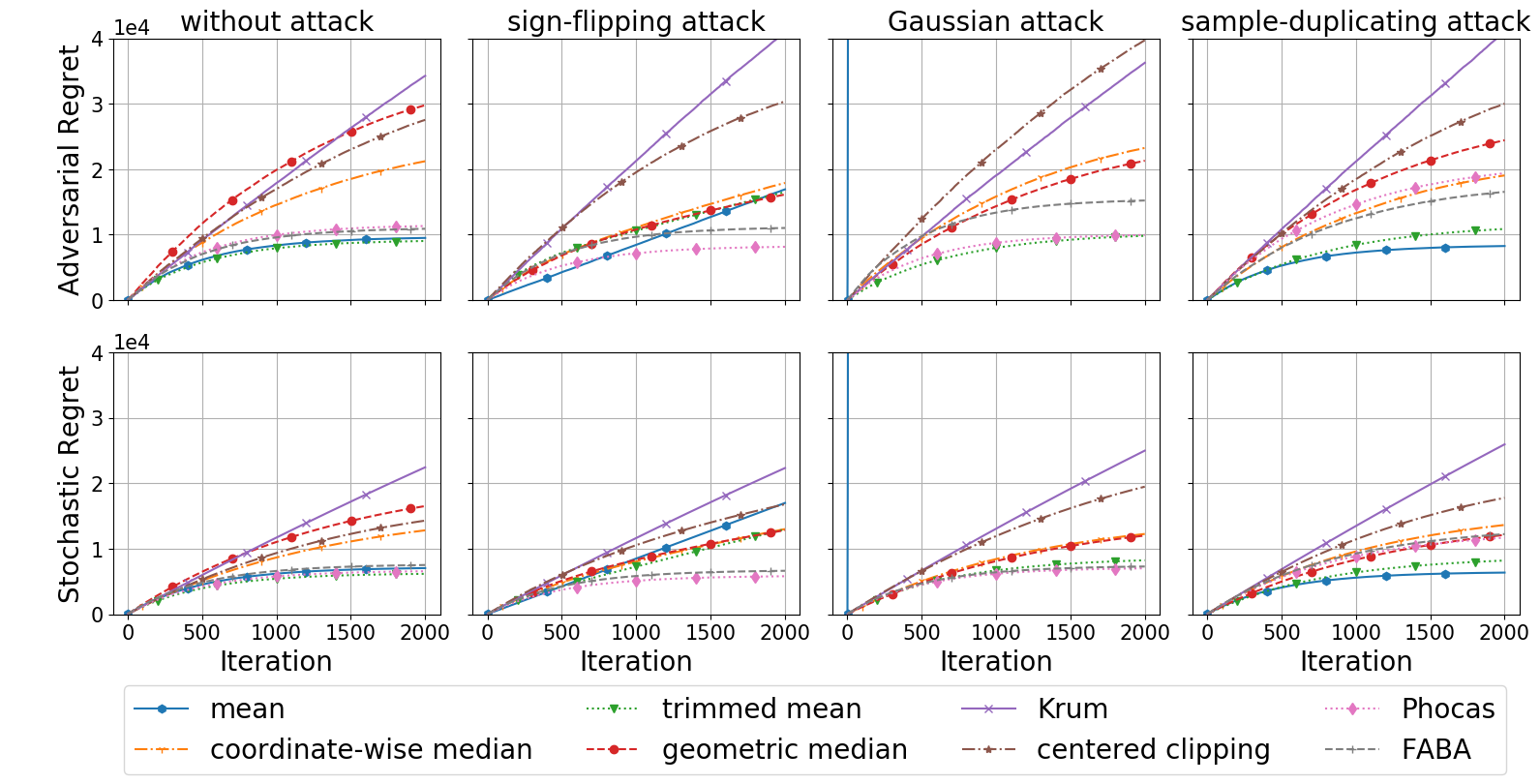}}
 		\caption{Byzantine-robust distributed online gradient descent for least-squares regression on synthetic i.i.d. data with constant step size.}
 		\label{fig:iid-constant-DROGD-artificial}
 	\end{center}
 \vspace{-2em}
 \end{figure*}

\begin{figure*}
	\begin{center}
		\centerline{\includegraphics[width=0.8\textwidth]{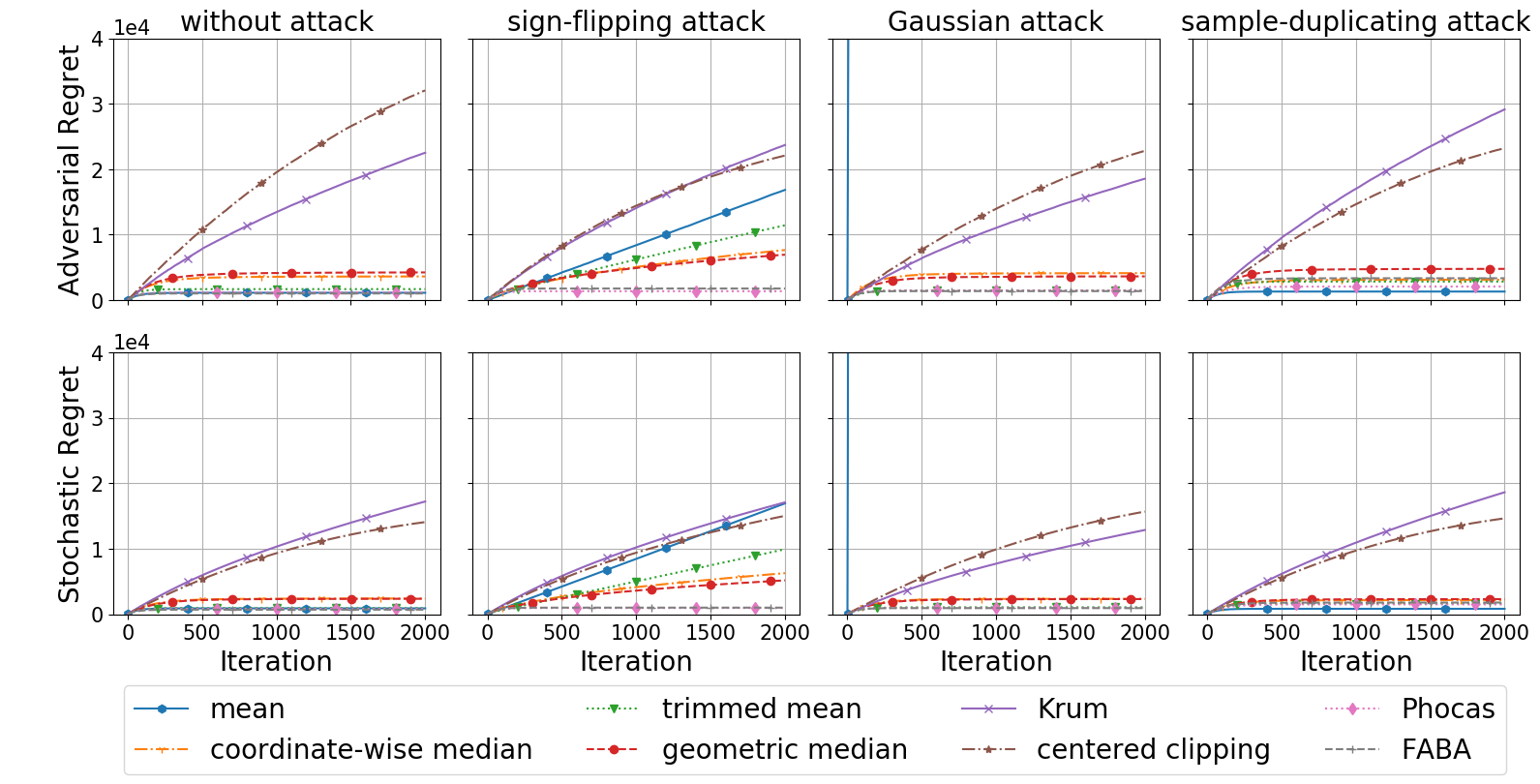}}
		\caption{Byzantine-robust distributed online gradient descent  for least-squares regression on synthetic i.i.d. data with diminishing step size.}
		\label{fig:iid-diminishing-DROGD-artificial}
	\end{center}
	\vspace{-2em}
\end{figure*}

\begin{figure*}
	\begin{center}
		\centerline{\includegraphics[width=0.8\textwidth]{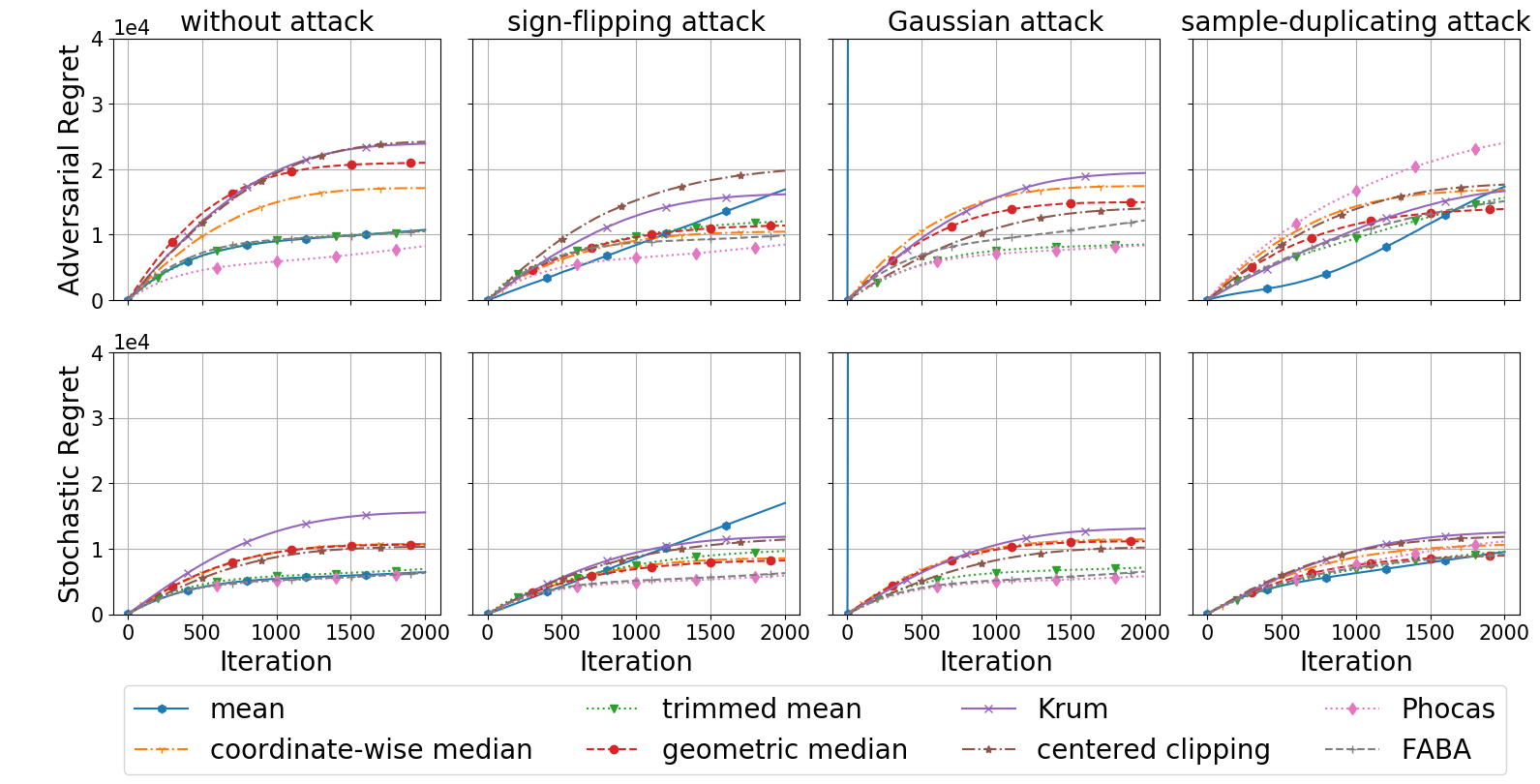}}
		\caption{Byzantine-robust distributed online momentum for least-squares regression on synthetic i.i.d. data with constant step size.}
		\label{fig:iid-constant-DROGDM-artificial}
	\end{center}
\vspace{-2em}
\end{figure*}

\begin{figure*}
	\begin{center}
		\centerline{\includegraphics[width=0.8\textwidth]{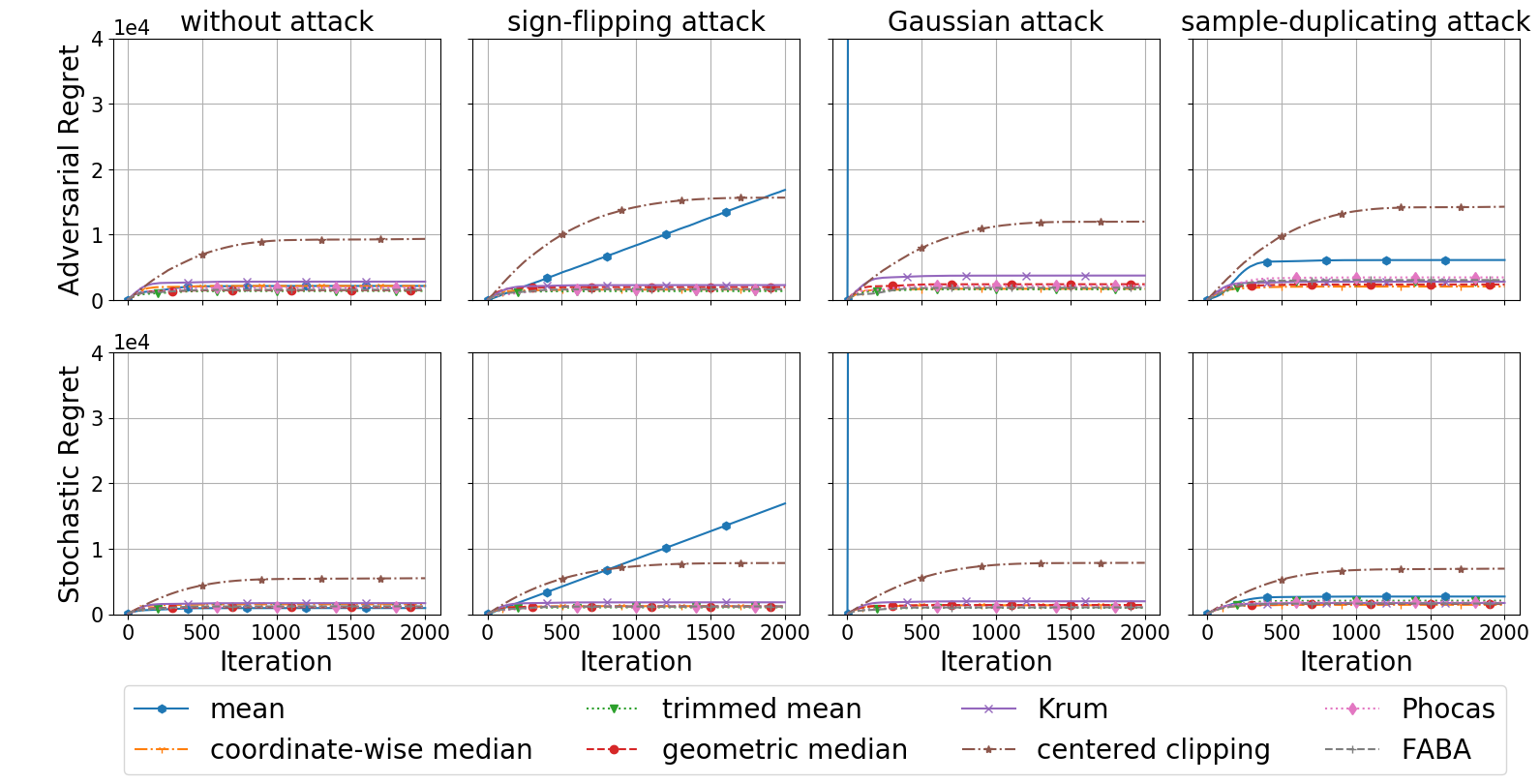}}
		\caption{Byzantine-robust distributed online momentum for least-squares regression on synthetic i.i.d. data with diminishing step size.}
		\label{fig:iid-diminishing-DROGDM-artificial}
	\end{center}
\vspace{-2em}
\end{figure*}

\noindent \textbf{Numerical experiments on i.i.d. data.}
As shown in Figs. \ref{fig:iid-constant-DROGD-artificial} and \ref{fig:iid-diminishing-DROGD-artificial}, the Byzantine-robust distributed online gradient descent algorithms equipped with robust bounded aggregation rules demonstrate trends of linear regret bounds, no matter using constant or diminishing step size. Take Fig. \ref{fig:iid-diminishing-DROGD-artificial} as an example. Although trimmed mean, Phocas, coordinate-wise median, geometric median and FABA show sublinear regret bounds under the Gaussian and sample-duplicating attacks, they yield to linear regret bounds under the sign-flipping attack. This validates the tightness of Theorem \ref{theorem:bound-T} even on i.i.d. data. The Byzantine-robust distributed online momentum algorithms significantly improves the regret bounds, as shown in Figs. \ref{fig:iid-constant-DROGDM-artificial} and \ref{fig:iid-diminishing-DROGDM-artificial}. Their regret bounds are all sublinear, which corroborate with Theorem \ref{theorem-DROGDM}.

\begin{figure*}
	\begin{center}
		\centerline{\includegraphics[width=0.8\textwidth]{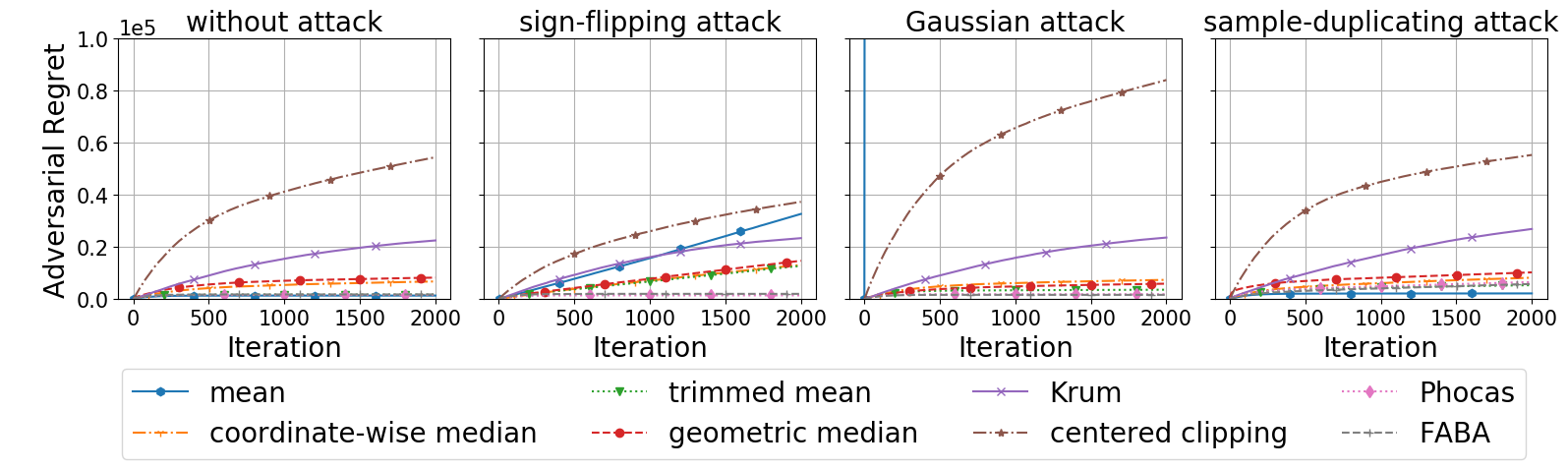}}
		\caption{Byzantine-robust distributed online gradient descent for least-squares regression on synthetic non-i.i.d. data with constant step size.}
		\label{fig:noniid-constant-DROGD-artificial}
	\end{center}
\vspace{-2em}
\end{figure*}

\begin{figure*}
	\begin{center}
		\centerline{\includegraphics[width=0.8\textwidth]{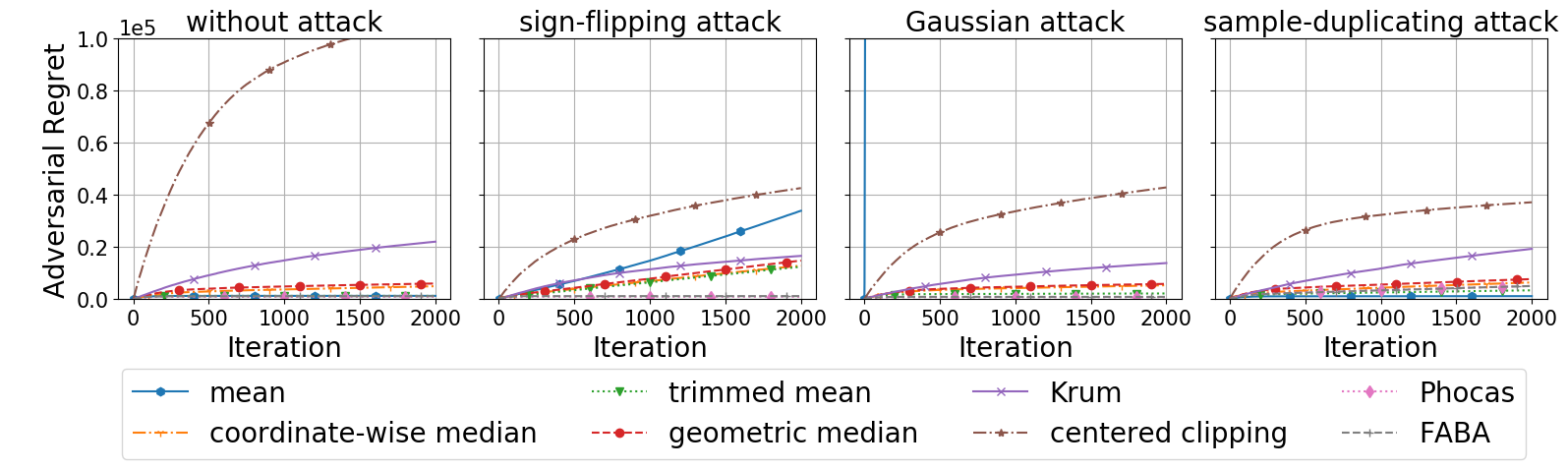}}
		\caption{Byzantine-robust distributed online gradient descent for least-squares regression on synthetic non-i.i.d. data with diminishing step size.}
		\label{fig:noniid-diminishing-DROGD-artificial}
	\end{center}
	\vspace{-2em}
\end{figure*}

\begin{figure*}
	\begin{center}
		\centerline{\includegraphics[width=0.8\textwidth]{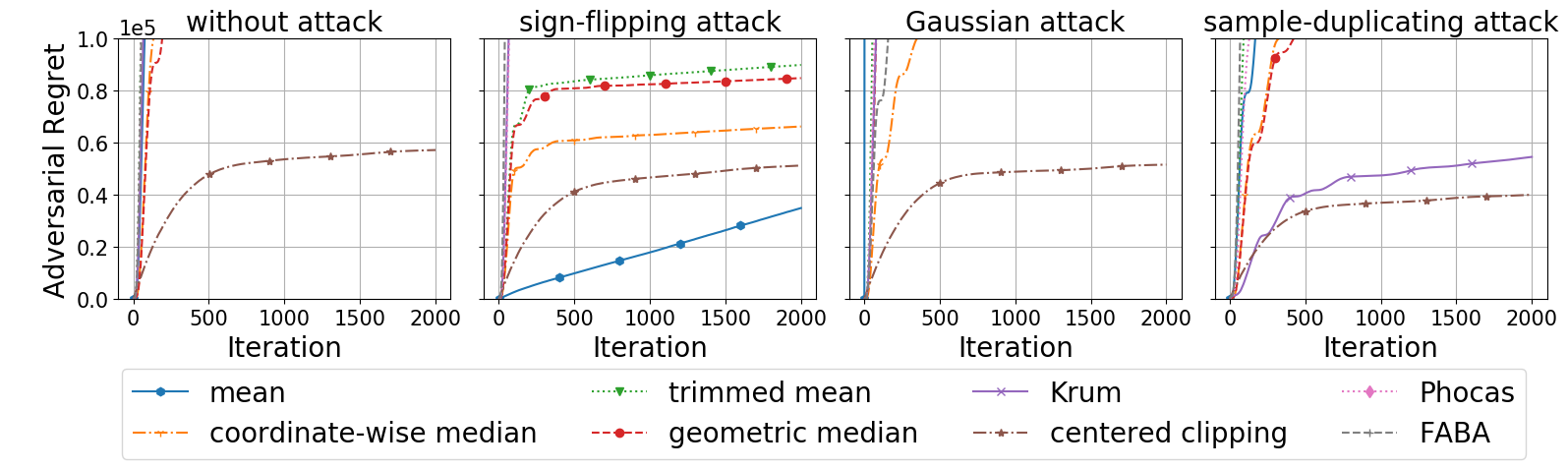}}
		\caption{Byzantine-robust distributed online momentum for least-squares regression on synthetic non-i.i.d. data with constant step size.}
		\label{fig:noniid-constant-DROGDM-artificial}
	\end{center}
\vspace{-2em}
\end{figure*}

\begin{figure*}
	\begin{center}
		\centerline{\includegraphics[width=0.8\textwidth]{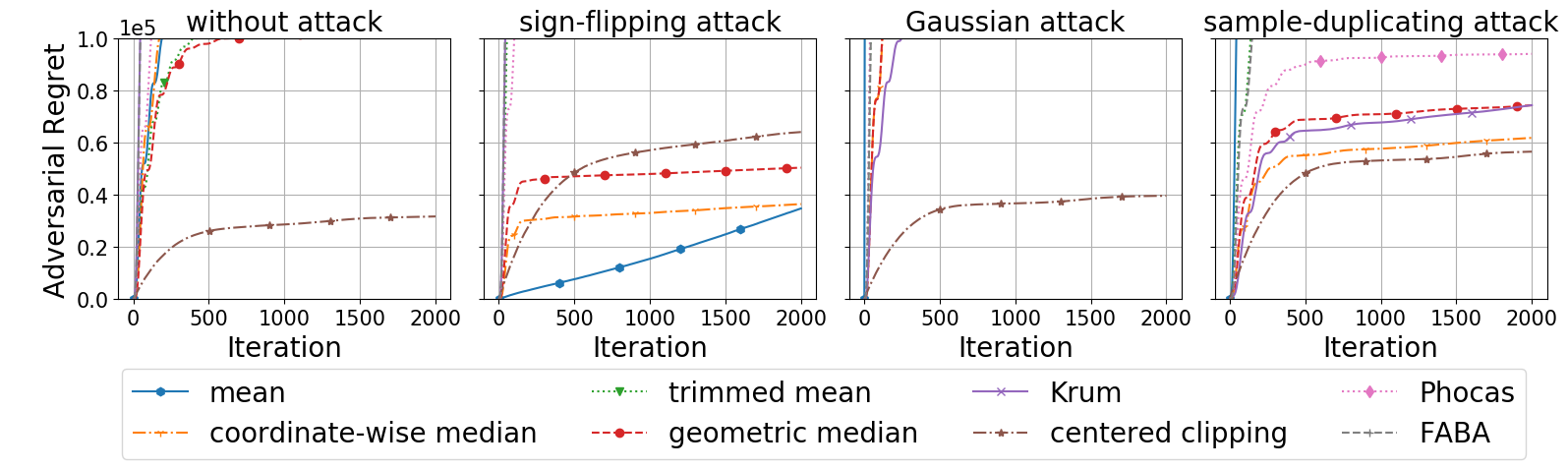}}
		\caption{Byzantine-robust distributed online momentum for least-squares regression on synthetic non-i.i.d. data with diminishing step size.}
		\label{fig:noniid-diminishing-DROGDM-artificial}
	\end{center}
\vspace{-2em}
\end{figure*}

\noindent \textbf{Numerical experiments on non-i.i.d. data.}
On the non-i.i.d. data, the environment is more adversarial than that on the i.i.d. data. As shown in Figs. \ref{fig:noniid-constant-DROGD-artificial} and \ref{fig:noniid-diminishing-DROGD-artificial}, the Byzantine-robust distributed online gradient descent algorithms, whether under attack or not, exhibit linear adversarial regret bounds. The Byzantine-robust distributed online momentum algorithms, as shown in Figs. \ref{fig:noniid-constant-DROGDM-artificial} and \ref{fig:noniid-diminishing-DROGDM-artificial}, may have even larger regrets than those without momentum. This phenomenon underscores the importance of i.i.d. data distribution to Byzantine-robustness.

\subsection{Softmax regression on the MNIST dataset}
We next consider softmax regression on the MNIST dataset, which contains 60,000 training samples and 10,000 testing samples. The batch size is set to 32 during training. We launch one server and 30 participants, and consider two data distributions. In the i.i.d. setting, all the training samples are randomly and evenly allocated to all participants. In the non-i.i.d. setting, each class of the training samples are randomly and evenly distributed to 3 participants. Under Byzantine attacks, 5 randomly chosen participants are adversarial.

\begin{figure*}
	\begin{center}
		\centerline{\includegraphics[width=0.8\textwidth]{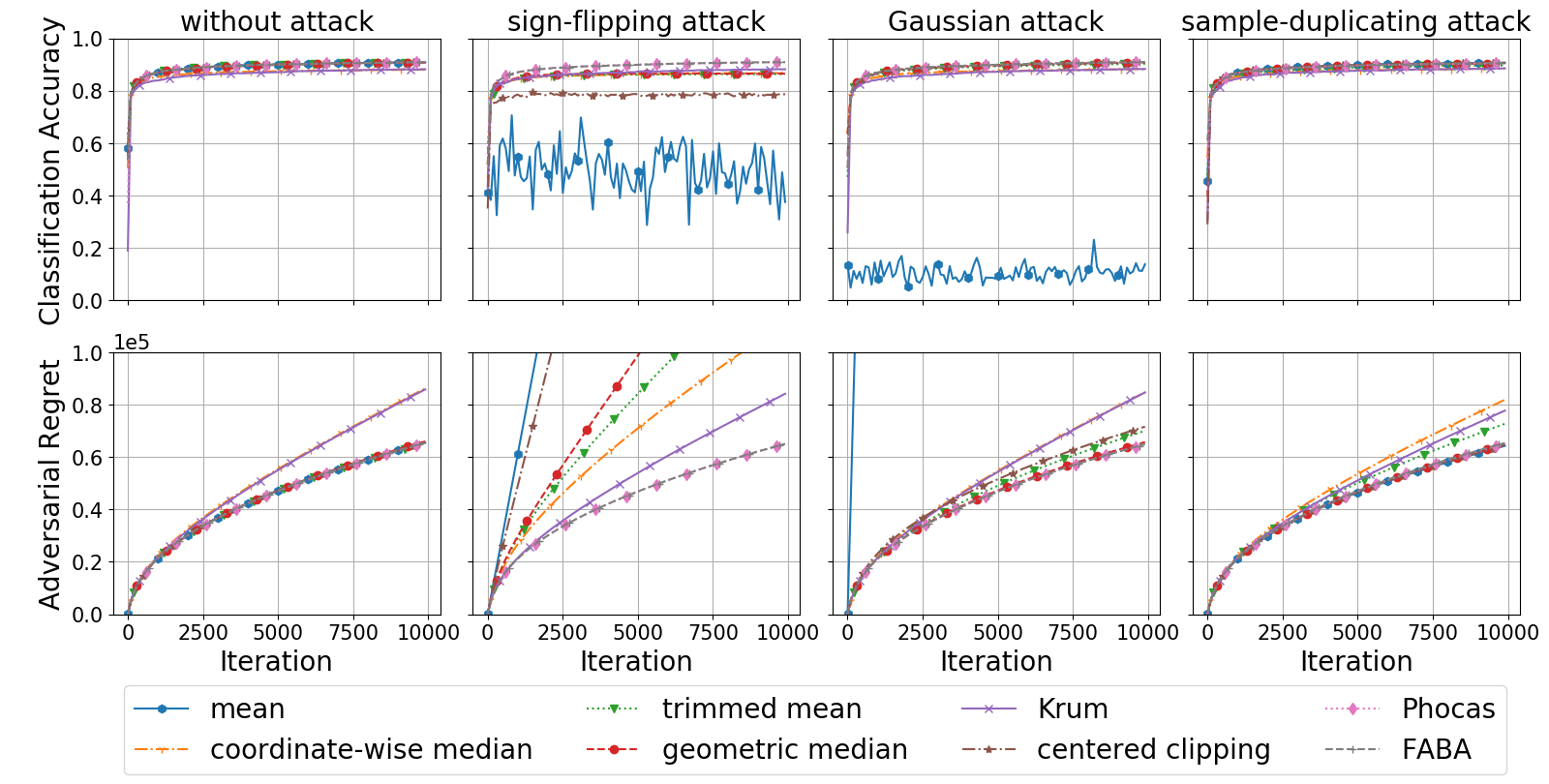}}
		\caption{Byzantine-robust distributed online gradient descent for softmax regression on MNIST i.i.d. data with constant step size.}
		\label{fig:iid-sqrtT-DROGD}
	\end{center}
	\vspace{-2em}
\end{figure*}
 \begin{figure*}
	\begin{center}
		\centerline{\includegraphics[width=0.8\textwidth]{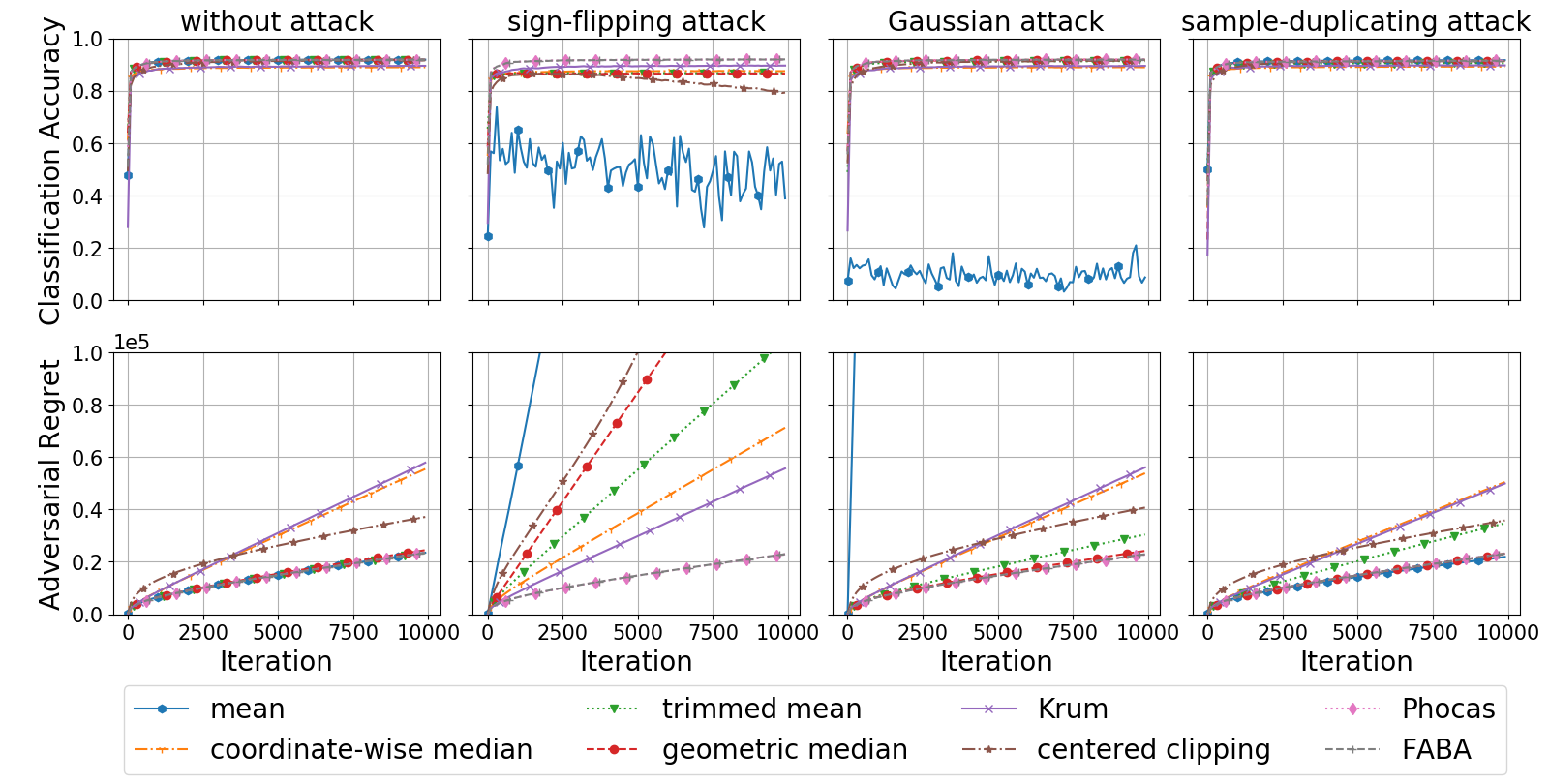}}
		\caption{Byzantine-robust distributed online gradient descent for softmax regression on MNIST i.i.d. data with diminishing step size.}
		\label{fig:iid-diminishing-DROGD-mnist}
	\end{center}
	\vspace{-2em}
\end{figure*}
\begin{figure*}
	\begin{center}
		\centerline{\includegraphics[width=0.8\textwidth]{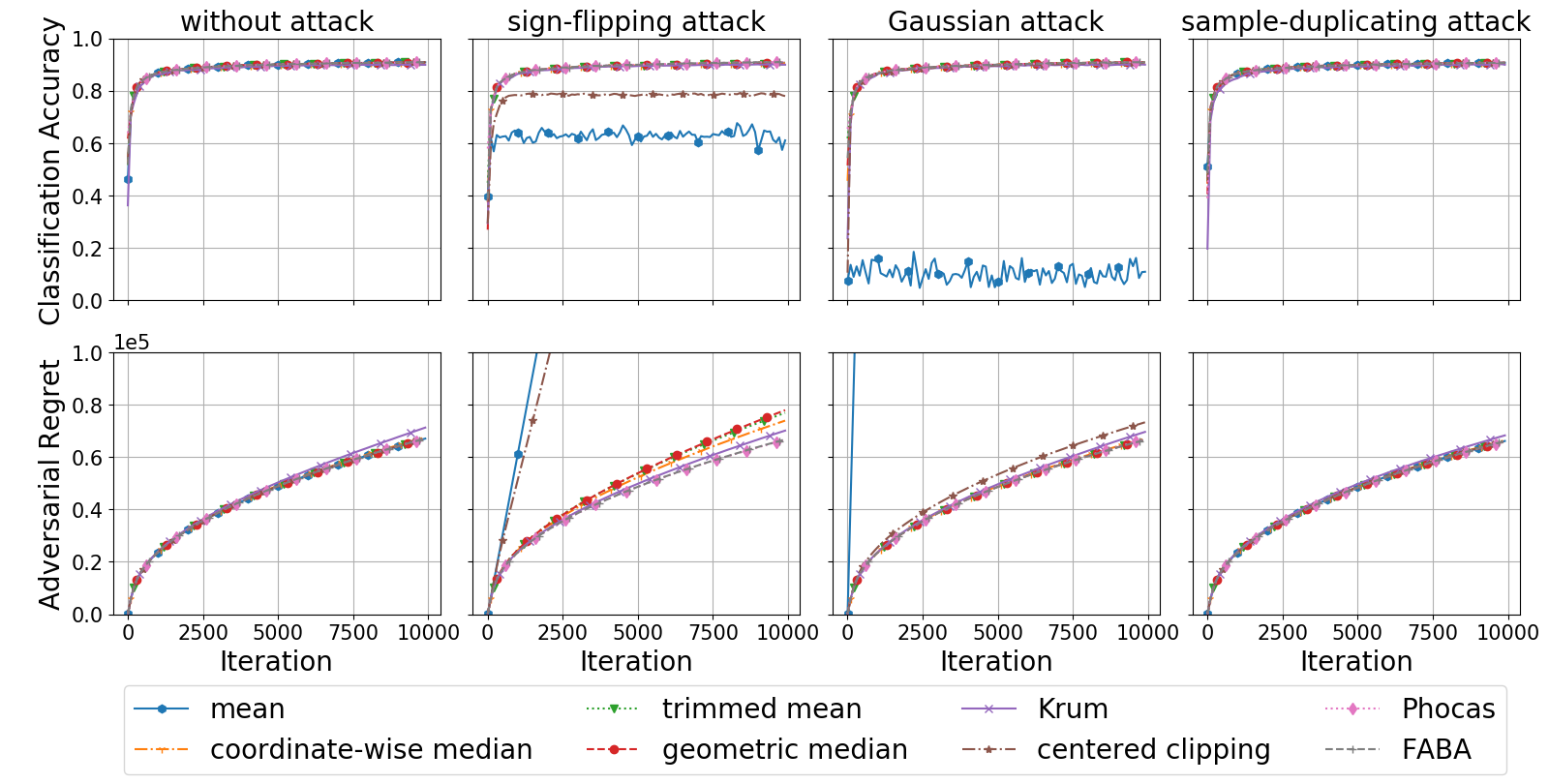}}
		\caption{Byzantine-robust distributed online momentum for softmax regression on MNIST i.i.d. data with constant step size.}
		\label{fig:iid-sqrtT-momentum-DROGDM}
	\end{center}
	\vspace{-2em}
\end{figure*}
 \begin{figure*}
	\begin{center}
		\centerline{\includegraphics[width=0.8\textwidth]{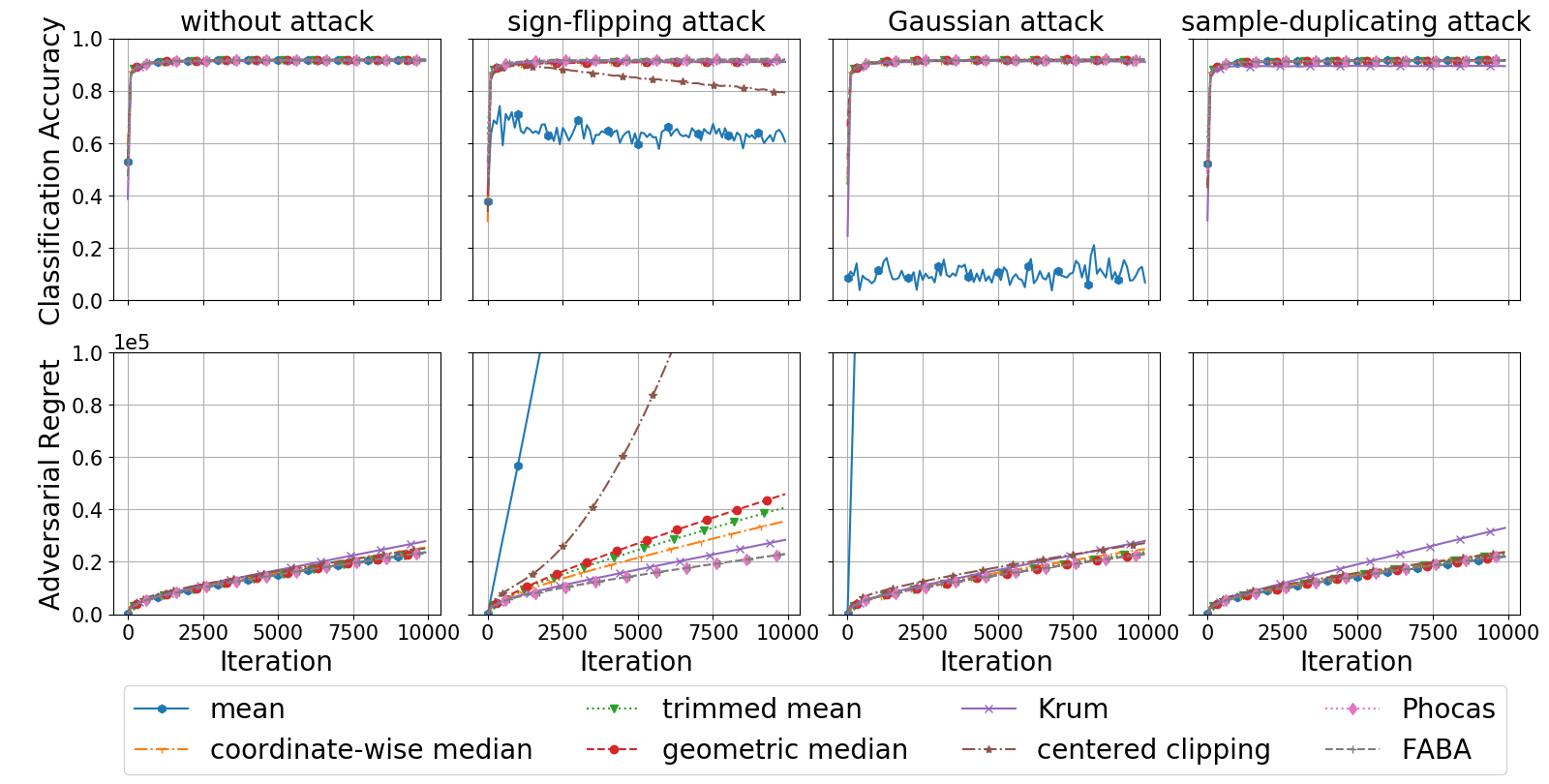}}
		\caption{Byzantine-robust distributed online momentum for softmax regression on MNIST i.i.d. data with diminishing step size.}
		\label{fig:iid-diminishing-DROGDM-mnist}
	\end{center}
	\vspace{-2em}
\end{figure*}
\begin{figure*}
	\begin{center}
		\centerline{\includegraphics[width=0.8\textwidth]{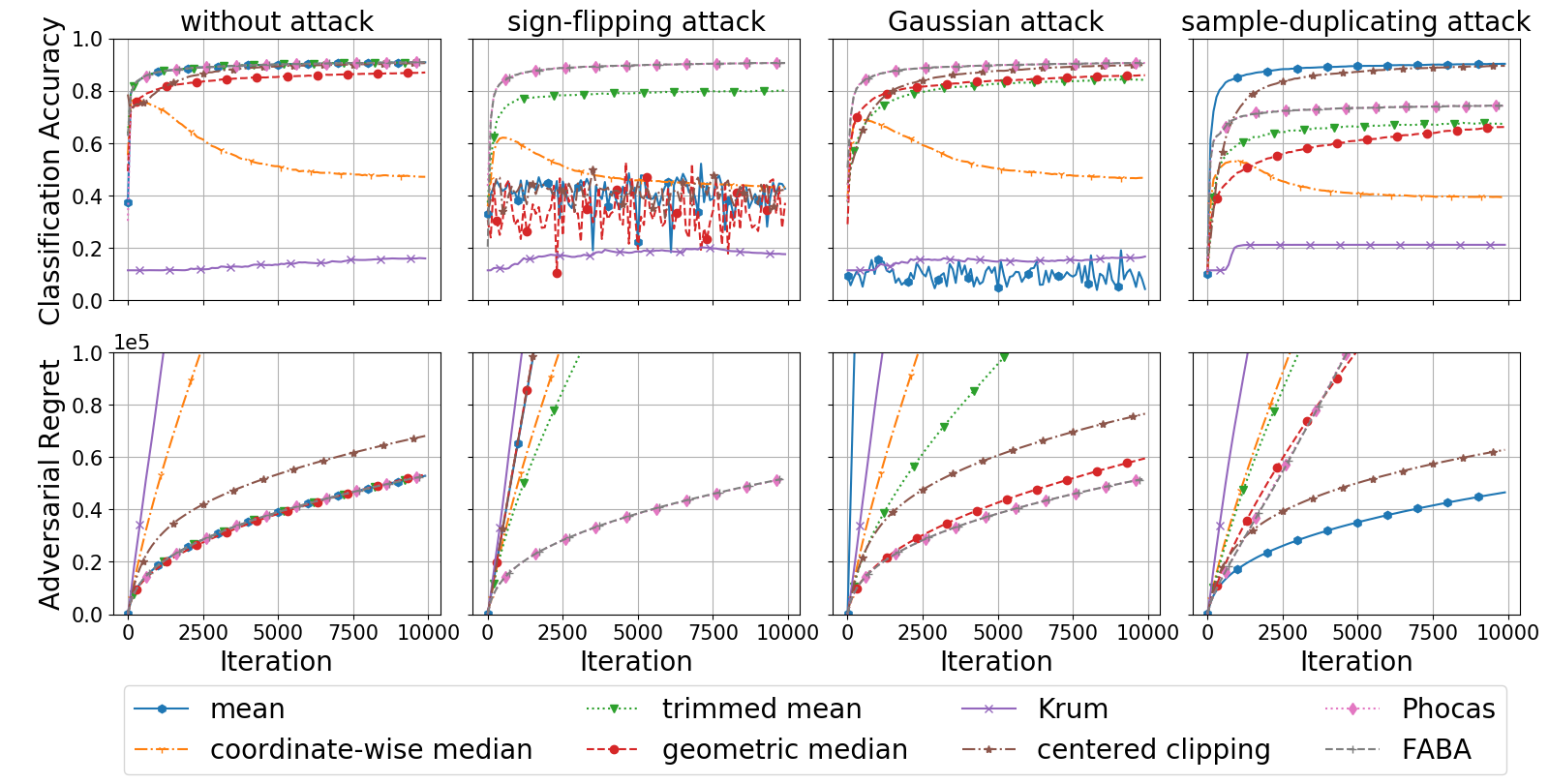}}
		\caption{Byzantine-robust distributed online gradient descent for softmax regression on MNIST non-i.i.d. data with constant step size.}
		\label{fig:noniid-sqrtT-DROGD}
	\end{center}
	\vspace{-2em}
\end{figure*}
 \begin{figure*}
	\begin{center}
		\centerline{\includegraphics[width=0.8\textwidth]{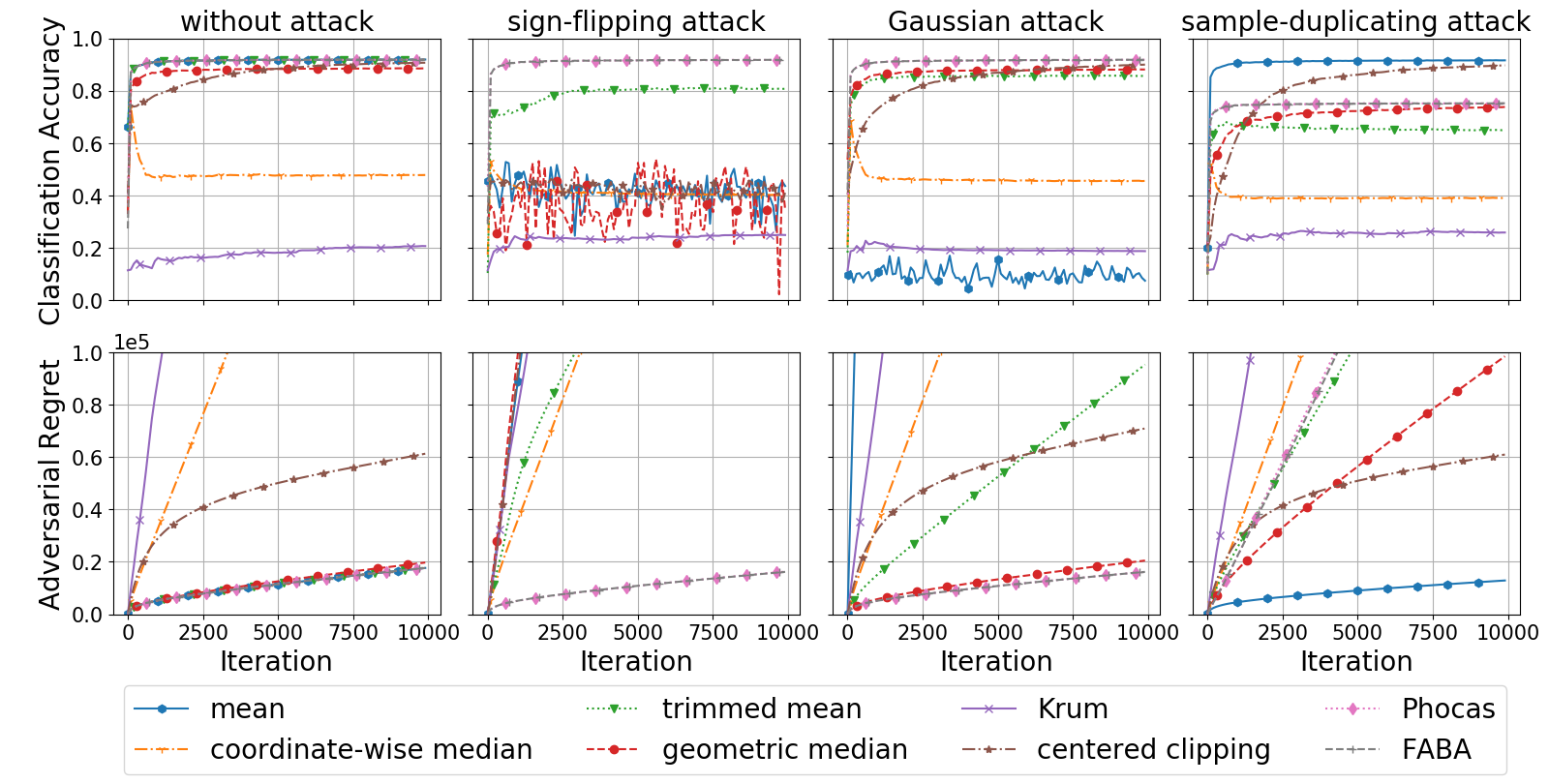}}
		\caption{Byzantine-robust distributed online gradient descent for softmax regression on MNIST non-i.i.d. data with diminishing step size.}
		\label{fig:noniid-diminishing-DROGD-mnist}
	\end{center}
	\vspace{-2em}
\end{figure*}
\begin{figure*}
	\begin{center}
		\centerline{\includegraphics[width=0.8\textwidth]{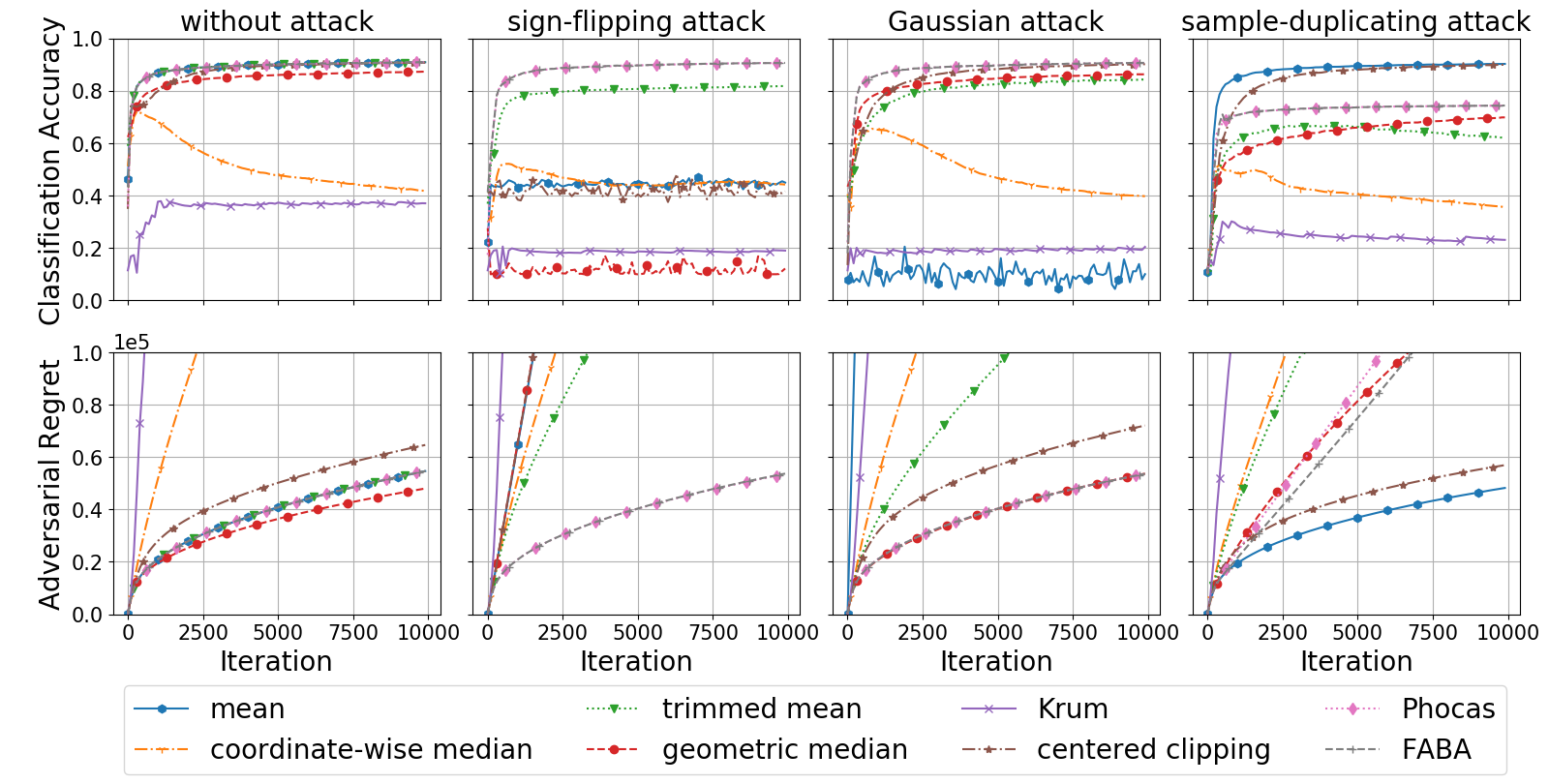}}
		\caption{Byzantine-robust distributed online momentum for softmax regression on MNIST non-i.i.d. data with constant step size.}
		\label{fig:noniid-sqrtT-momentum-DROGDM}
	\end{center}
	\vspace{-2em}
\end{figure*}
\begin{figure*}
	\begin{center}
		\centerline{\includegraphics[width=0.8\textwidth]{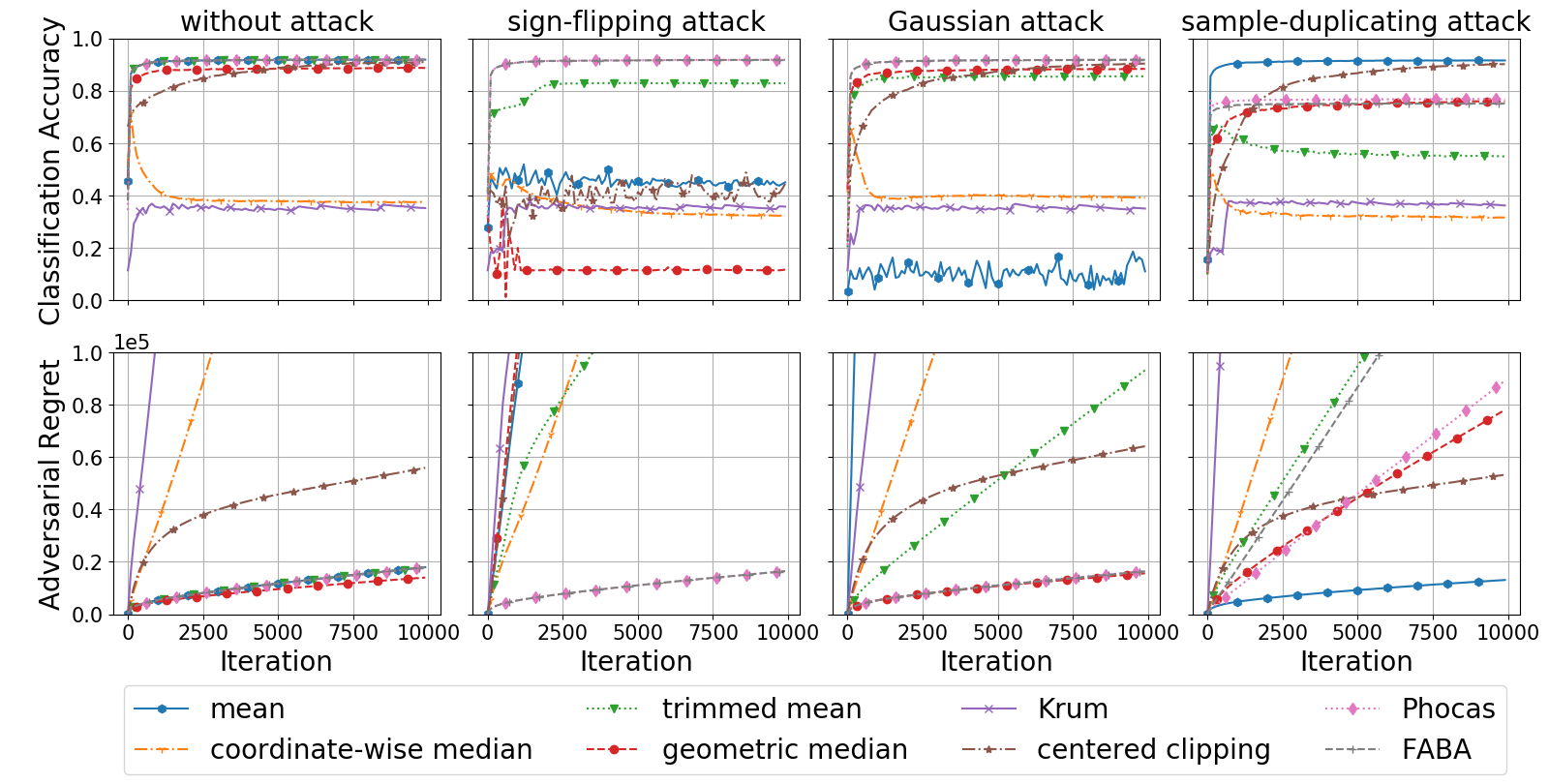}}
		\caption{Byzantine-robust distributed online momentum for softmax regression on MNIST non-i.i.d. data with diminishing step size.}
		\label{fig:noniid-diminishing-DROGDM-mnist}
	\end{center}
	\vspace{-1em}
\end{figure*}

The performance metrics are classification accuracy on the testing samples and adversarial regret on the training samples. Since accurately calculating the adversarial and stochastic regret bounds is computationally demanding on such a large dataset, we only conduct the numerical experiments once, and calculate the adversarial regret to approximate its bound.  Note that in the i.i.d. setting, adversarial regret is an approximation of stochastic regret, but there is still a substantial gap between the two.

When the step size $\eta$ is constant, it is set to 0.01 and the momentum parameter $\nu$ is also set to 0.01. For the diminishing step size $\eta_t$ and momentum parameter $\nu_t$, they are set to 0.1 in the first 500 steps and $\frac{1}{t}$ afterwards.

\noindent \textbf{Numerical experiments on i.i.d. data.} As shown in Figs. \ref{fig:iid-sqrtT-DROGD} and \ref{fig:iid-diminishing-DROGD-mnist}, on the i.i.d. data, Byzantine-robust distributed online gradient descent equipped with robust bounded aggregation rules all perform well when no attack presents or under the sample-duplicating attack. Under the sign-flipping and Gaussian attacks, the algorithm with mean aggregation fails, and the others demonstrate satisfactory robustness. The sign-flipping attack turns to be slightly stronger than the Gaussian attack; under the former, the algorithm with centered clipping performs worse, but is still much better than the one with mean aggregation.

The Byzantine-robust distributed online gradient descent algorithms with momentum improve over the ones without momentum in terms of classification accuracy and adversarial regret, as shown in Figs. \ref{fig:iid-sqrtT-momentum-DROGDM} and \ref{fig:iid-diminishing-DROGDM-mnist}. However, no sublinear adversarial regret bound is guaranteed, which confirms our theoretical prediction.

\noindent \textbf{Numerical experiments on non-i.i.d. data.} On the non-i.i.d. data, the environment is more adversarial than on the i.i.d. data. In this case, Byzantine-robust distributed online gradient descent, no matter with or without momentum, does not perform well, as in Figs. \ref{fig:noniid-sqrtT-DROGD}, \ref{fig:noniid-diminishing-DROGD-mnist}, \ref{fig:noniid-sqrtT-momentum-DROGDM}, and \ref{fig:noniid-diminishing-DROGDM-mnist}. This observation matches our conclusion on the hardness of handling adversarial participants in the adversarial environment.

%


\section{Conclusions}
\label{Conclusions}

This paper is among the first efforts to investigate the Byzantine-robustness of distributed online learning. We show that Byzantine-robust distributed online gradient descent has linear adversarial regret, and the constant of the linear term is determined by the robust aggregation rule. On the other hand, we also establish the sublinear stochastic regret bound for Byzantine-robust distributed online momentum under the i.i.d. assumption.

Our future focus is to improve the Byzantine-robustness of distributed online learning algorithms in the non-i.i.d. setting, which is of practical importance in processing streaming data.


\bibliography{refs}
\bibliographystyle{IEEEtran}

%
%

%
%


\clearpage
\appendices

\section{Proof of Theorem \ref{theorem:bound-T}}
\label{theorem-T}
In this section, we establish the adversarial regret bound of Byzantine-robust distributed online gradient descent under Assumptions \ref{ass-1}, \ref{ass-2}, \ref{ass-3} and \ref{ass-4}.
We start from two supporting lemmas whose proofs are in Appendix \ref{lemma-proof}.

First, Lemma \ref{mean-bound} shows that the average gradient of the honest participants is upper bounded.
\begin{lemma}
Under Assumptions \ref{ass-1}, \ref{ass-2} and \ref{ass-3}, for any step $t \in [1,T]$, the average gradient of the honest participants $\nabla f_t(w_t):=\|\frac{1}{h}\sum_{j\in \mathcal{H}} \nabla{f_t^j(w_t)}\|^2$ is bounded by
	\begin{align}
		\label{mean-bound-eq}
		 \|\nabla f_t(w_t)\|^2
		\leq &
		4L(f_t(w_t)-f_t(w^*)) \\
		 \nonumber & + \frac{2L}{\rho_t}\|w_t-w^*\|^2 + 2(1+L\rho_t)\xi^2,
	\end{align}
	where $w^* = \arg \min_{w\in \mathbb{R}^d}\sum_{t=1}^{T}f_t(w)$ is the overall best solution and $\rho_t$ is a positive time-varying parameter.
	\label{mean-bound}
\end{lemma}

Second, Lemma \ref{recursion} characterizes the recursion of $\|w_t - w^*\|^2$.

\begin{lemma}
	Under Assumptions \ref{ass-1}, \ref{ass-2}, \ref{ass-3} and \ref{ass-4}, for any step $t \in [1,T]$, the distance between the decision $w_{t+1}$ and the overall best solution $w^* := \arg \min_{w\in \mathbb{R}^d}\sum_{t=1}^{T}f_t(w)$ follows
	\begin{align}
\label{eq-lemma-2}
		 &\|w_{t+1}-w^*\|^2\\
		\nonumber \leq&
		(1+\hat{\rho}_t)(1+\frac{2L\eta_t^2}{\rho_t}- \eta_t\mu)\|w_t - w^*\|^2 \\
	\nonumber &+ (1+\hat{\rho}_t)(4L\eta_t^2-2 \eta_t)(f_t(w_t)-f_t(w^*))\\
	\nonumber &+ (1+\hat{\rho}_t)2(1+L\rho_t)\eta_t^2\xi^2
	+(1+\frac{1}{\hat{\rho}_t})\eta_t^2C_{\alpha}^2\sigma^2,
		\end{align}
	where $\rho_{t}$ and $\hat{\rho_t}$ are positive time-varying parameters.
	\label{recursion}
\end{lemma}

Now we prove Theorem \ref{theorem:bound-T}. Rewrite \eqref{eq-lemma-2} of Lemma \ref{recursion} as	
\begin{align}
	\label {1-rewrite-lemma}
	 &(1+\hat{\rho}_t)\eta_t(2 - 4L\eta_t)(f_t(w_t)-f_t(w^*))\\
	\nonumber \leq &\|w_t - w^*\|^2 - \|w_{t+1}-w^*\|^2 \\
	\nonumber  &+(\hat{\rho}_t+(1+\hat{\rho}_t)(\frac{2L\eta_t^2}{\rho_t}- \eta_t\mu))\|w_t - w^*\|^2  \\
	\nonumber &+(1+\hat{\rho}_t)2(1+L\rho_t)\eta_t^2\xi^2
+(1+\frac{1}{\hat{\rho}_t})\eta_t^2C_{\alpha}^2\sigma^2.
\end{align}
For any step $t \in[1,T]$, let $\eta_t \in (0, \frac{1}{4L})$ and $\hat{\eta}_t:= \frac{1}{2- 4L\eta_t}$. It holds that $\frac{1}{2}<\hat{\eta}_t\leq{1}$ and $\frac{\eta_t}{\hat{\eta}_t}=\eta_t(2- 4L\eta_t) > 0$. Dividing both sides of (\ref{1-rewrite-lemma}) by $\frac{(1+\hat{\rho}_t)\eta_t}{\hat{\eta}_t}$, we have
\begin{align}
	\label {3-rewrite-lemma}
	  &f_t(w_t)-f_t(w^*)\\
	 \nonumber\leq& \frac{\hat{\eta}_t}{(1+\hat{\rho}_t)\eta_t}(\|w_t - w^*\|^2 - \|w_{t+1}-w^*\|^2)\\
	\nonumber  & + \hat{\eta}_t(\frac{\hat{\rho}_t}{(1+\hat{\rho}_t)\eta_t}+\frac{2L\eta_t}{\rho_t}- \mu)\|w_t - w^*\|^2 \\
	\nonumber &+\hat{\eta}_t2(1+L\rho_t)\eta_t\xi^2
+\hat{\eta}_t\frac{1}{(1+\hat{\rho}_t)}(1+\frac{1}{\hat{\rho}_t})\eta_t C_{\alpha}^2\sigma^2.
\end{align}

Summing up (\ref{3-rewrite-lemma}) from $t=1$ to $T$, we have
\begin{align}
	 &\sum_{t=1}^T(f_t(w_t)-f_t(w^*))\\
	\nonumber \leq & \sum_{t=1}^T\frac{\hat{\eta}_t}{(1+\hat{\rho}_t)\eta_t}(\|w_t - w^*\|^2 - \|w_{t+1}-w^*\|^2)\\
	\nonumber
	&+\sum_{t=1}^T	\hat{\eta}_t(\frac{\hat{\rho}_t}{(1+\hat{\rho}_t)\eta_t}+\frac{2L\eta_t}{\rho_t}- \mu)\|w_t - w^*\|^2 \\
	&+\sum_{t=1}^T\hat{\eta}_t(2(1+L\rho_t)\eta_t\xi^2
	\nonumber+\frac{1}{(1+\hat{\rho}_t)}(1+\frac{1}{\hat{\rho}_t})\eta_t C_{\alpha}^2\sigma^2).
\end{align}
Since $\|w_{T+1}-w^*\|^2 \geq 0$ and $\hat{\eta}_t \leq 1$, we have
\begin{align}
	\label {4-rewrite-lemma}
	&\sum_{t=1}^T(f_t(w_t)-f_t(w^*))\\
	\nonumber\leq & \underbrace{\sum_{t=1}^T(\frac{\hat{\eta}_t}{(1+\hat{\rho}_t)\eta_t}-\frac{\hat{\eta}_{t-1}}{(1+\hat{\rho}_{t-1})\eta_{t-1}})\|w_t - w^*\|^2}_{U_a}\\
	\nonumber &+\underbrace{\sum_{t=1}^T\hat{\eta}_t(\frac{\hat{\rho}_t}{(1+\hat{\rho}_t)\eta_t}+\frac{2L\eta_t}{\rho_t}- \mu)\|w_t - w^*\|^2}_{U_b} \\
	\nonumber &+\sum_{t=1}^T(2(1+L\rho_t)\eta_t\xi^2
	+\frac{1}{(1+\hat{\rho}_t)}(1+\frac{1}{\hat{\rho}_t})\eta_t C_{\alpha}^2\sigma^2),
\end{align}
in which for convenience we specify $\hat{\eta}_0=\hat{\rho}_0=0$ and $\frac{1}{\eta_0}:= 0$ so that $\frac{\hat{\eta}_0}{(1+\hat{\rho}_0)\eta_0}=0$.

In the following, we establish the adversarial regret bounds for constant and diminishing step sizes, respectively.

\textbf{Case 1: Constant $\eta_t$.}

For each step $t\in[1,T]$, let $\eta_t = \eta$ where $0<\eta\leq \frac{1}{4L}$ and $\hat{\eta}_t = \hat{\eta} = \frac{1}{2-4L\eta}$ with $\frac{1}{2}<\hat{\eta}\leq 1$. Also let $\rho_t=\rho= \frac{4L\eta}{\mu}$ and $\hat{\rho}_t = \hat{\rho}= \frac{\mu\eta}{2}$.

With $\frac{1}{1+\hat{\rho}} < 1$, $U_a$ in (\ref{4-rewrite-lemma}) is bounded by
\begin{align}
	\label{U_a-drogd-c}
	 {U_a} =& (\frac{\hat{\eta}}{\eta(1+\hat{\rho})}-0)\|w_1 - w^*\|^2 \\
	\nonumber &+ \sum_{t=2}^T(\frac{\hat{\eta}}{ \eta(1+\hat{\rho})}-\frac{\hat{\eta}}{\eta(1+\hat{\rho})})\|w_t - w^*\|^2 \\
	\nonumber\leq &\frac{1}{\eta}\|w_1 - w^*\|^2,	
\end{align}
and $U_b$ in (\ref{4-rewrite-lemma}) is bounded by
\begin{align}
	\label{U_b-drogd-c}
	{U_b} \leq &\sum_{t=1}^T\hat{\eta}(\frac{\hat{\rho}}{\eta}+\frac{2L\eta}{\rho}-\mu)\|w_t - w^*\|^2\\
	\nonumber= & \sum_{t=1}^T\hat{\eta}(\frac{\mu}{2}+\frac{\mu}{2}- \mu)\|w_t - w^*\|^2\\
	\nonumber= & 0.	
\end{align}

Combining (\ref{4-rewrite-lemma}) with (\ref{U_a-drogd-c}) and (\ref{U_b-drogd-c}), we conclude that Byzantine-robust distributed online gradient descent with robust bounded aggregation rules using constant step size has a linear adversarial regret bound, given by
\begin{align}
	\label {5-rewrite-lemma}
  R_{T:\eta} =&\sum_{t=1}^T(f_t(w_t)-f_t(w^*))\\
	\nonumber \leq& {U_a} +{U_b}
	+\sum_{t=1}^T(2(1+L\rho)\eta\xi^2
	+\frac{1}{(1+\hat{\rho})}(1+\frac{1}{\hat{\rho}})\eta C_{\alpha}^2\sigma^2)\\
	\nonumber\leq& \frac{1}{\eta}\|w_1 - w^*\|^2 + \sum_{t=1}^T((2\eta+\frac{ 8L^2\eta^2}{\mu})\xi^2
	+\frac{2}{\mu} C_{\alpha}^2\sigma^2)\\
	\nonumber\leq &\frac{1}{\eta}\|w_1 - w^*\|^2 + ( \frac{8L^2\eta^2}{\mu}+2\eta)\xi^2 T
	+\frac{2}{\mu} C_{\alpha}^2\sigma^2T.	
\end{align}
Specifically, if $\eta = \frac{c}{\sqrt{T}}$ where $c=\min\{\frac{1}{4L},\sqrt{\frac{\|w_1 - w^*\|^2}{2\xi^2}}\}$, then the adversarial regret bound becomes
\begin{align}
	R_{T:\frac{1}{\sqrt{T}}} \leq & \frac{8L^2c^2}{\mu}\xi^2+(\frac{\|w_1 - w^*\|^2}{c}+2c\xi^2)\sqrt{T} \\
	& +	\frac{2}{\mu}C_{\alpha}^2\sigma^2T. \nonumber
\end{align}

\textbf{Case 2: Diminishing $\eta_t$.}

For each step $t\in[1,T]$, let $\eta_t = \min\{\frac{1}{4L},\frac{8}{\mu t}\}$, $\hat{\rho}_t=\frac{\mu\eta_t}{2}=\min\{\frac{\mu}{8L},\frac{4}{t}\}$,
and
$$\rho_t = \left\{
\begin{array}{lcl}
	\frac{1}{\mu} & & \text{if }\eta_t=\frac{1}{4L},\\
	\frac{64L}{\mu^2t} & & \text{otherwise.}
\end{array} \right. $$
Actually, $\eta_t = \min\{\frac{1}{4L},\frac{8}{\mu t}\}$ is equivalent to
$$\eta_t = \left\{
\begin{array}{lcl}
	\frac{1}{4L} & & t\leq \lfloor \frac{32L}{\mu} \rfloor,   \\
	\frac{8}{\mu t} & & \text{otherwise.}
\end{array} \right. $$
It is obvious that $0<\eta_t\leq \frac{1}{4L}$, $\frac{1}{2}<\hat{\eta}_t\leq 1$, $0<\rho_t\leq \frac{2}{\mu}$, $0<\hat{\rho}_t\leq \frac{1}{8}$, $\frac{1}{1+\hat{\rho}_t}<1$ and
$\hat{\eta}_t\leq \hat{\eta}_{t-1}$.

In this case, $U_a$ in (\ref{4-rewrite-lemma}) is bounded by
\begin{align}
	\label{U-a-drogd-d}
	 {U_a} &=\sum_{t=1}^T(\frac{\hat{\eta}_t}{(1+\hat{\rho}_{t})\eta_t}-\frac{\hat{\eta}_{t-1}}{(1+\hat{\rho}_{t-1})\eta_{t-1}})\|w_t - w^*\|^2\\
	&\nonumber \leq \sum_{t=1}^T \frac{\hat{\eta}_{t-1}}{(1+\hat{\rho}_{t})} (\frac{1}{\eta_t}-\frac{1}{\eta_{t-1}})\|w_t - w^*\|^2\\
	\nonumber& \leq \sum_{t=1}^T (\frac{1}{\eta_t}-\frac{1}{\eta_{t-1}}) \|w_t - w^*\|^2.
\end{align}
As $\frac{1}{1+\hat{\rho}_t} < 1$ and $\hat{\eta}_t>\frac{1}{2}$, $(\frac{\hat{\rho}_t}{\eta_t}+\frac{2L\eta_t}{\rho_t}- \mu) \leq 0$ under our parameter setting, and thus $U_b$ in (\ref{4-rewrite-lemma}) is bounded by
\begin{align}
	\label{U-b-drogd-d}
	 {U_b}
	&\leq \sum_{t=1}^T\hat{\eta}_t(\frac{\hat{\rho}_t}{\eta_t}+\frac{2L\eta_t}{\rho_t}- \mu)\|w_t - w^*\|^2\\
	&\nonumber \leq \frac{1}{2} \sum_{t=1}^T(\frac{\mu}{2}+\frac{2L\eta_t}{\rho_t}- \mu)\|w_t - w^*\|^2\\
	\nonumber& = \sum_{t=1}^T(\frac{L\eta_t}{\rho_t} - \frac{\mu}{4}) \|w_t - w^*\|^2.
\end{align}
Combining (\ref{U-a-drogd-d}) with (\ref{U-b-drogd-d}), we know $U_a+U_b$ is bounded by
\begin{align}
	\label{U_ab-drogd-d}
	 {U_a}+{U_b}
	\leq & \sum_{t=1}^T (\frac{1}{\eta_t}-\frac{1}{\eta_{t-1}}) \|w_t - w^*\|^2
	\\
	\nonumber&+\sum_{t=1}^T(\frac{L\eta_t}{\rho_t} - \frac{\mu}{4}) \|w_t - w^*\|^2 \\
	= &\nonumber  (\frac{1}{\eta_1}-\frac{1}{\eta_{0}} +\frac{L\eta_1}{\rho_1} - \frac{\mu}{4}) \|w_1 - w^*\|^2 \\
	&\nonumber+ \sum_{t=2}^T (\frac{1}{\eta_t}-\frac{1}{\eta_{t-1}} +\frac{L\eta_t}{\rho_t} - \frac{\mu}{4}) \|w_t - w^*\|^2\\
	\leq & \nonumber \frac{1}{\eta_1} \|w_1 - w^*\|^2 + \sum_{t=2}^T 0* \|w_t - w^*\|^2\\
	\nonumber = & 4L \|w_1 - w^*\|^2.	
\end{align}
Here we use the fact that $(\frac{1}{\eta_t}-\frac{1}{\eta_{t-1}}+\frac{L\eta_t}{\rho_t}-\frac{\mu}{4}) \leq 0$ when $t\geq 2$ under our parameter setting.

Combining (\ref{4-rewrite-lemma}) and (\ref{U_ab-drogd-d}), we know that
Byzantine-robust distributed online gradient descent with robust bounded aggregation rules using diminishing step size has a
linear adversarial regret bound, given by
\begin{align}
	  R_{T:\frac{1}{t}} =&\sum_{t=1}^T(f_t(w_t)-f_t(w^*)) \leq {U_a}+{U_b} \\
	\nonumber & + \sum_{t=1}^T(2(1+L\rho_t)\eta_t\xi^2
	+\frac{1}{(1+\hat{\rho}_t)}(1+\frac{1}{\hat{\rho}_t})\eta_t C_{\alpha}^2\sigma^2)\\
	\nonumber {\leq}& 4L\|w_1 - w^*\|^2 +\sum_{t=1}^T((2+\frac{4L}{\mu})\eta_t\xi^2
	+\frac{2}{\mu} C_{\alpha}^2\sigma^2)\\
	\nonumber\leq &4L\|w_1 - w^*\|^2 +\sum_{t=1}^T(\frac{6L}{\mu}\eta_t\xi^2
	+\frac{2}{\mu} C_{\alpha}^2\sigma^2)\\
	\nonumber\leq & 4L\|w_1 - w^*\|^2+\frac{48L}{\mu^2}\xi^2 \log T
	+\frac{2}{\mu}C_{\alpha}^2\sigma^2 T.
\end{align}

In summary, using both constant and diminishing step sizes, Byzantine-robust distributed online gradient descent with robust bounded aggregation rules has linear adversarial regret bounds. This completes the proof.

\section{Proof of Theorem \ref{theorem-DROGDM}}
\label{stochasic regret analysis}
In this section, we establish the stochastic regret bound of Byzantine-robust distributed online momentum under Assumptions \ref{ass-1-new}, \ref{ass-2-new} and \ref{ass-3-new}. We start from two supporting lemmas whose proofs are in Appendix \ref{lemma-proof}.


First, denote $\bar{m}_t := \frac{1}{h}\sum_{j\in\mathcal{H}}m_t^j$ as the average momentum of the honest participants. Lemma \ref{lemma-momentum error} characterizes the recursion of $\sum_{j\in \mathcal{H}}\mathbb{E} \|m_t^j-\bar{m}_t\|^2$.

\begin{lemma}
Under Assumption \ref{ass-3-new}, for any step $t \in [1, T]$, it holds that
	\begin{align}
		\label{ineq-momentum-error}
		 & \sum_{j\in \mathcal{H}}\mathbb{E}\|m_{t+1}^j-\bar{m}_{t+1}\|^2 \\
		\nonumber\leq& (1-\nu_{t+1})\sum_{j\in \mathcal{H}}\mathbb{E}\|m_t^j-\bar{m}_t\|^2 + 2 h \nu_{t+1}^2\sigma^2,
	\end{align}
		where $\bar{m}_t := \frac{1}{h}\sum_{j\in\mathcal{H}}m_t^j$ the average momentum of the honest participants and $\nu_{t+1}$ is the momentum parameter at step $t+1$.
	\label{lemma-momentum error}
\end{lemma}

Lemma \ref{lemma:momentum recursion} further characterizes the recursion of $\mathbb{E}\|\bar{m}_{t} -\nabla F(w_{t})\|^2$.

\begin{lemma}
	Under Assumptions \ref{ass-1-new}, \ref{ass-2-new} and \ref{ass-3-new}, for any step $t \in [1, T]$, the distance between the average momentum of the honest participant $\bar{m}_{t+1}$ and the expected gradient $\nabla F(w_{t+1})$  follows
	\begin{align}
		\label{ineq-lemma-momentum}
		 &\mathbb{E}\|\bar{m}_{t+1}-\nabla F(w_{t+1})\|^2 \\
		\nonumber  \leq  & (1-\frac{\nu_{t+1}}{2}+\frac{6 \nu_{t+1}}{\lambda})\mathbb{E}\|\bar{m}_{t} -\nabla F(w_{t})\|^2 \\
		\nonumber &+\frac{12 \nu_{t+1} L^2}{\lambda}\mathbb{E}\|w_t-w^*\|^2-\frac{6 \nu_{t+1}}{\lambda}\mathbb{E}\|\nabla F(w_t)\|^2\\
		\nonumber &+\frac{6 \nu_{t+1}}{\lambda}C_{\alpha}^2\sum_{j\in\mathcal{H}}\mathbb{E}\|m_t^j-\bar{m}_t\|^2+\frac{\nu_{t+1}^2\sigma^2}{h},
	\end{align}	
where $\bar{m}_t := \frac{1}{h}\sum_{j\in\mathcal{H}}m_t^j$ is the average momentum of the honest participants and $w^*=\arg\min_{w\in \mathbb{R}^d}F(w)$ is the overall best solution. Therein, $\lambda=\frac{192L^2}{\mu^2}$ and $\sqrt{\lambda}L\eta_{t}\leq \nu_{t+1}\leq 1$ such that $\eta_t\leq\frac{\mu}{16L^2}$.
	\label{lemma:momentum recursion}
\end{lemma}

Now we prove Theorem \ref{theorem-DROGDM}. We bound $\mathbb{E} \|w_{t+1}-w^*\|^2 $ by
\begin{align}
	\label{ineq-A}
	 &\mathbb{E}\|w_{t+1}-w^*\|^2\\
	\nonumber \leq& \mathbb{E}\|w_{t+1}-\bar{z}_{t+1}+\bar{z}_{t+1}-w^*\|^2\\
	\nonumber\leq&
	\frac{1}{\rho_{t}}\mathbb{E}\||w_{t+1}-\bar{z}_{t+1}\|^2+\frac{1}{1-\rho_{t}}\mathbb{E}\|\bar{z}_{t+1}-w^*\|^2\\
	\nonumber=&\frac{1}{\rho_{t}}\underbrace{\mathbb{E}\|w_{t+1}-\bar{z}_{t+1}\|^2}_{U_{\hat{a}}}\\
	\nonumber &+\frac{1}{1-\rho_{t}} \underbrace{\mathbb{E}\|w_{t}-\eta_{t} \bar{m}_t-w^*\|^2}_{U_{\hat{b}}},
\end{align}
using the inequality $\|x+y\|^2 \leq \frac{1}{\rho_t}\|x\|^2 + \frac{1}{1-\rho_t}\|y\|^2$ that holds for any vectors $x$, $y$ and scalar $\rho_t \in (0,1)$.

According to Definition \ref{def:robust-bounded}, $U_{\hat{a}}$ can be bounded by
\begin{align}
\label{eq:new-0005}
		{U_{\hat{a}}}\leq \eta_t^2C_{\alpha}^2\mathbb{E}\max_{j\in \mathcal{H}}\| m_t^j-\bar{m}_{t}\|^2.
\end{align}
With $\nabla F(w^*)=0$, $U_{\hat{b}}$ can be bounded by
\begin{align}
	{U_{\hat{b}}}=&\mathbb{E}\|w_{t}-\eta_{t} \bar{m}_t-w^*\|^2\\
	\nonumber=&\mathbb{E}\|w_{t}-w^*-\eta_{t} \bar{m}_t+\eta_{t}\nabla F(w_t)-\eta_{t}\nabla F(w_t)\|^2\\
	\nonumber=&\mathbb{E}\|w_{t}-w^*\|^2+\eta_{t}^2 \mathbb{E}\|\bar{m}_t-\nabla F(w_t)+\nabla F(w_t)\|^2\\
	\nonumber &-2\eta_{t}\mathbb{E}\langle \bar{m}_t+\nabla F(w_t)-\nabla F(w_t),w_t-w^* \rangle \\
	\nonumber\leq&\mathbb{E}\|w_{t}-w^*\|^2+2\eta_{t}^2 \mathbb{E}\|\bar{m}_t-\nabla F(w_t)\|^2\\
	\nonumber&+2\eta_{t}^2 \mathbb{E}\|\nabla F(w_t)-\nabla F(w^*)\|^2 \\
	\nonumber&-2\eta_{t}\mathbb{E}\langle \bar{m}_t-\nabla F(w_t),w_t-w^* \rangle \\	
	\nonumber&-2\eta_{t}\mathbb{E}\langle \nabla F(w_t)-\nabla F(w^*),w_t-w^* \rangle,
\end{align}
Due to $L$-smoothness and $\mu$-strong convexity of $F$ in Assumptions \ref{ass-1-new} and \ref{ass-2-new}, $U_{\hat{b}}$ can be further bounded by
\begin{align}
\label{eq:new-0006}
	{U_{\hat{b}}}\leq&\mathbb{E}\|w_{t}-w^*\|^2+2\eta_{t}^2 \mathbb{E}\|\bar{m}_t-\nabla F(w_t)\|^2\\
	\nonumber &+2L^2\eta_{t}^2 \mathbb{E}\|w_{t}-w^*\|^2 \\
	\nonumber &+\frac{1}{\hat{\rho}_t}\eta_{t} \mathbb{E}\| \bar{m}_t-\nabla F(w_t)\|^2+\hat{\rho}_t\eta_{t}\mathbb{E}\|w_t-w^* \|^2 \\	
	\nonumber &-2\mu\eta_{t} \mathbb{E}\|w_{t}-w^*\|^2\\
	\nonumber=&(1+2L^2\eta_{t}^2+\hat{\rho}_t\eta_{t}-2\mu\eta_{t})\mathbb{E}\|w_{t}-w^*\|^2\\
	\nonumber &+(2\eta_{t}^2+\frac{1}{\hat{\rho}_t}\eta_{t})\mathbb{E}\|\bar{m}_t-\nabla F(w_t)\|^2,
\end{align}
where $\hat{\rho}_t>0$ is an arbitrary positive scalar.

Substituting the upper bounds of ${U_{\hat{a}}}$ in \eqref{eq:new-0005} and ${U_{\hat{b}}}$ in \eqref{eq:new-0006} into (\ref{ineq-A}) yields
\begin{align}
	\label{ineq-A1}
	& \mathbb{E}\|w_{t+1}-w^*\|^2 \\
     \leq & \frac{1+2L^2\eta_{t}^2+\hat{\rho}_t\eta_{t}-2\mu\eta_{t}}{1-\rho_{t}}\mathbb{E}\|w_{t}-w^*\|^2 \nonumber \\
	\nonumber + & \frac{2\eta_{t}^2+\frac{1}{\hat{\rho}_t}\eta_{t}}{1-\rho_{t}}\mathbb{E}\|\bar{m}_t-\nabla F(w_t)\|^2 + \frac{C_{\alpha}^2\eta_t^2}{\rho_t}\sum_{j\in\mathcal{H}}\mathbb{E}\| m_t^j-\bar{m}_{t}\|^2.	
\end{align}

Letting $\rho_{t} = \frac{\mu\eta_{t}}{2}$, $\hat{\rho}_t=\frac{3\mu}{8}$ and also requiring $\eta_t \leq \frac{\mu}{16L^2}$ as in Lemma \ref{lemma:momentum recursion}, we know that $L^2\eta_{t}^2\leq \frac{\mu\eta_{t}}{16}$. Therefore, we have
$$\frac{1+2L^2\eta_{t}^2+\hat{\rho}_t\eta_{t}-2\mu\eta_{t}}{1-\rho_{t}}\leq 1-\mu\eta_{t}, \frac{2\eta_{t}^2+\frac{1}{\hat{\rho}_t}\eta_{t}}{1-\rho_{t}}\leq \frac{4}{\mu}\eta_t.$$

With these inequalities, multiplying the both sides of (\ref{ineq-A1}) by $\mu$ yields
\begin{align}
	\label{ineq-A2}
	& \mu\mathbb{E}\|w_{t+1}-w^*\|^2
	\leq (\mu-\mu^2\eta_{t})\mathbb{E}\|w_{t}-w^*\|^2 \\
	\nonumber&+4\eta_{t}\mathbb{E}\|\bar{m}_t-\nabla F(w_t)\|^2 +2\eta_{t}C_{\alpha}^2\sum_{j\in\mathcal{H}}\mathbb{E}\|m_t^j-\bar{m}_t\|^2.
\end{align}

Motivated by \eqref{ineq-A2}, we construct a Lyapunov function $A_{t}$ to enable the subsequent proof, as
\begin{align}
	 A_{t} := &\mu\mathbb{E}\|w_{t}-w^*\|^2 +{k}\mathbb{E}\|\bar{m}_{t} -\nabla F(w_{t})\|^2 \\
	\nonumber&+ {p} C_{\alpha}^2\sum_{j\in \mathcal{H}}\mathbb{E}\|m_t^j-\bar{m}_t\|^2,
\end{align}
where {$k$} and {$p$} are two positive constants.


According to Lemmas \ref{lemma-momentum error} and \ref{lemma:momentum recursion}, multiplying (\ref{ineq-lemma-momentum}) by $k$ and (\ref{ineq-momentum-error}) by $p$ and then followed by combining with (\ref{ineq-A2}), we have
\begin{align}
	 &\mu\mathbb{E}\|w_{t+1}-w^*\|^2  + {k}\mathbb{E}\|\bar{m}_{t+1}-\nabla F(w_{t+1})\|^2 \\
	\nonumber&+ {p} C_{\alpha}^2\sum_{j\in\mathcal{H}}\mathbb{E}\|m_{t+1}^j-\bar{m}_{t+1}\|^2\\
	\nonumber \leq & (\mu-\mu^2\eta_{t})\mathbb{E}\|w_{t}-w^*\|^2
	+4\eta_{t}\mathbb{E}\|\bar{m}_t-\nabla F(w_t)\|^2 \\
	\nonumber&+2\eta_{t}C_{\alpha}^2\sum_{j\in\mathcal{H}}\mathbb{E}\|m_t^j-\bar{m}_t\|^2\\
	\nonumber&+  (1- \frac{\nu_{t+1}}{2} + \frac{6 \nu_{t+1}}{\lambda}) {k}\mathbb{E}\|\bar{m}_{t} -\nabla F(w_{t})\|^2 \\
	\nonumber&+\frac{\nu_{t+1}^2\sigma^2 {k}}{h}+\frac{12 \nu_{t+1} {k}L^2}{\lambda}\mathbb{E}\|w_t-w^*\|^2\\
	\nonumber&-\frac{6 \nu_{t+1} {k}}{\lambda}\mathbb{E}\|\nabla F(w_t)\|^2+\frac{6 \nu_{t+1} {k}}{\lambda}C_{\alpha}^2\sum_{j\in\mathcal{H}}\mathbb{E}\|m_t^j-\bar{m}_t\|^2\\
	\nonumber&+{p}(1-\nu_{t+1})C_{\alpha}^2\sum_{j\in\mathcal{H}}\mathbb{E}\|m_t^j-\bar{m}_t\|^2 + {2hp}C_{\alpha}^2 \nu_{t+1}^2\sigma^2.
\end{align}	
Reorganizing the terms yields
\begin{align}
	\label{ineq-A3}
	 &\mu\mathbb{E}\|w_{t+1}-w^*\|^2  + {k}\mathbb{E}\|\bar{m}_{t+1}-\nabla F(w_{t+1})\|^2 \\
	\nonumber&+ {p} C_{\alpha}^2\sum_{j\in \mathcal{H}}\mathbb{E}\|m_{t+1}^j-\bar{m}_{t+1}\|^2\\
	\nonumber \leq & (\mu-\mu^2\eta_{t}+\frac{12 \nu_{t+1} {k}L^2}{\lambda})\mathbb{E}\|w_{t}-w^*\|^2
	\\
	\nonumber & +(4\eta_{t}+{k}- \frac{\nu_{t+1} {k}}{2} + \frac{6 \nu_{t+1} {k}}{\lambda})\mathbb{E}\|\bar{m}_t-\nabla F(w_t)\|^2\\
	\nonumber&+(2\eta_{t}+\frac{6 \nu_{t+1} {k}}{\lambda}+p-p\nu_{t+1})C_{\alpha}^2\sum_{j\in \mathcal{H}}\mathbb{E}\|m_t^j-\bar{m}_t\|^2\\
	\nonumber&-\frac{6 \nu_{t+1} {k}}{\lambda}\mathbb{E}\|\nabla F(w_t)\|^2  +(\frac{k}{h}+2hpC_{\alpha}^2 )\nu_{t+1}^2\sigma^2.	
\end{align}

Letting $\lambda=\frac{192L^2}{\mu^2}$, $\nu_{t+1} =\sqrt{\lambda}L\eta_{t}=\frac{8\sqrt{3}L^2}{\mu}\eta_{t}$, ${k}=\frac{\sqrt{3}\mu}{2L^2}$ and ${p}=\frac{3\sqrt{3}\mu}{8L^2}$, we know that the coefficients in \eqref{ineq-A3} satisfy
\begin{align*}
	\mu-\mu^2\eta_{t}+\frac{12 \nu_{t+1} {k} L^2}{\lambda}\leq\mu(1-\frac{\mu}{4}\eta_t),
\end{align*}
\begin{align*}
	4\eta_{t}+{k}- \frac{\nu_{t+1} {k}}{2} + \frac{6 \nu_{t+1} {k}}{\lambda}\leq {k}(1-\frac{\sqrt{3}L^2}{\mu}\eta_t),
\end{align*}
\begin{align*}
	2\eta_{t}+\frac{6 \nu_{t+1} {k}}{\lambda}+p-p\nu_{t+1} \leq {p}(1-\frac{4\sqrt{3}L^2}{\mu}\eta_t).
\end{align*}
Further choose $\theta_t = \min \{\frac{\mu}{4}\eta_t,\frac{\sqrt{3}L^2}{\mu}\eta_t,\frac{4\sqrt{3}L^2}{\mu}\eta_t\}=\frac{\mu}{4}\eta_t$. Thus, (\ref{ineq-A3}) can be rewritten as
\begin{align}
	\label{ineq-A4}
	 A_{t+1} \leq & (1-\frac{\mu}{4}\eta_t)A_{t} - \frac{3\mu^2}{8L^2}\eta_t\mathbb{E}\|\nabla F(w_t)\|^2  \\
	\nonumber &+ \frac{96\sqrt{3}+144\sqrt{3}h^2 C_{\alpha}^2}{\mu h} L^2\sigma^2\eta_t^2.	
\end{align}
Rearranging the terms again yields
\begin{align}
	\label{ineq-nablaF}
	 \frac{3\mu^2}{8L^2}\eta_t\mathbb{E}\|\nabla F(w_t)\|^2  \leq & (1-\frac{\mu}{4}\eta_t)A_{t} - A_{t+1} \\
	\nonumber&+ \frac{96\sqrt{3}+144\sqrt{3}h^2C_{\alpha}^2}{\mu h}L^2\sigma^2\eta_t^2.	
\end{align}
By $\mu$-strong convexity of $F$ in Assumption \ref{ass-2-new}, for any $w \in \mathbb{R}^d$, it holds that
\begin{align}
	F(w)-F(w^*)\leq \frac{1}{2\mu}\|\nabla F(w)\|^2.
	\label{ineq-FFW}
\end{align}
Thus, combining (\ref{ineq-FFW}) with (\ref{ineq-nablaF}) yields
\begin{align}
\label{eq:new-0007}
	&\mathbb{E}(F(w_t)-F(w^*))
	\leq \frac{1}{2\mu}\mathbb{E}\|\nabla F(w_t)\|^2\\
	\nonumber \leq & \frac{4L^2}{3\mu^3}((\frac{1}{\eta_t}-\frac{\mu}{4})A_{t} - \frac{1}{\eta_t}A_{t+1}) \\
	\nonumber&+ \frac{128\sqrt{3}+192\sqrt{3}h^2C_{\alpha}^2}{\mu^4 h}L^4\sigma^2\eta_t.
\end{align}	

For convenience, define $\frac{1}{\eta_0} := 0$ and $\nu_1 = \eta_1$. Summing up \eqref{eq:new-0007} from $t=1$ to $T$, we bound the stochastic regret of distributed online momentum gradient descent with robust bounded aggregation as
\begin{align}
	\label{eq:sum-DROGDM}
	 S_T = & \sum_{t=1}^{T}\mathbb{E}(F(w_t)-F(w^*))\\
	\nonumber \leq & \sum_{t=1}^{T}\frac{4L^2}{3\mu^3}((\frac{1}{\eta_t}-\frac{\mu}{4})A_{t} - \frac{1}{\eta_t}A_{t+1}) \\
	\nonumber&+ \sum_{t=1}^{T}\frac{128\sqrt{3}+96\sqrt{3}(h+1)C_{\alpha}^2}{\mu^4 h}L^4\sigma^2\eta_t\\
	\nonumber  \leq & \frac{4L^2}{3\mu^3}\underbrace{\sum_{t=1}^{T}(\frac{1}{\eta_t}-\frac{1}{\eta_{t-1}}-\frac{\mu}{4})A_{t}}_{U_{\hat{c}}} \\
	\nonumber& + \frac{128\sqrt{3}+192\sqrt{3}h^2C_{\alpha}^2}{\mu^4 h}L^4\sigma^2\sum_{t=1}^{T}\eta_t.	
\end{align}

In the following, we establish the stochastic regret bounds for constant and diminishing step sizes and momentum parameters, respectively.

\textbf{Case 1: Constant $\eta_t$ and $\nu_t$}

For each step $t \in [1,T]$, let $\eta_t=\eta$  where $0<\eta \leq \frac{\mu}{16L^2}$ and $\nu_{t+1} = \nu = \frac{8\sqrt{3}L^2}{\mu}\eta < 1$.

With these parameter settings, $U_{\hat{c}}$ in (\ref{eq:sum-DROGDM}) is bounded by
\begin{align}
	 U_{\hat{c}} &=(\frac{1}{\eta}-\frac{\mu}{4})A_{1}+ \sum_{t=2}^{T}(\frac{1}{\eta}-\frac{1}{\eta}-\frac{\mu}{4})A_{t}\leq \frac{1}{\eta}A_{1}.
\end{align}
Therefore, the stochastic regret in \eqref{eq:sum-DROGDM} is bounded by
\begin{align}
	S_{T:\eta}\leq \frac{1}{\eta}A_{1} +\frac{128\sqrt{3}+192\sqrt{3}h^2C_{\alpha}^2}{\mu^4 h}L^4\sigma^2\eta T,	
\end{align}
where
\begin{align}
	A_1 =&\mu\mathbb{E}\|w_{1}-w^*\|^2 +\frac{\sqrt{3}\mu}{2L^2}\mathbb{E}\|\bar{m}_{1} -\nabla F(w_{1})\|^2 \\
	\nonumber&+ \frac{3\sqrt{3}\mu}{8L^2} C_{\alpha}^2\sum_{j\in \mathcal{H}}\mathbb{E}\|m_1^j-\bar{m}_1\|^2.
\end{align}

Specifically, if $\eta = \frac{c}{\sqrt{T}}\leq\frac{\mu}{16L^2}$ where $c$ is a constant, then the stochastic regret bound becomes
\begin{align}
\hspace{-1.5em}S_{T:\frac{1}{\sqrt{T}}} \leq \frac{A_1}{c}\sqrt{T} +
	\frac{128\sqrt{3}+192\sqrt{3}h^2C_{\alpha}^2}{\mu^4 h}L^4\sigma^2 c \sqrt{T}.
\end{align}

\textbf{Case 2: Diminishing $\eta_t$ and $\nu_t$}

For each step $t \in [1,T]$, let $\eta_t=\min\{\frac{\mu}{16L^2},\frac{c}{t}\}$ where $c\geq \frac{4}{\mu}$ is a constant $\nu_{t+1}=\frac{8\sqrt{3}L^2}{\mu}\eta_t$.

With these parameter settings, for $t \in [2,T]$ we always have $(\frac{1}{\eta_t}-\frac{1}{\eta_{t-1}}-\frac{\mu}{4})\leq 0$. Thus, $U_{\hat{c}}$ in (\ref{eq:sum-DROGDM}) is bounded by
\begin{align}
	 U_{\hat{c}} &=(\frac{1}{\eta_1}-\frac{\mu}{4})A_{1}+ \sum_{t=2}^{T}(\frac{1}{\eta_t}-\frac{1}{\eta_{t-1}}-\frac{\mu}{4})A_{t}\\
	\nonumber&\leq \frac{1}{\eta_1}A_{1} \leq \frac{16L^2A_{1}}{\mu}.
\end{align}
Therefore, the stochastic regret bound is given by
\begin{align}
\hspace{-1.5em} S_{T:\frac{1}{t}}
\leq \frac{16L^2A_{1}}{\mu} + \frac{128\sqrt{3}+192\sqrt{3}h^2C_{\alpha}^2}{\mu^4 h}L^4\sigma^2 c \log{T}.
\end{align}

\newpage

\newpage

\section{Proof of Supporting Lemmas and Corollary}
\label{lemma-proof}

\subsection{Proof of Lemma \ref{mean-bound}}
\label{lemma-proof-1}

\begin{proof}
	The squared norm of the average gradient can be bounded by
	\begin{align}
		&\|\nabla{f_t(w_t)}\|^2 \\
		\nonumber=&\|\nabla{f_t(w_t)}- \nabla{f_t(w^*)}+ \nabla{f_t(w^*)}\|^2\\
		\nonumber\leq& 2\|\nabla{f_t(w_t)}-\nabla{f_t(w^*)}\|^2
		+2\|\nabla{f_t(w^*)}\|^2  \\
		\nonumber\overset{(\ref{ass-4-equa})}{\leq}& 2 \|\nabla{f_t(w_t)}-\nabla{f_t(w^*)}\|^2+2\xi^2  \\
		\nonumber \leq & 4L(f_t(w_t)-f_t(w^*)-\langle \nabla f_t(w^*),w_t-w^*\rangle) +2\xi^2.
	\end{align}
The last inequality uses the fact that an $L$-smooth and convex function $f_t^j$ satisfies
	\begin{align}
		\label{ineq2}
		&\frac{1}{2L}\|\nabla f_t^j(y)-\nabla f_t^j(x)\|^2 \\	
		\leq & f_t^j(y) - f_t^j(x) - \langle \nabla{f_t^j(x)},y-x\rangle, \nonumber
	\end{align}
for any vectors $x$, $y$, according to Theorem 2.1.5 in \cite{nesterov1998introductory}.

Due to $ - 2\langle x,y\rangle \leq \frac{1}{\rho_t}\|x\|^2 + \rho_t\|y\|^2$ for any vectors $x$, $y$ and scalar $\rho_t > 0$, we further have
	\begin{align}
		&\|\nabla{f_t(w_t)}\|^2\\
		\nonumber\leq & 4L(f_t(w_t)-f_t(w^*)) \\
		\nonumber&+ 2L(\rho_t\|\nabla f_t(w^*)\|^2 +\frac{1}{\rho_t}\|w_t - w^*\|^2)+ 2\xi^2\\
	\nonumber\overset{(\ref{ass-4-equa})}{\leq} &
		4L(f_t(w_t)-f_t(w^*))  + \frac{2L}{\rho_t}\|w_t-w^*\|^2 + 2(1+L\rho_t)\xi^2,
	\end{align}
which completes the proof.
\end{proof}

\subsection{Proof of Lemma \ref{recursion}}
\label{lemma-proof-2}

\begin{proof}
With the inequality $\|x+y\|^2 \leq (1+\hat{\rho}_t)\|x\|^2 + (1+\frac{1}{\hat{\rho}_t})\|y\|^2$ that holds for any vectors $x$, $y$ and scalar $\hat{\rho}_t>0$, $\|w_{t+1}-w^*\|^2$ can be bounded as
\begin{align}
	 \|w_{t+1}-w^*\|^2 &=\|w_{t+1}- \bar{z}_{t+1}+\bar{z}_{t+1} - w^*\|^2 \\
	\nonumber&\leq(1+\hat{\rho}_t) \|\bar{z}_{t+1}-w^*\|^2 \\
	\nonumber  & + (1+\frac{1}{\hat{\rho}_t}) \|w_{t+1}-\bar{z}_{t+1}\|^2.
	\label{eq:U1U2}
\end{align}
where $\bar{z}_{t+1}: = \frac{1}{h}\sum_{j\in\mathcal{H}}z_{t+1}^j=\frac{1}{h}\sum_{j\in\mathcal{H}}(w_t-\eta_t\nabla f_t^j(w_t)) = w_t - \eta_t \nabla f_t(w_t)$. Below, we respectively bound $\|\bar{z}_{t+1}-w^*\|^2$ and $\|w_{t+1}-\bar{z}_{t+1}\|^2$.

We expand $\|\bar{z}_{t+1}-w^*\|^2$ to
	\begin{align}
			&\|\bar{z}_{t+1}-w^*\|^2  \\
			\nonumber=&\|w_t - \eta_t \nabla f_t(w_t)-w^*\|^2\\
			\nonumber=&\|w_t -  w^*\|^2+\eta_t^2\|\nabla f_t(w_t)\|^2 -2\eta_t \langle w_t- w^*,\nabla f_t(w_t) \rangle.
	\end{align}
Substituting $\|\nabla f_t(w_t)\|^2$ with (\ref{mean-bound-eq}) in Lemma \ref{mean-bound} and $\langle w_t- w^*,\nabla f_t(w_t) \rangle$ with (\ref{ass-2-equa}) in Assumption \ref{ass-2}, we further have
	\begin{align}
		&\|\bar{z}_{t+1}-w^*\|^2 \label{U_1-eq} \\
		\nonumber	\leq & \|w_t - w^*\|^2+ \eta_t^2(4L(f_t(w_t)-f_t(w^*))  \\
		\nonumber  & + \frac{2L}{\rho_t}\|w_t-w^*\|^2 + 2(1+L\rho_t)\xi^2) \\
		\nonumber  & - 2 \eta_t \frac{1}{h}\sum_{j\in \mathcal{H}} (f_t^j(w_t)-f_t^j(w^*)+\frac{\mu}{2}\|w_t - w^*\|^2) \\
		\nonumber	\leq&\|w_t - w^*\|^2 + 4L\eta_t^2(f_t(w_t)-f_t(w^*))  + \frac{2L\eta_t^2}{\rho_t}\|w_t-w^*\|^2 \\
		\nonumber	 & + 2(1+L\rho_t)\eta_t^2\xi^2  -2 \eta_t (f_t(w_t)-f_t(w^*))- \eta_t\mu\|w_t - w^*\|^2 \\
		\nonumber	 \leq & (1+\frac{2L\eta_t^2}{\rho_t}- \eta_t\mu)\|w_t - w^*\|^2 \\
	    \nonumber & + (4L\eta_t^2-2 \eta_t)(f_t(w_t)-f_t(w^*))+2(1+L\rho_t)\eta_t^2\xi^2.
		\end{align}

For $\|w_{t+1}-\bar{z}_{t+1}\|^2$, by Definition \ref{def:robust-bounded} and Assumption \ref{ass-3}, we have
\begin{align}
	\label{U_3-eq}
 &\|w_{t+1}-\bar{z}_{t+1}\|^2 \\
\nonumber \leq & C_{\alpha}^2\max_{j\in\mathcal{H}}\|w_t^j-\bar{z}_{t+1}\|^2 \\
\nonumber = & C_{\alpha}^2\max_{j\in\mathcal{H}}\|w_t -\eta_t\nabla f_t^j(w_t)-w_t + \eta_t \nabla f_t(w_t)\|^2 \\
\nonumber \leq & \eta_t^2 C_{\alpha}^2\max_{j\in\mathcal{H}}\|\nabla f_t(w_t) - \nabla f_t^j(w_t)\|^2 \\
\nonumber\leq &\eta_{t}^2C_{\alpha}^2\sigma^2.	
\end{align}

Substituting (\ref{U_1-eq}) and (\ref{U_3-eq}) into \eqref{eq:U1U2} yields
\begin{align}
	&\|w_{t+1}-w^*\|^2\\
	\nonumber {\leq}&
	(1+\hat{\rho}_t)(1+\frac{2L\eta_t^2}{\rho_t}- \eta_t\mu)\|w_t - w^*\|^2 \\
	 \nonumber &+(1+\hat{\rho}_t)(4L\eta_t^2-2 \eta_t)(f_t(w_t)-f_t(w^*))\\
	\nonumber &+2(1+\hat{\rho}_t)(1+L\rho_t)\eta_t^2\xi^2
	+(1+\frac{1}{\hat{\rho}_t})\eta_t^2C_{\alpha}^2\sigma^2,
\end{align}
which completes the proof.
\end{proof}

\subsection{Proof of Lemma \ref{lemma-momentum error}}
\label{lemma-proof-3}

\begin{proof}	
	Given i.i.d. data, Assumption \ref{ass-3-new} yields
	\begin{align}
		\mathbb{E}\|f_{t+1}(w_{t+1})-\nabla F(w_{t+1}) \|^2 \leq \frac{1}{h}\sigma^2.
		\label{ass-3-new-equa-h}
	\end{align}
According to the definitions of
	$$m_{t+1}^j=\nu_{t+1}\nabla f_{t+1}^j(w_{t+1})+(1-\nu_{t+1})m_{t}^j,$$ and
	$$\bar{m}_{t+1}=\nu_{t+1}\nabla f_{t+1}(w_{t+1})+(1-\nu_{t+1})\bar{m}_{t},$$
	we have
	\begin{align}
		 &\sum_{j\in \mathcal{H}}\mathbb{E}\|m_{t+1}^j-\bar{m}_{t+1}\|^2 \\
		\nonumber\leq& (1-\nu_{t+1})^2\sum_{j\in \mathcal{H}}\mathbb{E}\|m_{t}^j-\bar{m}_{t}\|^2 \\
		\nonumber & + \nu_{t+1}^2\sum_{j\in \mathcal{H}}\mathbb{E}\|\nabla f_{t+1}^j(w_{t+1})-\nabla f_{t+1}(w_{t+1})\|^2\\
		\nonumber\leq& (1-\nu_{t+1})\sum_{j\in \mathcal{H}}\mathbb{E}\|m_{t}^j-\bar{m}_{t}\|^2\\
		\nonumber &+ \nu_{t+1}^2\sum_{j\in \mathcal{H}}\mathbb{E}\|\nabla f_{t+1}^j(w_{t+1})-\nabla F(w_{t+1})\|^2\\
		\nonumber &+ \nu_{t+1}^2\sum_{j\in \mathcal{H}}\mathbb{E}\|\nabla F(w_{t+1})-\nabla f_{t+1}(w_{t+1})\|^2\\
		\nonumber\overset{(\ref{ass-3-new-equa}),(\ref{ass-3-new-equa-h})}{\leq}& (1-\nu_{t+1})\sum_{j\in \mathcal{H}}\mathbb{E}\|m_{t}^j-\bar{m}_{t}\|^2 +\nu_{t+1}^2 h\sigma^2+\nu_{t+1}^2\sigma^2\\
		\nonumber\leq& (1-\nu_{t+1})\sum_{j\in \mathcal{H}}\mathbb{E}\|m_{t}^j-\bar{m}_{t}\|^2 + (h+1)\nu_{t+1}^2\sigma^2\\
		\nonumber\leq& (1-\nu_{t+1})\sum_{j\in \mathcal{H}}\mathbb{E}\|m_t^j-\bar{m}_t\|^2 + 2 h \nu_{t+1}^2\sigma^2,
	\end{align}	
which completes the proof.
\end{proof}

\subsection{Proof of Lemma \ref{lemma:momentum recursion}}
\label{lemma-proof-4}

\begin{proof}
	We first define the following filtration
	$$
	\mathcal{W}_t := \hat{\sigma}(\bigcup_{j\in\mathcal{N}}\{w_{0}^j, w_{1}^j,...,w_{t}^j\}),
	$$
	where $\hat{\sigma}$ stands for the $\sigma$-field.
	
	Observe that
	\begin{align}
		\label{eq:67}
		&\mathbb{E} [ \|\bar{m}_{t+1}-\nabla F(w_{t+1}) \|^2 |\mathcal{W}_{t}] \\
		\nonumber=& \mathbb{E}[ \|\nu_{t+1}\nabla f_{t+1}(w_{t+1})+(1-\nu_{t+1})\bar{m}_{t}  \\
		 \nonumber& -\nabla F(w_{t+1}) \|^2 | \mathcal{W}_{t} ] \\
		\nonumber=&\mathbb{E}[\|\nu_{t+1}\nabla f_{t+1}(w_{t+1})-\nu_{t+1}\nabla F(w_{t+1})+\nu_{t+1}\nabla F(w_{t+1})\\
		 \nonumber & + (1-\nu_{t+1})\bar{m}_{t}-\nabla F(w_{t+1}) \|^2|  \mathcal{W}_{t} ] \\
		\nonumber\leq& \nu_{t+1}^2\mathbb{E}[\|\nabla f_{t+1}(w_{t+1})-\nabla F(w_{t+1})\|^2|  \mathcal{W}_{t}]\\
		\nonumber& + (1-\nu_{t+1})^2\mathbb{E}[\|\bar{m}_{t}-\nabla F(w_{t+1})\|^2|  \mathcal{W}_{t}] \\
		\nonumber\overset{(\ref{ass-3-new-equa})}{\leq}&(1-\nu_{t+1})\mathbb{E}[\|\bar{m}_{t}-\nabla F(w_{t})+\nabla F(w_{t})\\
		\nonumber &-\nabla F(w_{t+1})\|^2|  \mathcal{W}_{t}] +\frac{\nu_{t+1}^2\sigma^2}{h}.
	\end{align}
Taking the expectation on both sides of \eqref{eq:67}, we obtain
	\begin{align}
	\label{eq:67-2}
	&\mathbb{E}\|\bar{m}_{t+1}-\nabla F(w_{t+1})\|^2 \\
	\nonumber\leq&(1-\nu_{t+1})\mathbb{E}\|\bar{m}_{t}-\nabla F(w_{t})+\nabla F(w_{t})-\nabla F(w_{t+1})\|^2\\
	\nonumber & +\frac{\nu_{t+1}^2\sigma^2}{h}.
\end{align}

	By $\|x+y\|^2 \leq (1+\frac{\nu_{t+1}}{2})\|x\|^2 + (1+\frac{2}{\nu_{t+1}})\|y\|^2$ that holds for any vectors $x$ and $y$, where $\nu_{t+1} > 0$ is the momentum parameter, using $L$-smoothness of $F$ in Assumption \ref{ass-1-new}, we bound the first term at the right-hand side of \eqref{eq:67-2} as
	\begin{align}
		\label{eq:68} & \hspace{-1em} (1-\nu_{t+1})\mathbb{E}\|\bar{m}_{t}-\nabla F(w_{t})+\nabla F(w_{t})-\nabla F(w_{t+1})\|^2\\
		\nonumber\hspace{-1em}\leq& (1-\nu_{t+1})(1+\frac{\nu_{t+1}}{2})\mathbb{E}\|\bar{m}_{t}-\nabla F(w_{t})\|^2\\
		 \nonumber\hspace{-1em} &+(1-\nu_{t+1})(1+\frac{2}{\nu_{t+1}})\mathbb{E}\|\nabla F(w_{t})-\nabla F(w_{t+1})\|^2\\
		\nonumber\hspace{-1em}\leq & (1-\frac{\nu_{t+1}}{2})\mathbb{E} \|\bar{m}_{t}-\nabla F(w_{t})\|^2+\frac{2L^2}{\nu_{t+1}}\mathbb{E} \|w_t-w_{t+1}\|^2.
		\end{align}
	Define $\bar{z}_{t+1}:= \frac{1}{h}\sum_{j\in\mathcal{H}}w_{t+1}^j$ and hence $\bar{z}_{t+1} = \frac{1}{h}\sum_{j\in\mathcal{H}}$ $(w_{t}-\eta_t m_t^j)={w}_{t} - \eta_t\bar{m}_t$. The term $\mathbb{E}\|w_t-w_{t+1}\|^2$ in \eqref{eq:68} satisfies
	\begin{align}
\label{eq:new-0001}
		&\mathbb{E}\|w_t-w_{t+1}\|^2\\
		\nonumber= & \mathbb{E}\|w_t - \bar{z}_{t+1} + \bar{z}_{t+1}-w_{t+1} - \eta_t\nabla F(w_{t})+\eta_t\nabla F(w_{t}) \|^2\\
		\nonumber= & \mathbb{E}\|\bar{z}_{t+1}-w_{t+1} +\eta_t(\bar{m}_t- \nabla F(w_{t}))+\eta_t\nabla F(w_{t})\|^2\\
		\nonumber\leq & 3\mathbb{E}\|\bar{z}_{t+1}-w_{t+1}\|^2 +3\eta_t^2\mathbb{E}\|\bar{m}_t- \nabla F(w_{t})\|^2\\
		\nonumber &+ 3\eta_t^2\mathbb{E}\|\nabla F(w_{t})\|^2.
	\end{align}
By Definition \ref{def:robust-bounded}, we have
\begin{align}
	\mathbb{E}\|\bar{z}_{t+1}-w_{t+1}\|^2 \leq & C_{\alpha}^2\mathbb{E}\max_{j\in\mathcal{H}}\|w_{t+1}^j-\bar{z}_{t+1}\|^2\\
	\nonumber=&C_{\alpha}^2\mathbb{E}\max_{j\in \mathcal{H}}\|\eta_t m_t^j-\eta_t\bar{m}_{t}\|^2\\
\nonumber	\leq&\eta_t^2C_{\alpha}^2\sum_{j\in \mathcal{H}}\mathbb{E}\| m_t^j-\bar{m}_{t}\|^2,
\end{align}
such that \eqref{eq:new-0001} turns to
\begin{align}
	\label{eq:71}
	&\mathbb{E}\|w_t-w_{t+1}\|^2\\
	\nonumber\leq & 3\eta_t^2\sum_{j\in \mathcal{H}}\mathbb{E}\| m_t^j-\bar{m}_{t}\|^2 +3\eta_t^2\mathbb{E}\|\bar{m}_t- \nabla F(w_{t})\|^2\\
	\nonumber & +3\eta_t^2\mathbb{E}\|\nabla F(w_{t})\|^2.
\end{align}

Hence, combining \eqref{eq:67-2} with \eqref{eq:68} and \eqref{eq:71} yields
		\begin{align}
\label{eq:new-0002}
		\hspace{-1em} & \mathbb{E}\|\bar{m}_{t+1}-\nabla F(w_{t+1})\|^2 \\
		\nonumber\hspace{-1em} \leq& (1-\frac{\nu_{t+1}}{2})\mathbb{E} \|\bar{m}_{t}-\nabla F(w_{t})\|^2\\
		\nonumber\hspace{-1em} &+\frac{6L^2\eta_{t}^2}{\nu_{t+1}} C_{\alpha}^2\sum_{j\in\mathcal{H}}\mathbb{E}\| m_t^j-\bar{m}_{t}\|^2
		+\frac{6L^2\eta_{t}^2}{\nu_{t+1}}\mathbb{E} \|\bar{m}_t-\nabla F(w_t)\|^2 \\
		\nonumber\hspace{-1em} &+\frac{6L^2\eta_{t}^2}{\nu_{t+1}} \mathbb{E} \|\nabla F(w_t)\|^2 +\frac{\nu_{t+1}^2\sigma^2}{h}.	
	\end{align}
Observing that $\nabla F(w^*) = 0$, we have
	\begin{align}
\label{eq:new-0003}
		\hspace{-1em}&\mathbb{E}\|\bar{m}_{t+1}-\nabla F(w_{t+1})\|^2 \\
		\nonumber \hspace{-1em}\leq & (1-\frac{\nu_{t+1}}{2})\mathbb{E} \|\bar{m}_{t}-\nabla F(w_{t})\|^2\\
		\nonumber\hspace{-1em}  &+\frac{6L^2\eta_{t}^2}{\nu_{t+1}} C_{\alpha}^2\sum_{j\in\mathcal{H}}\mathbb{E}\| m_t^j-\bar{m}_{t}\|^2 +\frac{6L^2\eta_{t}^2}{\nu_{t+1}}\mathbb{E} \|\bar{m}_t-\nabla F(w_t)\|^2 \\
		\nonumber\hspace{-1em} &+\frac{6L^2\eta_{t}^2}{\nu_{t+1}} \mathbb{E} \|\nabla F(w_t) -\nabla F(w^*)\|^2
		+\frac{\nu_{t+1}^2\sigma^2}{h}.
	\end{align}
	
Letting $\lambda=\frac{192L^2}{\mu^2}$ and $\sqrt{\lambda}L\eta_{t}\leq \nu_{t+1}\leq 1$ such that $\eta_t\leq\frac{\mu}{16L^2}$ and $L^2\eta_{t}^2\leq \frac{\nu_{t+1}^2}{\lambda}$, we can rewrite \eqref{eq:new-0003} as
	\begin{align}
\label{eq:new-0004}
		&\mathbb{E}\|\bar{m}_{t+1}-\nabla F(w_{t+1})\|^2\\
		\nonumber\leq& (1-\frac{\nu_{t+1}}{2}+\frac{6 \nu_{t+1}}{\lambda})\mathbb{E} \|\bar{m}_{t}-\nabla F(w_{t})\|^2 \\
		\nonumber &+\frac{6 \nu_{t+1}}{\lambda} \mathbb{E} \|\nabla F(w_t) -\nabla F(w^*)\|^2\\
		\nonumber&+\frac{6 \nu_{t+1}}{\lambda} C_{\alpha}^2\sum_{j\in\mathcal{H}}\mathbb{E}\| m_t^j-\bar{m}_{t}\|^2+\frac{\nu_{t+1}^2\sigma^2}{h}.
	\end{align}
	
Due to $L$-smoothness of $F$ and $\nabla F(w^*) = 0$, it holds that
\begin{align}
	 & \|\nabla F(w_t)-\nabla F(w^*)\|^2 \\
\leq & 2L^2 \|w_t - w^*\|^2 - \|\nabla F(w_t)\|^2. \nonumber
		\label{eq:change}
\end{align}
	With (\ref{eq:change}), \eqref{eq:new-0004} turns to
	\begin{align}
		&\mathbb{E}\|\bar{m}_{t+1}-\nabla F(w_{t+1})\|^2\\
		 \nonumber\leq& (1-\frac{\nu_{t+1}}{2}+\frac{6 \nu_{t+1}}{\lambda}) \mathbb{E}\|\bar{m}_{t} -\nabla F(w_{t})\|^2 \\
		\nonumber &+\frac{12 \nu_{t+1} L^2}{\lambda}\mathbb{E}\|w_t-w^*\|^2-\frac{6 \nu_{t+1}}{\lambda}\mathbb{E}\|\nabla F(w_t)\|^2\\
		\nonumber &+\frac{6 \nu_{t+1}}{\lambda}C_{\alpha}^2\sum_{j\in\mathcal{H}}\mathbb{E}\|m_t^j-\bar{m}_t\|^2+\frac{\nu_{t+1}^2\sigma^2}{h},	
	\end{align}
which completes the proof.
\end{proof}

\subsection{Corollary of Theorem \ref{theorem:bound-T}}
\label{app:corollary-DOGD}
By Definition \ref{def:robust-bounded}, $C_{\alpha}$ for mean aggregation without Byzantine participants is equal to $0$. As a result, the adversarial regret bounds of distributed online gradient descent with mean aggregation can be obtained by dropping the terms containing $C_{\alpha}$ in Theorem \ref{theorem:bound-T}. The results are given in Corollary \ref{corollary-DOGD}.

\begin{corollary}
	\label{corollary-DOGD}
	Under Assumptions \ref{ass-1}, \ref{ass-2} and \ref{ass-4}, the distributed online gradient descent updates \eqref{eq:descent} and \eqref{eq:mean} with mean aggregation and constant step size $\eta_t = \eta \in (0,\frac{1}{4L}]$ have a linear adversarial regret bound
	\begin{align}
		R_{T:\eta} \leq  \frac{1}{\eta} \| w_1 - w^*\|^2 +(\frac{8L^2\eta^2}{\mu}+2\eta)\xi^2 T.
	\end{align}
If $\eta_t=\eta=\frac{c}{\sqrt{T}}$ where $c=\min\{\frac{1}{8L},\sqrt{\frac{\| w_1 - w^*\|^2}{2\xi^2}}\}$ in particular, then the adversarial regret bound is in the order of $\mathcal{O}(\sqrt{T})$, given by
	\begin{align}
		R_{T:\frac{1}{\sqrt{T}}} \leq \frac{8L^2c^2}{\mu}\xi^2+(\frac{\|w_1 - w^*\|^2}{c}+2c\xi^2)\sqrt{T}.
	\end{align}
	
If we use diminishing step size $\eta_t = \min\{\frac{1}{4L},\frac{8}{\mu t}\}$, the adversarial regret bound is in the order of $\mathcal{O}(\log T)$, given by
	\begin{align}
		R_{T:\frac{1}{t}}\leq 4L\| w_1 - w^*\|^2 + \frac{48L}{\mu^2}\xi^2 \log T.
	\end{align}
\end{corollary}

The $\mathcal{O}(\log T)$ adversarial regret bound is optimal in order under the $L$-smoothness and $\mu$-strong convexity assumptions \cite{wan2020projection}.

\section{Constants $C_\alpha$ of robust bounded aggregation rules}
\label{C-alpha-proof}

At step $t$ of a Byzantine-robust distributed online learning algorithm, the server aggregates the received messages to make a decision. We denote $\mathcal{Z}_t=\{z_t^1, z_t^2, \cdots, z_t^n\}$ as the set of the $n$ received messages and $\mathcal{Z}_t[k]=\{z_t^1[k], z_t^2[k], \cdots, z_t^n[k]\}$ as the set of their $k$-th elements, where $k\in[d]$.

Next, we give the definitions and the corresponding constants $C_{\alpha}$ of several popular robust bounded aggregation rules that satisfy Definition \ref{def:robust-bounded}, including coordinate-wise median \cite{yin2018byzantine}, trimmed mean \cite{yin2018byzantine,su2020byzantine}, geometric median \cite{chen2017distributed}, Krum \cite{blanchard2017machine}, centered clipping \cite{karimireddy2021learning}, Phocas \cite{xie2018phocas}, and FABA \cite{xia2019faba}.


\subsection{Coordinate-wise median}

Coordinate-wise median yields the median for each dimension, given by
\begin{align}
   & \operatorname{coomed}({\mathcal{Z}_t}) \\
:= & [\operatorname{med}(\mathcal{Z}_t[1]); \operatorname{med}(\mathcal{Z}_t[2]); \cdots; \operatorname{med}(\mathcal{Z}_t[d])], \notag
\end{align}
where $\operatorname{med}(\cdot)$ calculates the median of the input scalars.

Since $\|\bar{z}_t - z_t^j \|^2 \leq \zeta^2$ for all $j \in \mathcal{H}$ by hypothesis, we have $\|z_t^j - z_t^i\|^2\leq 2\zeta^2$ for all $i,j\in \mathcal{H}$. According to Proposition $8$ in \cite{farhadkhani2022byzantine}, if $\alpha = \frac{b}{n} < \frac{1}{2}$, it holds that
\begin{align}	 \nonumber\|\operatorname{coomed}(\mathcal{Z}_t)-\bar{z}_t\|^2
	&\leq (\frac{n}{2(n-b)}\Delta\max_{i,j\in\mathcal{H}} \|z_t^i-z_t^j\|)^2\\
	&\leq (\frac{1}{(1-\alpha)})^2 \frac{1}{2}\Delta^2\zeta^2,
\end{align}
where $\Delta =\min \{2\sqrt{n-b},\sqrt{d}\}$.

\subsection{Trimmed mean}

Trimmed mean is also coordinate-wise. Let $q < \frac{n}{2}$ be the estimated number of Byzantine participants. Given $\mathcal{Z}_t[k]$, trimmed mean removes the largest $q$ inputs and the smallest $q$ inputs, and then averages the rest. The results of the $d$ dimensions are stacked to yield
\begin{align}
   & \operatorname{trimean}({\mathcal{Z}_t}) \\
:= & [\operatorname{trimean}(\mathcal{Z}_t[1]); \operatorname{trimean}(\mathcal{Z}_t[2]); \cdots; \operatorname{trimean}(\mathcal{Z}_t[d])]. \notag
\end{align}


When the messages $z_t^j$ for all $j\in\mathcal{H}$ are i.i.d. $d$-dimensional random vectors with expectation $\bar{z}_t$ and subject to $\mathbb{E}\|z_t^j-\bar{z}_t\|^2\leq \zeta^2$, Theorem 1 of \cite{xie2018phocas} bounds $\mathbb{E} \| \operatorname{trimean}(\mathcal{Z}_t)- \bar{z}_t \|^2$.


Here we consider the deterministic case and accordingly modify the analysis. Supposing that the number of Byzantine participants are correctly estimated such that $q=b$, when $\alpha< \frac{1}{2}$ we have
\begin{align}
	\|\operatorname{trimean}(\mathcal{Z}_t)-\bar{z}_t\|^2
	&\leq \frac{2b(n-b)}{(n-2b)^2}\zeta^2\\
	\nonumber&\leq  \frac{2\alpha(1-\alpha)}{(1-2\alpha)^2}\zeta^2.
\end{align}

\subsection{Geometric median}
Geometric median finds a vector that has the minimum summed distance to the elements in the set, given by
\begin{align}
	\operatorname{geomed}(\mathcal{Z}_t):= \arg \min_{z \in \mathbb{R}^d} \sum_{j=1}^{n} \| z - z_t^j \|.
\end{align}

According to Lemma 3 in \cite{xie2018generalized}, when $\alpha < \frac{1}{2}$, we have
\begin{align}
	\|\operatorname{geomed}(\mathcal{Z}_t)\| \leq \frac{2(1-\alpha)}{1-2\alpha}r,
\end{align}
where $r$ is the upper bound of the honest messages such that $\|z_t^j\| \leq r$ for all $j \in \mathcal{H}$. Therefore, we have
\begin{align}
	  & \|\operatorname{geomed}(\mathcal{Z}_t)-\bar{z}_t\|^2 \\
	\nonumber= & \|\operatorname{geomed}(\{z_t^1-\bar{z}_t,z_t^2-\bar{z}_t,\cdots, z_t^n-\bar{z}_t\})\|^2\\
	\nonumber\leq & (\frac{2(1-\alpha)}{1-2\alpha})^2\zeta^2.
\end{align}

\subsection{Krum}
For each $z_t^j \in \mathcal{N}$, we calculate the summed distance to the $n-q-2$ nearest elements in the set $\mathcal{Z}_t$ and denote it as $s_t^j$, where $q < \frac{n}{2}$ is the estimated number of Byzantine participants. Krum finds the participant $j_t^* = \arg\min_{j \in \mathcal{N}} s_t^j$ with the smallest summed distance, and yields
\begin{align}
	\operatorname{krum}(\mathcal{Z}_t) := z_t^{j_t^*}.
\end{align}

Suppose the number of Byzantine participants is correctly estimated such that $q=b$. According to Proposition 6 in \cite{farhadkhani2022byzantine}, when $\alpha< \frac{1}{2}$, we have
\begin{align}
	\nonumber \|\operatorname{krum}(\mathcal{Z}_t)-\bar{z}_t\|^2
	&\leq ((1+\sqrt{\frac{n-b}{n-2b}})\max_{i,j\in\mathcal{H}}\|z_t^i-z_t^j\|)^2\\
	&\leq (1+\sqrt{\frac{1-\alpha}{1-2\alpha}})^2 2\zeta^2.
\end{align}

\subsection{Centered clipping}
At step $t$, centered clipping has a hyperparameter $\tau_t$ and a number of inner iterations $\iota_{\max}$. It iteratively runs
\begin{align}
\hspace{-1.5em}	v_{t,\iota}= v_{t,\iota-1} +  \frac{1}{n}\sum_{j=1}^{n}(z_t^j-v_{t,\iota-1})\min (1,\frac{\tau_t}{\vert|z_t^j-v_{t,\iota-1}\vert|}),
\end{align}
for $\iota = 1, 2, \cdots, \iota_{\max}$, initialized by $v_{t,0} = w_{t-1}$ that is the aggregation result of the last step. Then, centered clipping yields
\begin{align}
	\operatorname{cc}(\mathcal{Z}_t):=v_{t,\iota_{\max}}.
\end{align}

When the messages $z_t^j$ for all $j\in\mathcal{H}$ are i.i.d. $d$-dimensional random vectors with expectation $\bar{z}_t$ and subject to $\mathbb{E}\|z_t^j-\bar{z}_t\|^2\leq \zeta^2$, Theorem 4 of \cite{karimireddy2021learning} bounds $\mathbb{E} \| \operatorname{cc}(\mathcal{Z}_t)- \bar{z}_t \|^2$.

Here we consider the deterministic case and accordingly modify the analysis. Supposing that the hyperparameter $\tau_t$ is properly chosen, when $\alpha \leq 0.1$ we have
\begin{align}
	\|\operatorname{cc}(Z_t)-\bar{z}_t\|^2 \leq (9.7\alpha)^{\iota_{\max}}\|v_{t,0}-\bar{z}_t\|^2+8000\alpha \zeta^2.
\end{align}

\subsection{Phocas}
Let $q < \frac{n}{2}$ be the estimated number of Byzantine participants. Phocas finds the $n-q$ nearest messages to the trimmed mean of $\mathcal{Z}_t$, places them in a set $\mathcal{V}_t$, and outputs the average as
\begin{align}
	\operatorname{phocas}(\mathcal{Z}_t) := \frac{\sum_{v\in \mathcal{V}_t}v}{n-q}.
\end{align}

When the messages $z_t^j$ for all $j\in\mathcal{H}$ are i.i.d. $d$-dimensional random vectors with expectation $\bar{z}_t$ and subject to $\mathbb{E}\|z_t^j-\bar{z}_t\|^2\leq \zeta^2$, Theorem 1 of \cite{xie2018phocas} bounds $\mathbb{E} \| \operatorname{phocas}(\mathcal{Z}_t)- \bar{z}_t \|^2$.

Here we consider the deterministic case and accordingly modify the analysis. Supposing that the number of Byzantine participants are correctly estimated such that $q=b$, when $\alpha< \frac{1}{2}$ we have
\begin{align}
	\|\operatorname{phocas}(\mathcal{Z}_t)-\bar{z}_t\|^2 \leq (4+\frac{12\alpha(1-\alpha)}{(1-2\alpha)^2})\zeta^2.
\end{align}

\subsection{FABA}
Let $q < \frac{n}{3}$ be the estimated number of Byzantine participants. FABA iteratively discards $q$ outliers from $\mathcal{Z}_t$. Starting from a remaining set $\mathcal{U}_t = \mathcal{Z}_t$, at each inner iteration, FABA calculates the mean of $\mathcal{U}_t$, deletes the message that is the farthest the mean to form the new remaining set. Eventually, FABA outputs the mean of the remaining set as
\begin{align}
	\operatorname{faba}(\mathcal{Z}_t) := \frac{\sum_{u\in \mathcal{U}_t}u}{n-q}.
\end{align}

	Suppose the number of Byzantine participants is correctly estimated such that $q=b$ and $\alpha<\frac{1}{3}$.
	If all Byzantine messages are removed, the corresponding constant $C_{\alpha}=0$.
	Otherwise, we have
	\begin{align} \hspace{-1em} \|\operatorname{faba}(\mathcal{Z}_t)-\bar{z}_t\|^2
		 \leq 4  \left(\frac{b}{n-b} + \frac{n+1-b}{n-b}\frac{b}{n-3b}\right) \zeta^2 .
	\end{align}
	
	\begin{proof}
At inner iteration $\iota$, define the remaining set as $\mathcal{U}_{t,\iota}$, while $\mathcal{H}_{t,\iota} = \mathcal{H} \cap \mathcal{U}_{t,\iota}$ and $\mathcal{B}_{t,\iota} = \mathcal{B} \cap \mathcal{U}_{t,\iota}$. Also define
$\bar z^{\mathcal{U}}_{t,\iota} = \frac{1}{|\mathcal{U}_{t,\iota}|} \sum_{j \in \mathcal{U}_{t,\iota}} z_t^j$, $\bar z_{t,\iota}^{\mathcal{H}} = \frac{1}{|\mathcal{H}_{t,\iota}|} \sum_{j \in \mathcal{H}_{t,\iota}} z_t^j$, $\bar z_{t,\iota}^{\mathcal{B}} = \frac{1}{|\mathcal{B}_{t,\iota}|} \sum_{j \in \mathcal{B}_{t,\iota}} z_t^j$ and $\bar z_t = \frac{1}{h} \sum_{j \in \mathcal{H}} z_t^j$. It is obvious that
		\begin{align}
			\label{fb-1}
				\bar z_{t,\iota}^{\mathcal{U}} =\left( 1- \frac{|\mathcal{B}_{t,\iota}|}{|\mathcal{U}_{t,\iota}|} \right) \bar z_{t,\iota}^{\mathcal{H}} + \frac{|\mathcal{B}_{t,\iota}|}{|\mathcal{U}_{t,\iota}|}\bar z_{t,\iota}^{\mathcal{B}}.
		\end{align}
		
		\textbf{Case 1:} At inner iteration $\iota$, according to Theorem 3 in \cite{wu2022byzantine}, if $\|\bar z_{t,\iota}^{\mathcal{H}} - \bar z_{t,\iota}^{\mathcal{B}} \| > \max_{j \in \mathcal{H}}\|z_t^j - \bar z_{t,\iota}^{\mathcal{H}} \|/(1-2\frac{|\mathcal{B}_{t,\iota}|}{|\mathcal{U}_{t,\iota}|})$, then a Byzantine message will be successfully removed.
		
		\textbf{Case 2:} Otherwise, if $\|\bar z_{t,\iota}^{\mathcal{H}} - \bar z_{t,\iota}^{\mathcal{B}} \| \leq \max_{j \in \mathcal{H}}\|z_t^j - \bar z_{t,\iota}^{\mathcal{H}} \|/(1-2\frac{|\mathcal{B}_{t,\iota}|}{|\mathcal{U}_{t,\iota}|})$, then a message denoted as $z_{t,\iota}^{j}$,
		that may be either Byzantine or honest,
	 will be removed. But we can assume it is honest; otherwise, FABA correctly removes a Byzantine message as described in Case 1. This case yields
		\begin{align}
			\label{fb-2}
				& \|\bar{z}_{t,\iota+1}^{\mathcal{U}}-\bar{z}_t \| \\
				\leq&  \|\bar{z}_{t,\iota+1}^{\mathcal{U}}- \bar{z}_{t,\iota}^{\mathcal{U}} \| + \|\bar{z}_{t,\iota}^{\mathcal{U}}- \bar z_{t,\iota}^{\mathcal{H}} \|+ \|\bar{z}_{t,\iota}^{\mathcal{H}}-\bar{z}_t \| \nonumber\\
				\leq & \frac{1}{|\mathcal{U}_{t,\iota+1}|} \|z_{t,\iota}^{j}- \bar{z}_{t,\iota}^{\mathcal{U}}\| + \|\bar{z}_{t,\iota}^{\mathcal{H}}-\bar{z}_t \|+ \|\bar{z}_{t,\iota}^{\mathcal{U}}- \bar z_{t,\iota}^{\mathcal{H}} \| \nonumber\\
				\leq & \frac{1}{|\mathcal{U}_{t,\iota+1}|} \|z_{t,\iota}^{j}- \bar{z}_{t,\iota}^{\mathcal{H}}\| + \|\bar{z}_{t,\iota}^{\mathcal{H}}-\bar{z}_t \| \nonumber\\
				&  + \left(1+\frac{1}{|\mathcal{U}_{t,\iota+1}|}\right)\|\bar{z}_{t,\iota}^{\mathcal{U}}- \bar z_{t,\iota}^{\mathcal{H}} \| \nonumber\\
				\leq & \frac{1}{|\mathcal{U}_{t,\iota+1}|} \|z_{t,\iota}^{j}- \bar{z}_{t,\iota}^{\mathcal{H}}\| + \|\bar{z}_{t,\iota}^{\mathcal{H}}-\bar{z}_t \| \nonumber\\
				&+ \frac{n+1-b}{n-b}\|\bar{z}_{t,\iota}^{\mathcal{U}}- \bar z_{t,\iota}^{\mathcal{H}} \|, \nonumber
		\end{align}
		where the second inequality is due to $\|\bar{z}_{t,\iota+1}^{\mathcal{U}}- \bar{z}_{t,\iota}^{\mathcal{U}} \| \leq  \frac{1}{|\mathcal{U}_{t,\iota+1}|} \|z_{t,\iota}^{j}- \bar{z}_{t,\iota}^{\mathcal{U}}\|$ given by Lemma 3 in \cite{wu2022byzantine}.
	
		For the second term at the right-hand side of (\ref{fb-2}), according to Lemma 3 in \cite{wu2022byzantine}, we have
		\begin{align}
			\label{fb-4}
				& \|\bar{z}_{t,\iota}^{\mathcal{H}}-\bar{z}_t \| \\
				\leq&  \frac{n-b-|\mathcal{H}_{t,\iota}|}{n-b}\max_{j \in \mathcal{H}}\|z_t^j - \bar z_{t,\iota}^{\mathcal{H}}  \|. \nonumber
		\end{align}
	
		For the third term at the right-hand side of (\ref{fb-2}), according to (\ref{fb-1}) we have
		\begin{align}
			\label{fb-3}
				& \|\bar{z}_{t,\iota}^{\mathcal{U}}- \bar z_{t,\iota}^{\mathcal{H}} \| \\
				=& \frac{|\mathcal{B}_{t,\iota}|}{|\mathcal{U}_{t,\iota}|} \|\bar{z}_{t,\iota}^{\mathcal{H}}- \bar z_{t,\iota}^{\mathcal{B}} \| \nonumber\\
				\leq& \frac{\frac{|\mathcal{B}_{t,\iota}|}{|\mathcal{U}_{t,\iota}|}}{1-2\frac{|\mathcal{B}_{t,\iota}|}{|\mathcal{U}_{t,\iota}|}} \max_{j \in \mathcal{H}}\|z_t^j - \bar z_{t,\iota}^{\mathcal{H}} \|. \nonumber
		\end{align}

		Substituting  (\ref{fb-4}) and (\ref{fb-3}) into (\ref{fb-2}), we obtain
		\begin{align}
			\label{fb-5}
				& \|\bar{z}_{t,\iota+1}^{\mathcal{U}}-\bar{z}_t \| \\
				\leq & \left( \frac{1}{|\mathcal{U}_{t,\iota+1}|}+  \frac{n-b-|\mathcal{H}_{t,\iota}|}{n-b} \right) \max_{j \in \mathcal{H}}\|z_t^j - \bar z_{t,\iota}^{\mathcal{H}}  \| \nonumber\\
				&+ \frac{n+1-b}{n-b}\frac{\frac{|\mathcal{B}_{t,\iota}|}{|\mathcal{U}_{t,\iota}|}}{1-2\frac{|\mathcal{B}_{t,\iota}|}{|\mathcal{U}_{t,\iota}|}}
				\max_{j \in \mathcal{H}}\|z_t^j - \bar z_{t,\iota}^{\mathcal{H}}  \|. \nonumber
		\end{align}
		Note that
		\begin{align}
			\label{fb-6}
				& \frac{1}{|\mathcal{U}_{t,\iota+1}|}+ \frac{n-b-|\mathcal{H}_{t,\iota}|}{n-b} \\
				\leq & \frac{|\mathcal{H}|+|\mathcal{B}|+1-b-|\mathcal{H}_{t,\iota}|}{n-b} \nonumber\\
				\leq &  \frac{|\mathcal{H}|+1-|\mathcal{H}_{t,\iota}|}{n-b}  \leq \frac{\iota}{n-b} \leq \frac{b}{n-b}. \nonumber
		\end{align}
		Further, since ${|\mathcal{B}_{t,\iota}|}/{|\mathcal{U}_{t,\iota}|} \leq \frac{b}{n-b}$, we have
		\begin{align}
			\label{fb-8}
				 \frac{n+1-b}{n-b}\frac{\frac{|\mathcal{B}_{t,\iota}|}{|\mathcal{U}_{t,\iota}|}}{1-2\frac{|\mathcal{B}_{t,\iota}|}{|\mathcal{U}_{t,\iota}|}} \leq \frac{n+1-b}{n-b}\frac{b}{n-3b}.
		\end{align}
		Combining (\ref{fb-6}) and (\ref{fb-8}) with (\ref{fb-5}) yields
		\begin{align}
			\label{fb-9}
				& \|\bar{z}_{t,\iota+1}^{\mathcal{U}}-\bar{z}_t \| \\
				\leq &\left( \frac{b+1}{n-b} + \frac{n+1-b}{n-b}\frac{b}{n-3b}  \right)  \max_{j \in \mathcal{H}}\|z_t^j - \bar z_{t,\iota}^{\mathcal{H}}  \|. \nonumber
		\end{align}
which ends the discussion of Case 2.


If FABA successfully eliminates one Byzantine message in each inner iteration, the condition $q=b$ implies that all Byzantine messages have been removed, resulting in $\|\bar z_{t,b}^{\mathcal{U}}-\bar z_t\|=0$. Instead, if at least one honest message has been removed during the inner iterations, we denote $\iota^*+1$ as the last inner iteration that eliminates the honest message, which only occurs in Case 2. For this scenario, observe that
\begin{align}
	\label{last-iteration}
	 & \|\bar z_{t,b}^{\mathcal{U}}-\bar z_t\| \\
\leq & \|\bar z_{t,b}^{\mathcal{U}} - \bar z_{t,\iota^*}^{\mathcal{U}}\|+\| \bar z_{t,\iota^*}^{\mathcal{U}}-\bar z_{t,\iota^*+1}^{\mathcal{U}}\|+\| \bar z_{t,\iota^*+1}^{\mathcal{U}}-\bar z_t\|. \nonumber
\end{align}

The first term at the right-hand side of (\ref{last-iteration}) can be bounded using Lemma 3 in \cite{wu2022byzantine}, as
\begin{align}
	\label{fb-12}
	\|\bar z_{t,b}^{\mathcal{U}} - \bar z_{t,\iota^*}^{\mathcal{U}}\|
	&\leq  \frac{b-\iota^*}{n-b}\max_{z_{t}^j\in \mathcal{U}_{t,\iota^*,b}}\|z_{t}^j - \bar z_{t,\iota^*}^{\mathcal{U}}\|\\
	\nonumber &\leq \frac{b}{n-b} \|z_{t,\iota^*+1}- \bar z_{t,\iota^*}^{\mathcal{U}}\|,
\end{align}
where $\mathcal{U}_{t,\iota^*,b} := \mathcal{U}_{t,\iota^*}/\mathcal{U}_{t,b}$ denotes the removed messages from inner iteration $\iota^*+1$ to $b$ . The second inequality comes from the removal rule which shows $z_{t,\iota^*+1}$ to be the furthest message to $\bar z_{t,\iota^*}^{\mathcal{U}}$ in $\mathcal{U}_{t,\iota^*}$, namely
\begin{align}
	\label{fb-13}
	\max_{z_{t}^j\in \mathcal{U}_{t,\iota^*,b}}\|z_{t}^j - \bar z_{t,\iota^*}^{\mathcal{U}}\| =\|z_{t,\iota^*+1}- \bar z_{t,\iota^*}^{\mathcal{U}}\|.
\end{align}
On the other hand, the definition of $\iota^*+1$ ensures $z_{t,\iota^*+1} \in \mathcal{H}_{t,\iota^*}$, and thus
\begin{align}
	\label{fb-14}
	\|z_{t,\iota^*+1}- \bar z_{t,\iota^*}^{\mathcal{U}}\|
	&\leq \|z_{t,\iota^*+1}- \bar z_{t,\iota^*}^{\mathcal{H}}\| + \|z_{t,\iota^*}^{\mathcal{H}}- \bar z_{t,\iota^*}^{\mathcal{U}}\| \\
	\nonumber &\leq (1+\frac{\frac{|\mathcal{B}_{t,\iota}|}{|\mathcal{U}_{t,\iota}|}}{1-2\frac{|\mathcal{B}_{t,\iota}|}{|\mathcal{U}_{t,\iota}|}})\max_{j\in\mathcal{H}} \|z_t^j-z_{t,\iota^*}^{\mathcal{H}}\| \\
	\nonumber &\leq (1+\frac{b}{n-3b})\max_{j\in\mathcal{H}}\|z_t^j-z_{t,\iota^*}^{\mathcal{H}}\|,
\end{align}
where the second inequality comes from (\ref{fb-3}) and the third inequality comes from (\ref{fb-8}).

The second term at the right-hand side of (\ref{last-iteration}) can be bounded using Lemma 3 in \cite{wu2022byzantine}, (\ref{fb-3}) and (\ref{fb-8}), as
\begin{align}
	\label{fb-15}
	\| \bar z_{t,\iota^*}^{\mathcal{U}}-\bar z_{t,\iota^*+1}^{\mathcal{U}}\|
	&\leq \frac{1}{|\mathcal{U}_{t,\iota^*+1}|} \|z_{t,\iota^*+1}^{j}- \bar{z}_{t,\iota^*}^{\mathcal{U}}\|\\
	\nonumber&\leq \frac{1}{|\mathcal{U}_{t,\iota^*+1}|} (\|z_{t,\iota^*+1}^{j}- \bar{z}_{t,\iota^*}^{\mathcal{H}}\|+ \|\bar{z}_{t,\iota^*}^{\mathcal{H}} -\bar{z}_{t,\iota^*}^{\mathcal{U}}\|)\\
	\nonumber &\leq \frac{1}{n-b}(1+\frac{b}{n-3b})\max_{j\in\mathcal{H}}\|z_t^j-z_{t,\iota^*}^{\mathcal{H}}\|.
\end{align}

The third term at the right-hand side of (\ref{last-iteration}) can be bounded by (\ref{fb-9}), as
		\begin{align}
	\label{fb-16}
		& \|\bar{z}_{t,\iota^*+1}^{\mathcal{U}}-\bar{z}_t \| \\
		\nonumber\leq &\left( \frac{b+1}{n-b} + \frac{n+1-b}{n-b}\frac{b}{n-3b}  \right)  \max_{j \in \mathcal{H}}\|z_t^j - \bar z_{t,\iota^*}^{\mathcal{H}}  \|.
\end{align}

Substituting (\ref{fb-13}), (\ref{fb-14}) (\ref{fb-15}) and (\ref{fb-16}) back into (\ref{fb-12}) yields
\begin{align}
	\label{fb-17}
	&\|\bar z_{t,b}^{\mathcal{U}}-\bar z_t\| \\
	\nonumber &\leq \frac{b}{n-b}(1+\frac{b}{n-3b})\max_{j \in \mathcal{H}}\|z_t^j - \bar z_{t,\iota^*}^{\mathcal{H}}  \|	\\
	\nonumber&+\frac{1}{n-b}(1+\frac{b}{n-3b})\max_{j \in \mathcal{H}}\|z_t^j - \bar z_{t,\iota^*}^{\mathcal{H}}\|\\
	\nonumber&+\left( \frac{b+1}{n-b} + \frac{n+1-b}{n-b}\frac{b}{n-3b}  \right)  \max_{j \in \mathcal{H}}\|z_t^j - \bar z_{t,\iota^*}^{\mathcal{H}}\|\\
	\nonumber &\leq \left(\frac{b+1}{n-b}+\frac{b+1}{n-b}\frac{b}{n-3b}\right) \max_{j \in \mathcal{H}}\|z_t^j - \bar z_{t,\iota^*}^{\mathcal{H}}\|\\
	\nonumber & + \left( \frac{b+1}{n-b} + \frac{n+1-b}{n-b}\frac{b}{n-3b}  \right)  \max_{j \in \mathcal{H}}\|z_t^j - \bar z_{t,\iota^*}^{\mathcal{H}}\|\\
	\nonumber &\leq 2\left( \frac{b+1}{n-b} + \frac{n+1-b}{n-b}\frac{b}{n-3b}  \right) \max_{j \in \mathcal{H}}\|z_t^j - \bar z_{t,\iota^*}^{\mathcal{H}}\|,
\end{align}
where the last inequality comes from $b< n-b$.
Obviously, we have
	\begin{align}
		\label{fb-10}
			\max_{j \in \mathcal{H}}\|z_t^j - \bar z_{t,\iota}^{\mathcal{H}}  \|
			&= \max_{j \in \mathcal{H}}\|z_t^j - \bar{z}_t+ \bar{z}_t- \bar z_{t,\iota}^{\mathcal{H}}  \|\\
			\nonumber&\leq  2\max_{j \in \mathcal{H}}\|z_t^j - \bar{z}_t \|.
	\end{align}
Combining (\ref{fb-17}) with (\ref{fb-10}), we have
\begin{align}
	&\|\bar z_{t,b}^{\mathcal{U}}-\bar z_t\| \\
	\nonumber &\leq 4\left( \frac{b+1}{n-b} + \frac{n+1-b}{n-b}\frac{b}{n-3b}  \right) \max_{j \in \mathcal{H}}\|z_t^j - \bar z_t\|,
\end{align}
which completes the proof.
\end{proof}

\section{Resnet18 training on the CIFAR10 dataset}
\label{experimentCifar10}
The previous numerical experiments involve convex losses. Now, we consider nonconvex losses. We train a Resnet18 neural network on the CIFAR10 dataset, which contains 50,000 training samples and 10,000 testing samples. The Resnet18 neural network is initialized using pre-trained parameters from the PyTorch Resnet18 package. The batch size is set as 32 during training. We launch one server and 5 participants, and consider two data distributions. In the i.i.d. setting, all training samples are divided evenly among all participants. In the non-i.i.d. setting, each class of training samples is assigned to 2 participants and eventually each participant has training samples from 4 classes. Under Byzantine attacks, 1 randomly chosen participant is adversarial.

The performance metrics are classification accuracy on the testing samples and adversarial regret on the training samples.

As in most neural network training tasks, we use the diminishing step size $\eta_t$. It starts from $0.05$ and decreases by a factor of $0.85$ after every four epochs. The momentum parameter $\nu_t$ is set as $2\eta_t$.

\begin{figure*}
	\begin{center}
		\centerline{\includegraphics[width=0.8\textwidth]{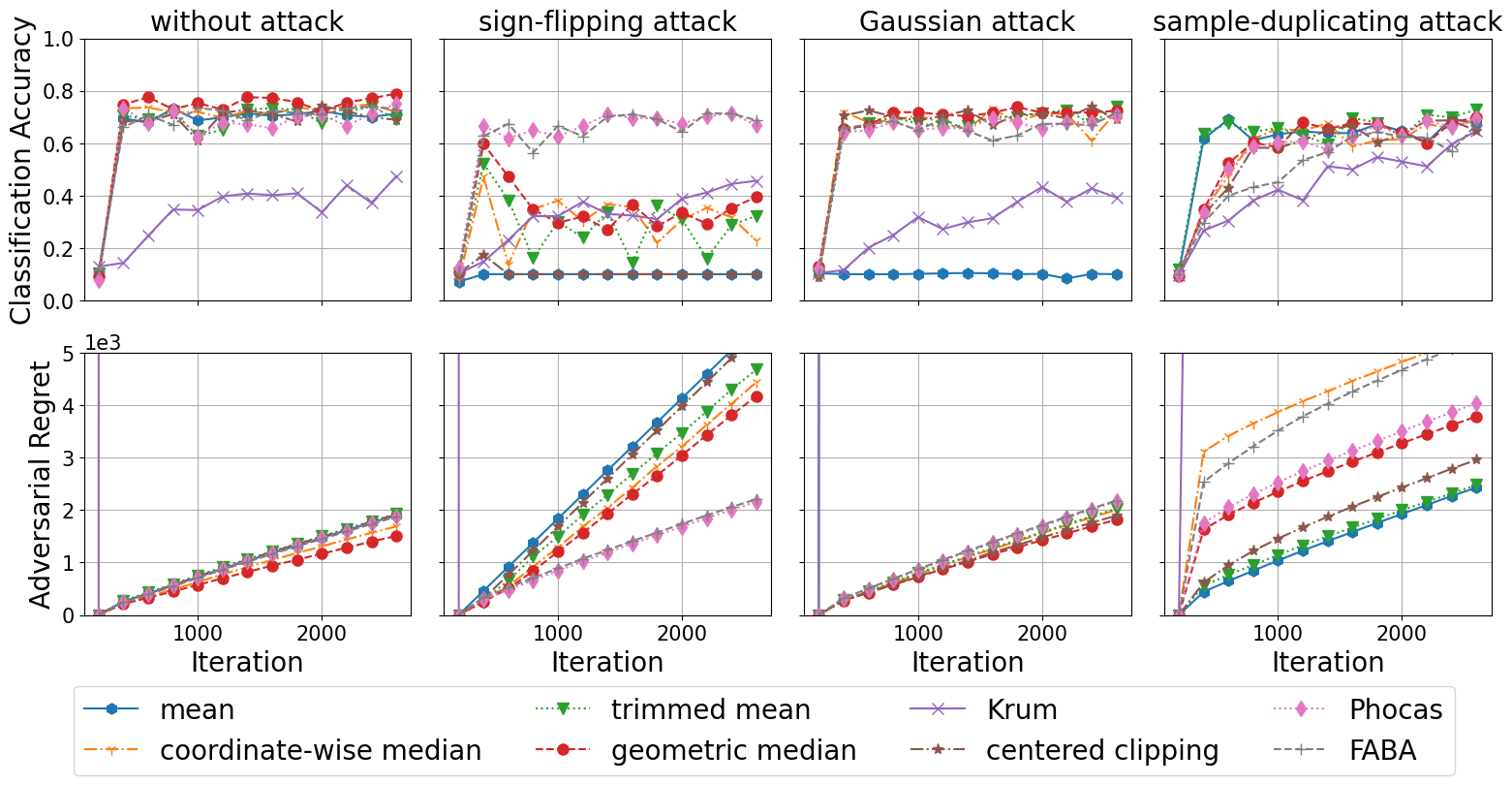}}
		\caption{Performance of Byzantine-robust distributed online gradient descent on CIFAR10 i.i.d. data with diminishing step size.}
		\label{fig:iid-diminishing-DROGD-cifar10}
	\end{center}
	\vspace{-2em}
\end{figure*}
\begin{figure*}
	\begin{center}
		\centerline{\includegraphics[width=0.8\textwidth]{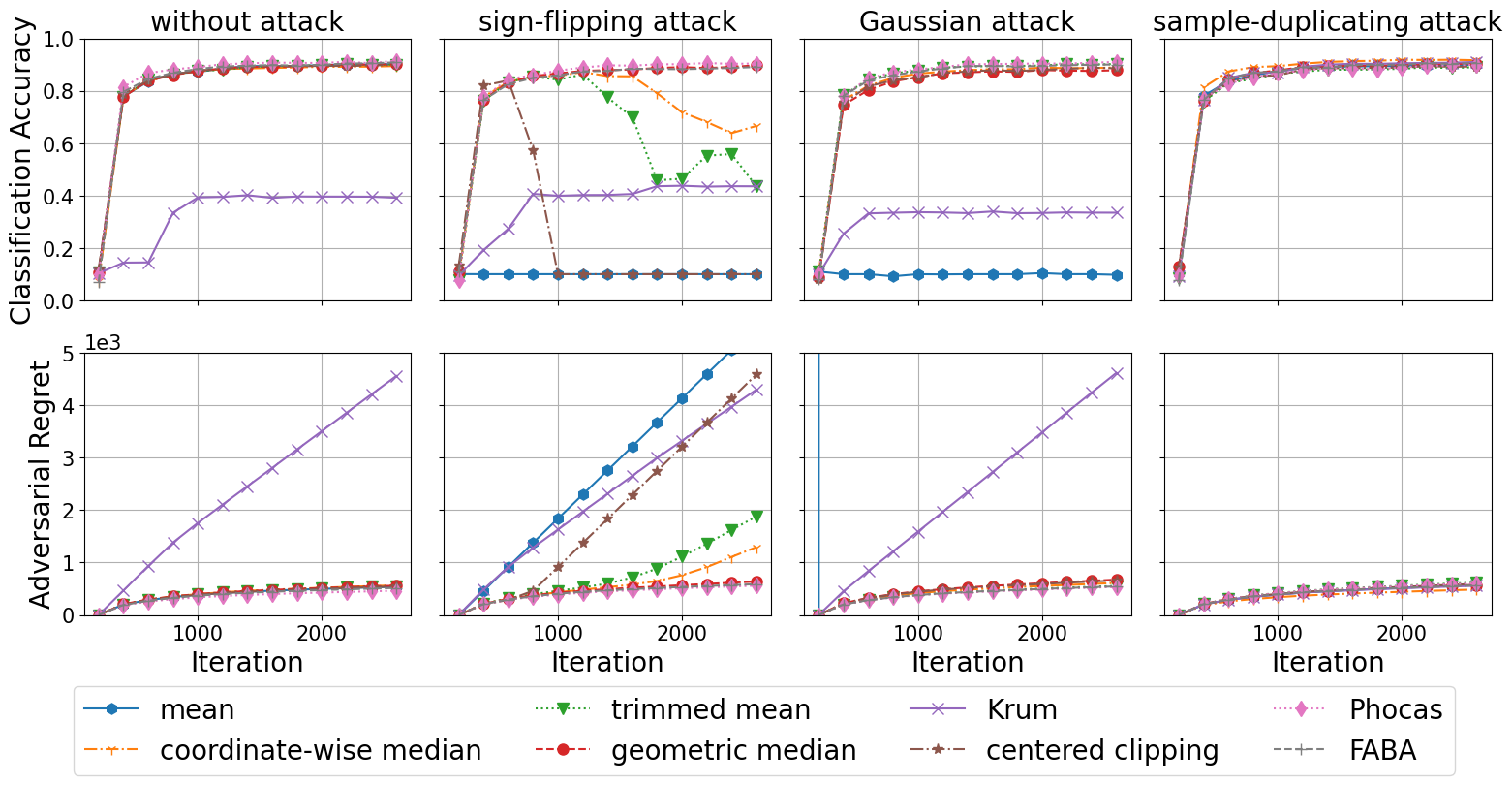}}
		\caption{Performance of Byzantine-robust distributed online momentum gradient descent on CIFAR10 i.i.d. data with diminishing step size.}
		\label{fig:iid-diminishing-DROGDM-cifar10}
	\end{center}
	\vspace{-2em}
\end{figure*}

\noindent \textbf{Numerical experiments on i.i.d. data.}
As shown in Fig. \ref{fig:iid-diminishing-DROGD-cifar10}, on the i.i.d. data, the Byzantine-robust distributed online gradient descent algorithms perform well under Byzantine attacks.
However, their performance is inferior to that with momentum, as shown in Fig. \ref{fig:iid-diminishing-DROGDM-cifar10}. This phenomenon is consistent with the convex case.

\begin{figure*}
	\begin{center}
		\centerline{\includegraphics[width=0.8\textwidth]{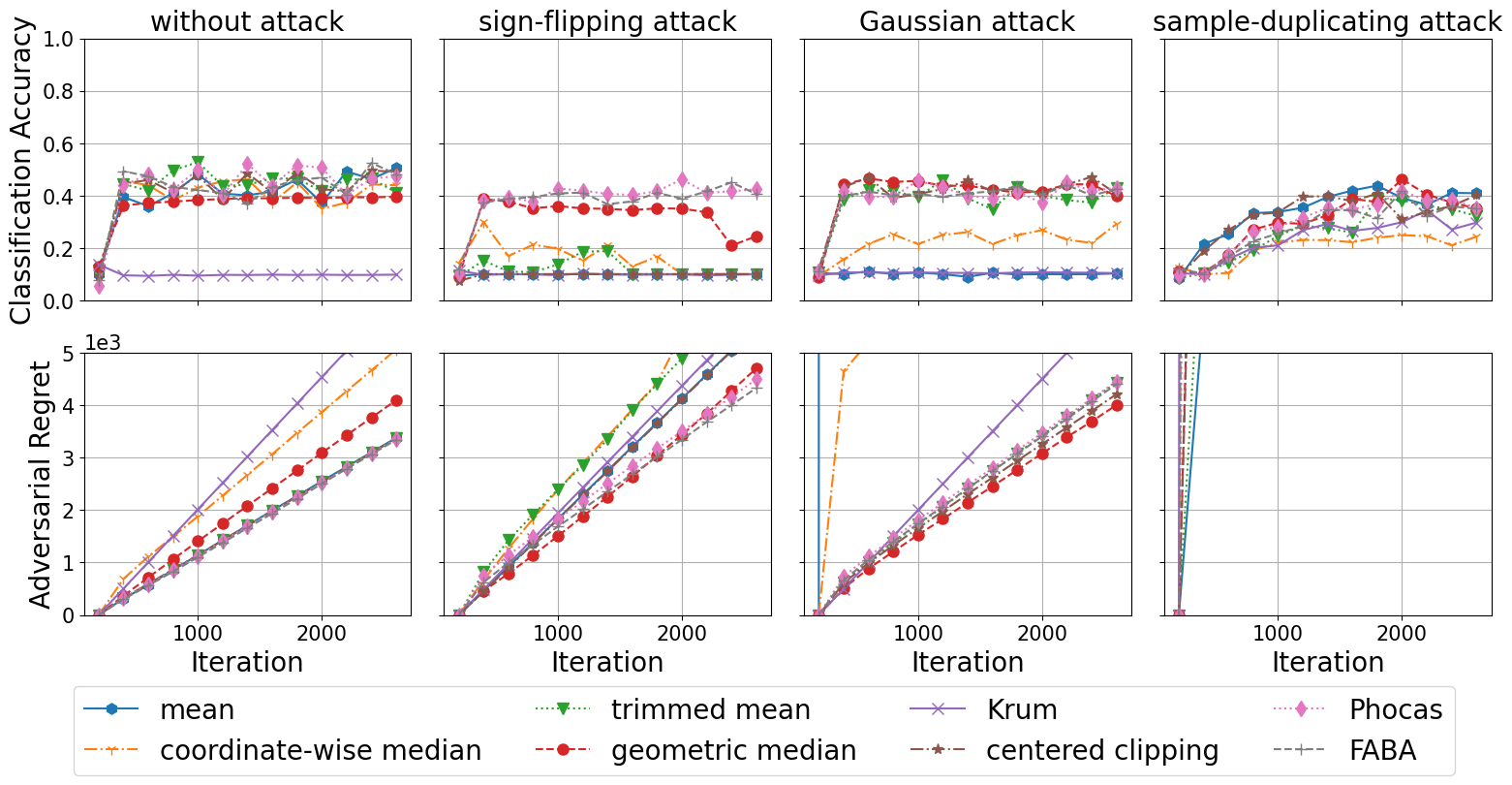}}
		\caption{Performance of Byzantine-robust distributed online gradient descent on CIFAR10 non-i.i.d. data with diminishing step size.}
		\label{fig:noniid-diminishing-DROGD-cifar10}
	\end{center}
	\vspace{-2em}
\end{figure*}

\begin{figure*}
	\begin{center}
		\centerline{\includegraphics[width=0.8\textwidth]{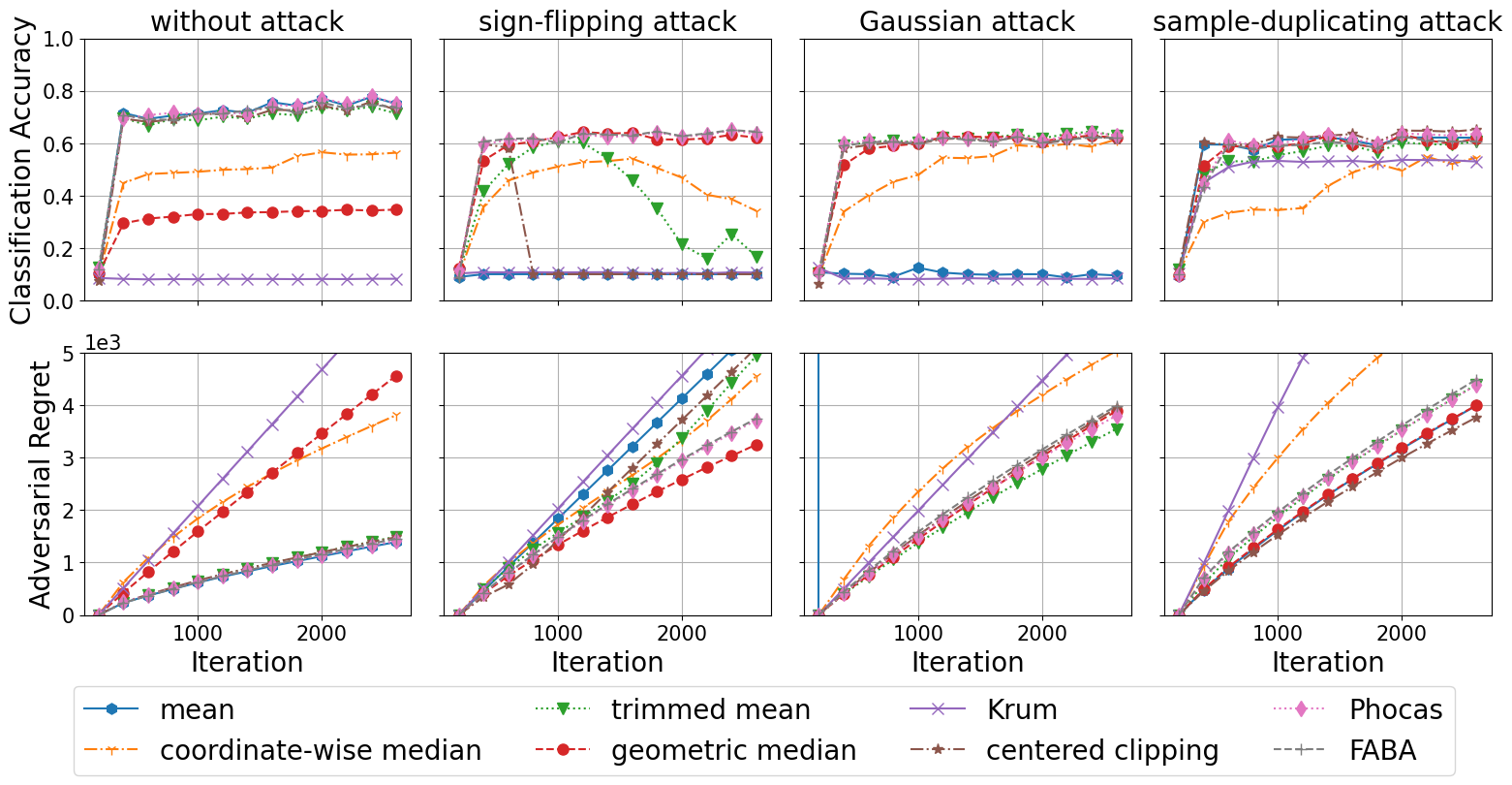}}
		\caption{Performance of Byzantine-robust distributed online momentum gradient descent on CIFAR10 non-i.i.d. data with diminishing step size.}
		\label{fig:noniid-diminishing-DROGDM-cifar10}
	\end{center}
	\vspace{-2em}
\end{figure*}

\noindent \textbf{Numerical experiments on non-i.i.d. data.}
As illustrated in Figs. \ref{fig:noniid-diminishing-DROGD-cifar10} and \ref{fig:noniid-diminishing-DROGDM-cifar10}, on the non-i.i.d. data, the Byzantine-robust distributed online gradient descent algorithms, no matter with or without momentum, perform worse than on the i.i.d. data. Nevertheless, they still show certain robustness to Byzantine attacks.


\end{document}